\documentclass[10pt]{IEEEtran}
\usepackage{amssymb}
\usepackage{graphicx}
\usepackage{url}
\usepackage{multirow}
\usepackage[linesnumbered,ruled]{algorithm2e}
\usepackage[caption=false,font=footnotesize]{subfig}
\usepackage{mathtools}
\usepackage{color}
\usepackage{capt-of}

\SetCommentSty{mycommfont}

\def\i{\boldsymbol{i}}
\def\c{\boldsymbol{c}}
\def\x{\boldsymbol{x}}
\def\b{\boldsymbol{b}}
\def\y{\boldsymbol{y}}
\def\j{\boldsymbol{j}}
\def\p{\boldsymbol{p}}
\def\f{\boldsymbol{f}}
\def\g{\boldsymbol{g}}

\def\amu{\boldsymbol{\mu}} 
\def\C{\mathcal{C}}

\newtheorem{theorem}{Theorem}
\newcommand\tabcaption{\def\@captype{table}\caption}
\newcommand\figcaption{\def\@captype{figure}\caption}

\begin{document}

% ---- Title -----------------------------------------
\title{Fast High-Dimensional Bilateral and \\ Nonlocal Means Filtering}
% ----------------------------------------------------

% ---- Authors ---------------------------------------
\author{Pravin~Nair,~\IEEEmembership{Student Member,~IEEE,} and~Kunal~N.~Chaudhury,~\IEEEmembership{Senior~Member,~IEEE}
        \thanks{The authors are with the Department of Electrical Engineering, Indian Institute of Science, Bangalore 560012, India. Correspondence: kunal@iisc.ac.in. K.~N.~Chaudhury was supported by a Startup Grant from Indian Institute of Science and EMR Grant SERB/F/6047/2016-2017 from the Department of Science and Technology, Government of India.}}

% ----Header ------------------------------
\markboth{IEEE Transactions on Image Processing}{}
% ----------------------------------------------------

% ---- Title ---------------------------
\maketitle
%\IEEEpeerreviewmaketitle
% ----------------------------------------------------

% ---- Abstract -----------------------------
\begin{abstract}
Existing fast algorithms for bilateral and nonlocal means filtering mostly work with grayscale images. They cannot easily be extended to high-dimensional data such as color and hyperspectral images, patch-based data, flow-fields, etc. In this paper, we propose a fast algorithm for high-dimensional bilateral and nonlocal means filtering. Unlike existing approaches, where the focus is on approximating the data (using quantization) or the filter kernel (via analytic expansions), we locally approximate the kernel using weighted and shifted copies of a Gaussian, where the weights and shifts are inferred from the data. The algorithm emerging from the proposed approximation essentially involves clustering and fast convolutions, and is easy to implement. Moreover, a variant of our algorithm comes with a guarantee (bound) on the approximation error, which is not enjoyed by existing algorithms. We present some results for high-dimensional bilateral and nonlocal means filtering to demonstrate the speed and accuracy of our proposal. Moreover, we also show that our algorithm can outperform state-of-the-art fast approximations in terms of accuracy and timing. 
\end{abstract} 
% -----------------------------------------------------

% ---- Keywords ---------------------------------------
% -----------------------------------------------------

\section{Introduction}

Smoothing images while preserving structures (edges, corners, lines, etc.) is a fundamental task in image processing.
A classic example in this regard is the diffusion framework of Perona and Malik \cite{perona1990scale}.
In the last few decades, several filtering based approaches have been proposed for this task.
Prominent examples include bilateral filtering \cite{aurich1995non,smith1997susan,tomasi1998bilateral}, mean shift filtering \cite{comaniciu1999mean}, weighted least squares \cite{farbman2008edge}, domain transform \cite{gastal2011domain}, guided filtering \cite{he2013guided}, and nonlocal means \cite{buades2005non}. 
The brute-force implementation of most of these filters is computationally prohibitive and cannot be used for real-time applications. To address this problem, researchers have come up with approximation algorithms that can significantly accelerate the filtering without compromising the quality. 
Unfortunately, for bilateral and nonlocal means filtering, most of these algorithms can be used only for grayscale images.
It is difficult to use them even for color filtering, while preserving their efficiency. 
The situation is more challenging for multispectral and hyperspectral images, where the range dimension is much larger.
Of course, any algorithm for grayscale filtering can be used for channel-by-channel processing of high-dimensional images.
However, by working in the combined intensity space, we can exploit the strong correlation between channels \cite{tomasi1998bilateral}. 

\subsection{High-Dimensional Filtering}

The focus of the present work is on two popular smoothers, the bilateral and the nonlocal means filters, and their application to high-dimensional data. The former is used for edge-preserving smoothing in a variety of applications \cite{paris2009bilateral}, while the latter is primarily used for denoising. 
Though nonlocal means has limited denoising capability compared to state-of-the-art denoisers \cite{dabov2006image,zhang2017beyond}, it continues to be of interest due to its simplicity and the availability of low-complexity algorithms \cite{darbon2008,gastal2012adaptive,wang2006}. 
The connection between these filters is that they can be interpreted as a multidimensional Gaussian filter operating in the joint spatio-range space~\cite{chen2007real}. 
The term \textit{high-dimensional filtering} is used when the dimension of the spatio-range space is large \cite{gastal2012adaptive,adams2010fast,adams2009gaussian}. In this paper, we will use this term when the range dimension is greater than one.

An unified formulation of high-dimensional bilateral and nonlocal means filtering  is as follows. Suppose that the input data is $\f: \Omega  \to [0,R]^n$, where $\Omega \subset \mathbb{Z}^{d}$ is the spatial domain, $[0,R]^n$ is the range space, and $d$ and $n$ are dimensions of the domain and range of $\f$. 
Let $\p : \Omega  \to \mathbb{R}^{\rho}$ be the \textit{guide} image, which is used to control the filtering \cite{paris2009bilateral}. The output $\g: \Omega  \to [0,R]^n$ is given by
\begin{equation}
 \label{num}
\g(\i) = \frac{1}{\eta({\i})}{\smashoperator[r]{\sum_{\j\in W}} \omega(\j) \ \varphi\big(\p(\i-\j) - \p(\i)\big) \f(\i-\j)},
\end{equation}
where 
\begin{equation}
\label{den}
\eta(\i) = {\smashoperator[r]{\sum_{\j\in W}} \omega(\j) \ \varphi\big(\p(\i-\j) - \p(\i)\big)}.
\end{equation}
\begin{figure*}
\centering
\subfloat[ \textit{Lena} ($3 \times 512 \times 512$)]{\includegraphics[width=0.2\linewidth]{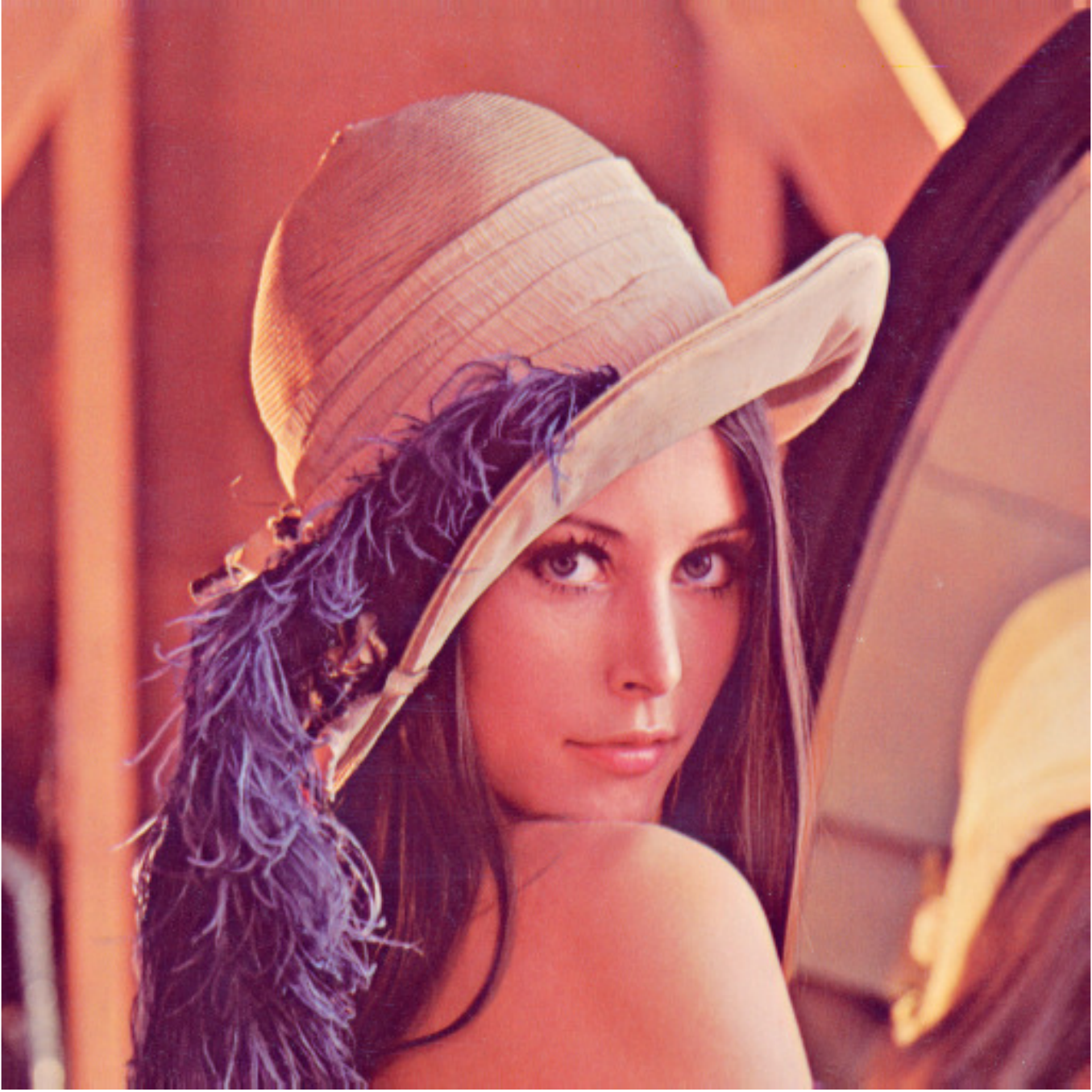}} \hspace{2mm}
\subfloat[\textbf{Proposed, $\textbf{274}$ms, $\textbf{55.36}$dB.}]{\includegraphics[width=0.2\linewidth]{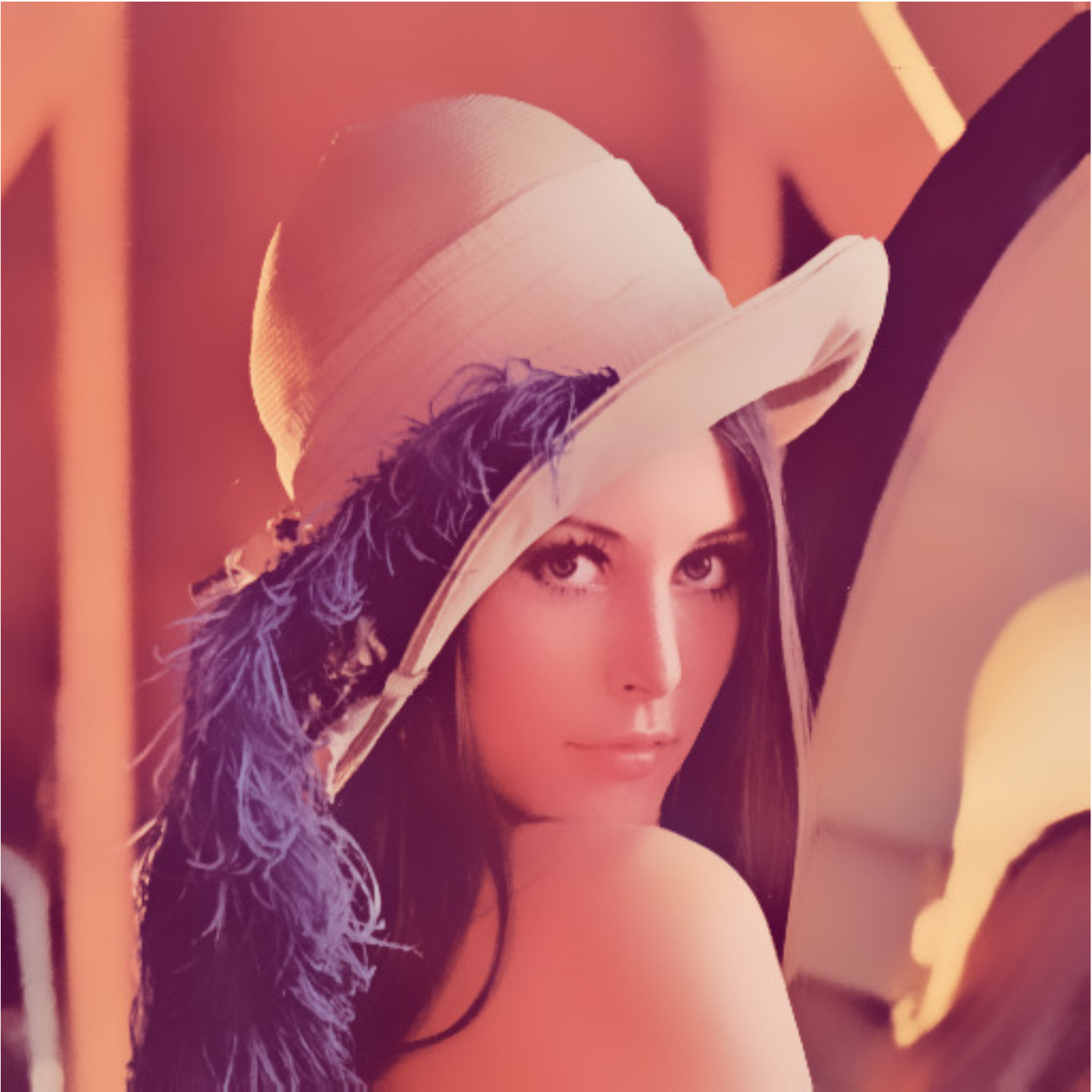}}\hspace{2mm}
\subfloat[GCS \cite{mozerov2015global}, $278$ms, $53.31$dB.]{\includegraphics[width=0.2\linewidth]{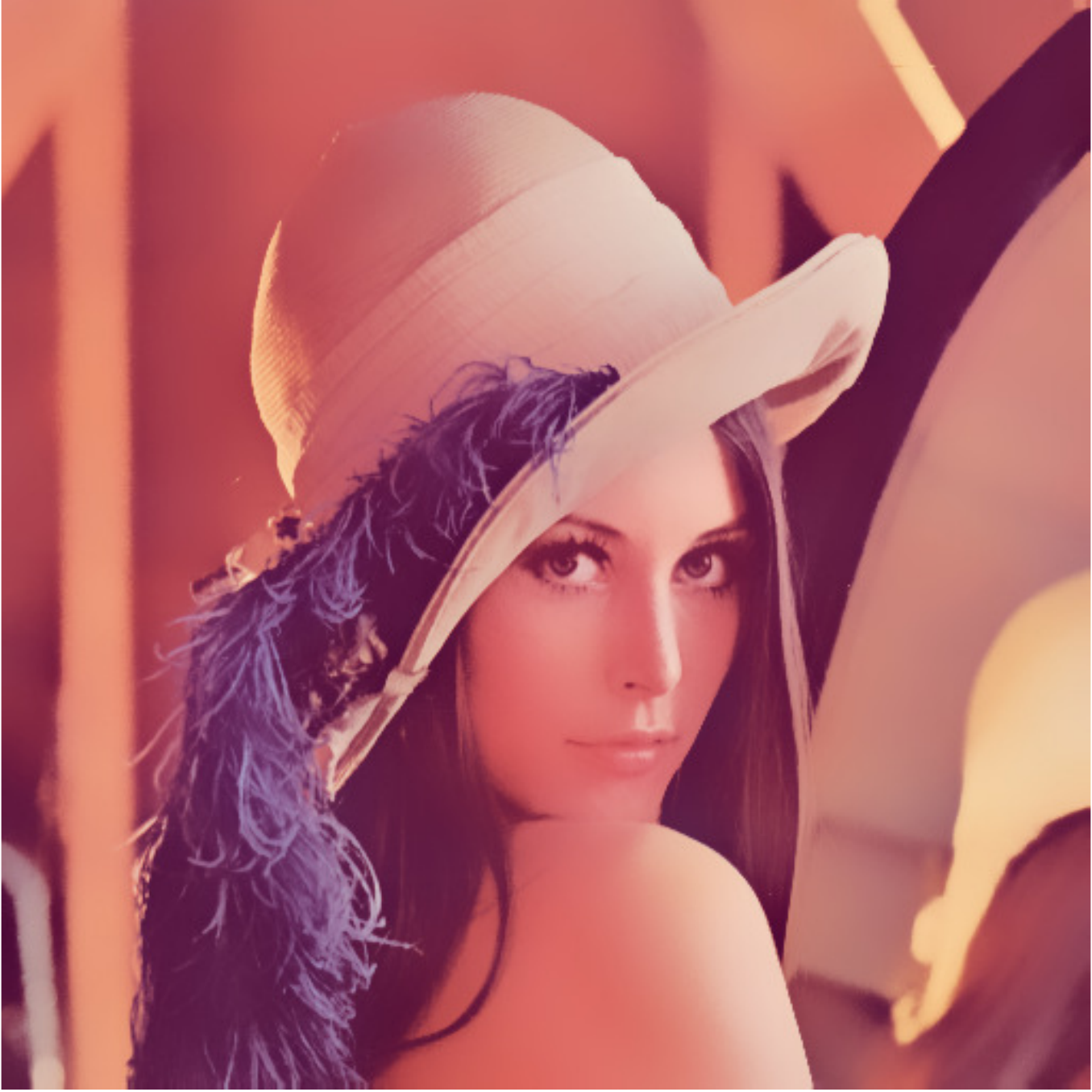}}\hspace{2mm}
\subfloat[AM \cite{gastal2012adaptive}, $272$ms, $40.58$dB.]{\includegraphics[width=0.2\linewidth]{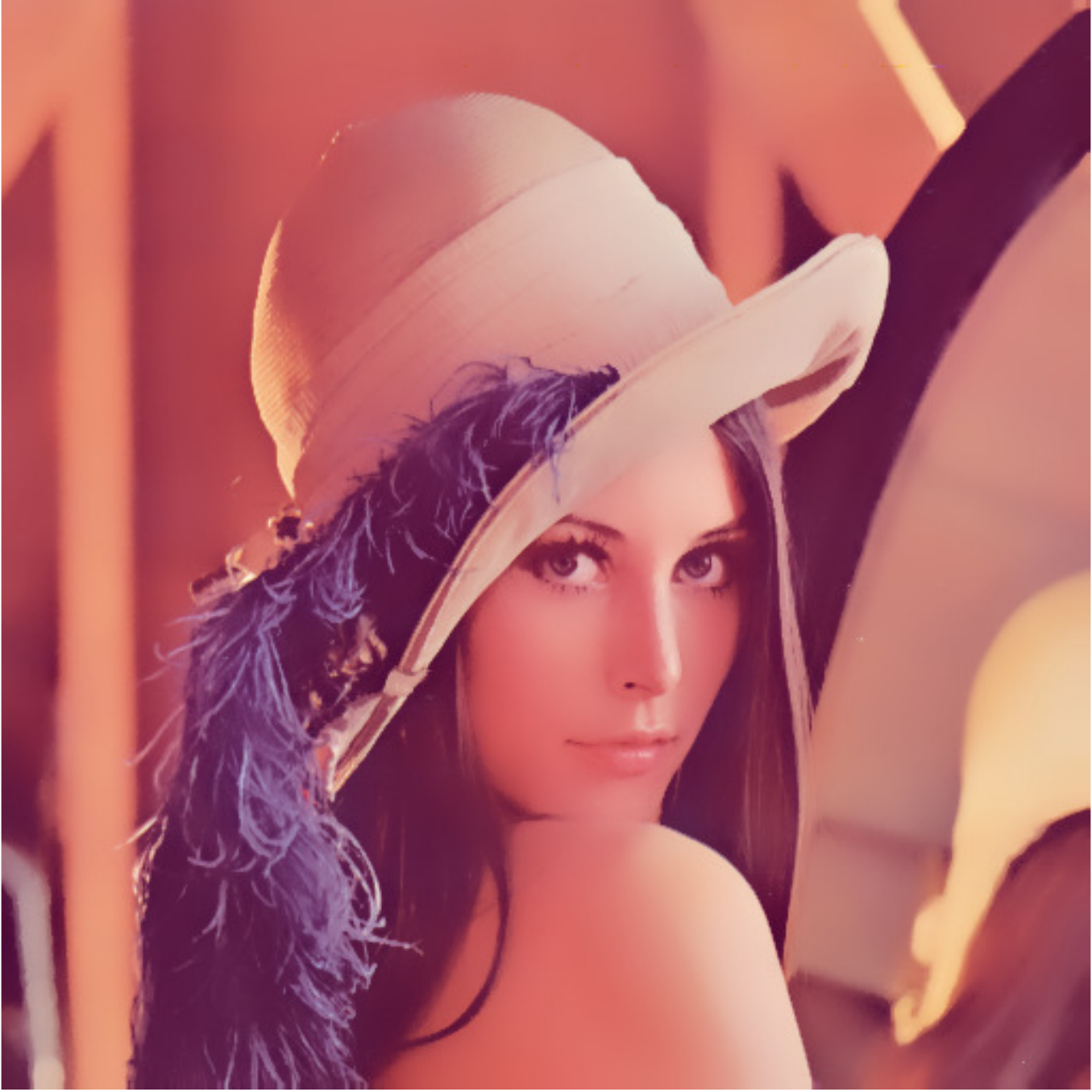}}

\subfloat[Brute-force, $19$ sec.]{\includegraphics[width=0.2\linewidth]{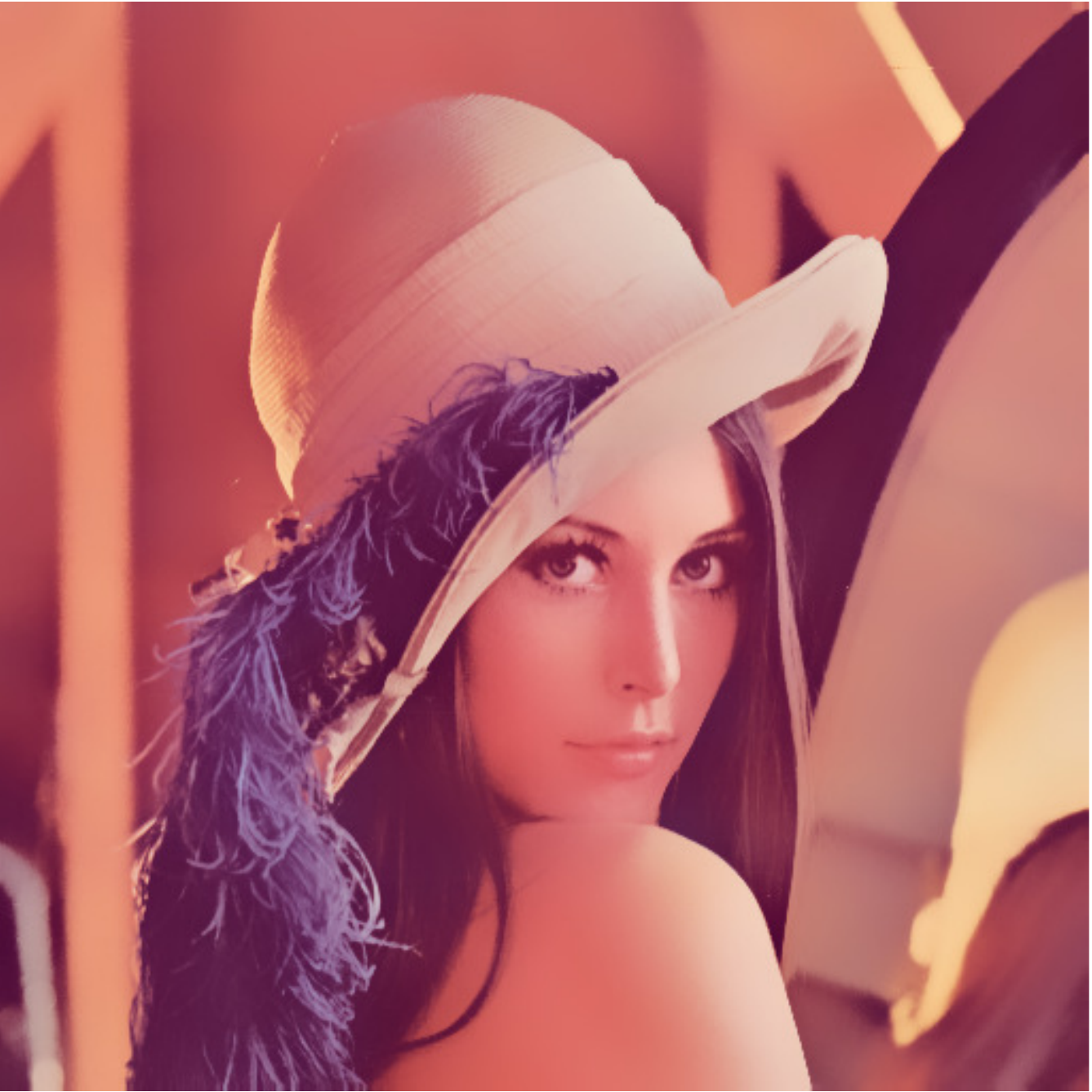}} \hspace{2mm}
\subfloat[Error: (b) - (e).]{\includegraphics[width=0.2\linewidth,height=0.2\linewidth]{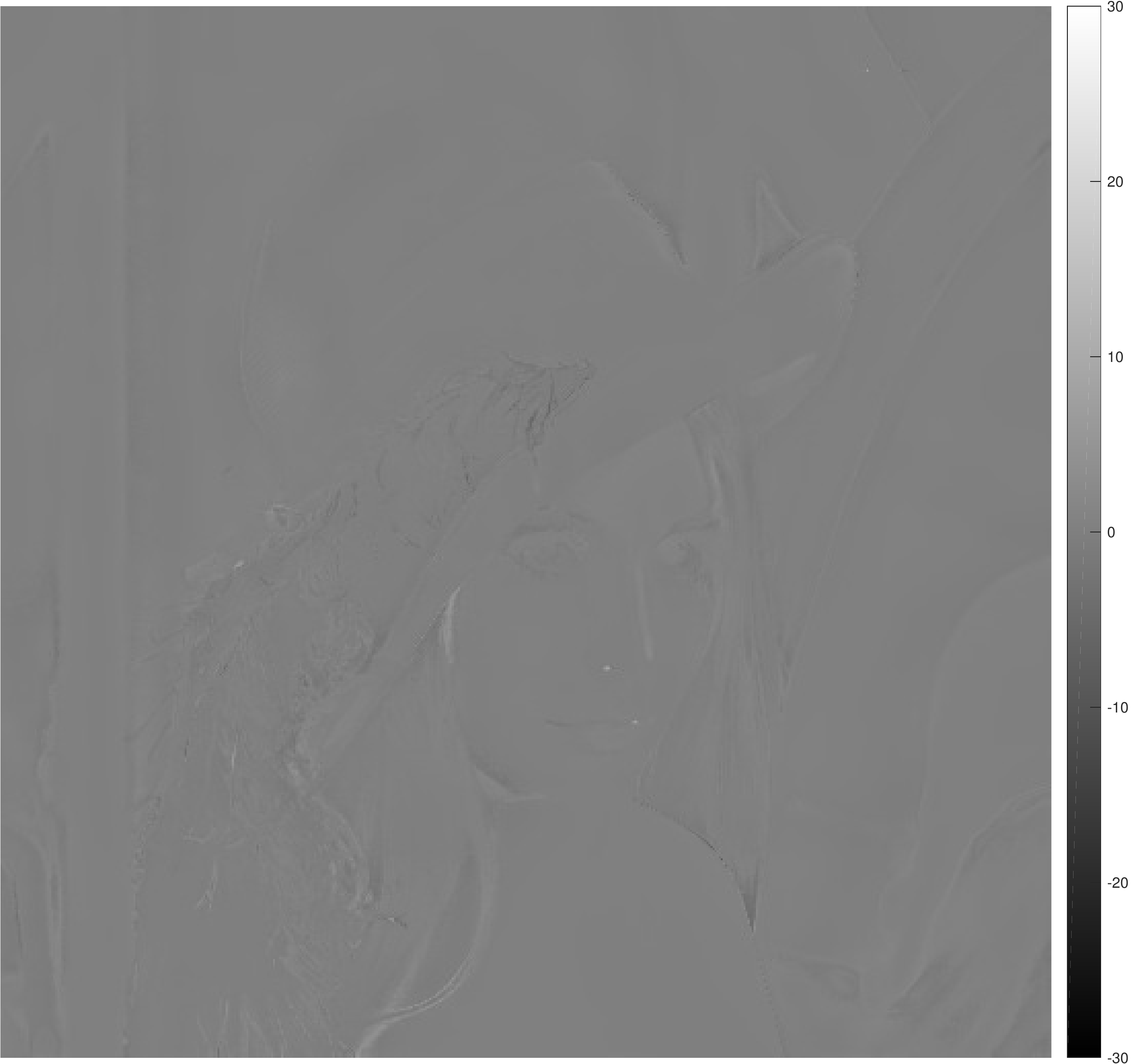}}\hspace{2mm}
\subfloat[Error: (c) - (e).]{\includegraphics[width=0.2\linewidth,height=0.2\linewidth]{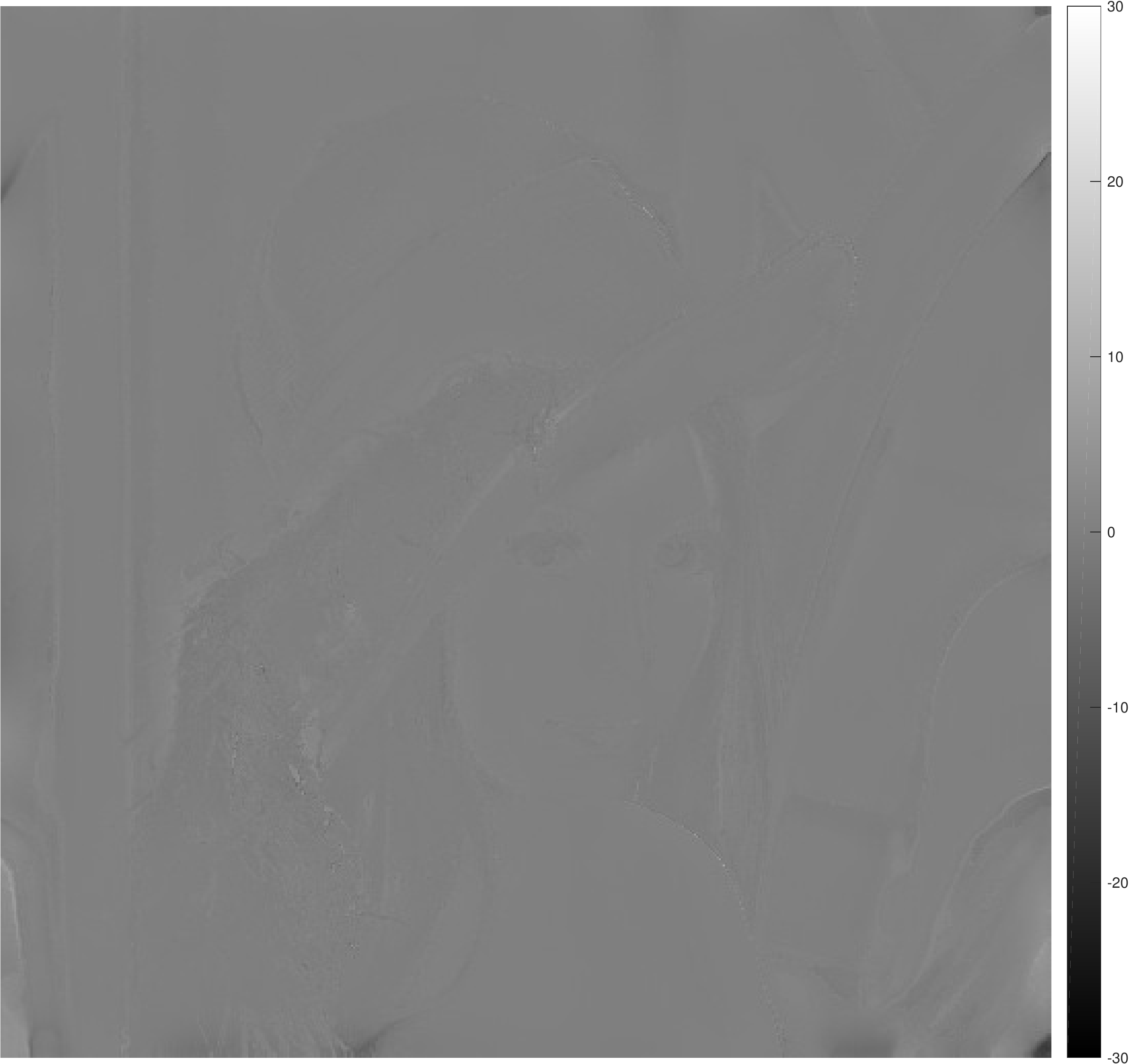}}\hspace{2mm}
\subfloat[Error: (d) - (e).]{\includegraphics[width=0.2\linewidth,height=0.2\linewidth]{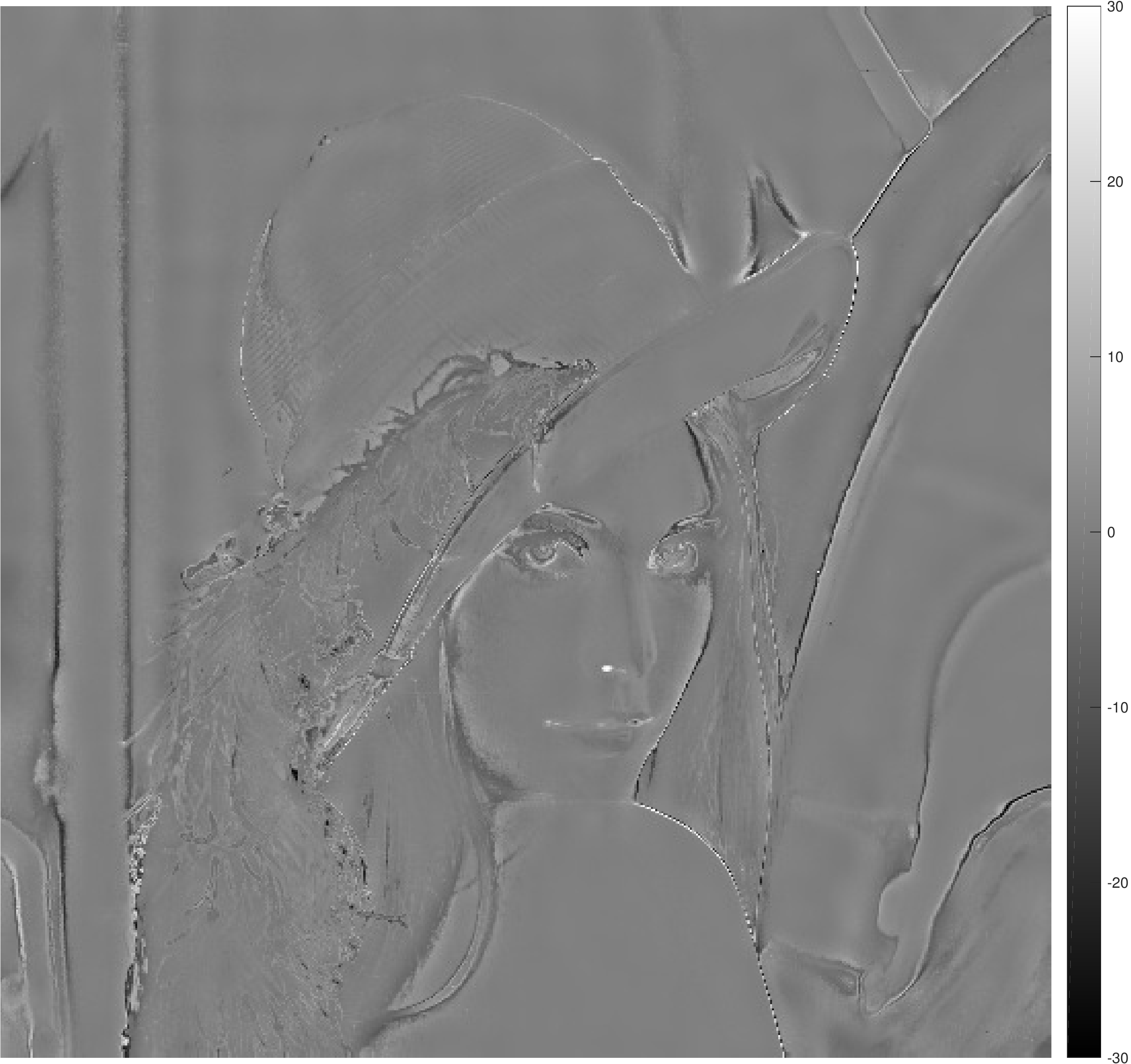}} 
\caption{Bilateral filtering of a color image \cite{paris2006fast}, where the filter parameters are $\sigma_s = 10$ and $\sigma_r = 40$. We have used $15$ clusters for GCS and the proposed method. The number of manifolds was automatically set to $3$ in AM. The run-time and $\mathrm{PSNR}$ (see definition in \eqref{rmse}) with $15$ manifolds are $1.48$ sec and $43.41$ dB. 
In the second row, we show the error between a particular method (first row) and the brute-force implementation for just the red channel.}
\label{Visualfig2}
\end{figure*} 
The aggregations in \eqref{num} and \eqref{den} are performed over a window around the pixel of interest, i.e., $W=[-S,S]^d$, where $S$ is the window length. 
We call $\omega: \mathbb{Z}^d \to \mathbb{R}$ the \textit{spatial}  kernel and $\varphi: \mathbb{R}^\rho \to \mathbb{R}$ the \textit{range} kernel \cite{tomasi1998bilateral}, where we 
%recall that $\rho$ is the dimension of the range space of $\p$.
denote the dimension of the range space of $\p$ as $\rho$. 
The spatial kernel is used to measure proximity in the spatial domain $\Omega$ (as in classical linear filtering). On the other hand, the range kernel is used to measure proximity in the range space of $\p$.  
%For some applications, $\p$ and $\f$ are identical. However, they are different for nonlocal means and infrared-guided low-light denoising (cf. Section \ref{Experiments}).

For (joint) bilateral filtering, $\omega$ and $\varphi$ are generally Gaussian. While $\f$ and $\p$ are identical for the bilateral filter ($\rho$=$n$), they are different for the joint bilateral filter \cite{paris2009bilateral}. 
In the original proposal for nonlocal means \cite{buades2005non}, $\p(\i)$ are the spatial neighbors of pixel $\i$ (extracted from a patch around $\i$) and hence $\rho$ is $n$ times the size of image patch, $\varphi$ is a multidimensional Gaussian, and $\omega$ is a box  filter. Later, it was shown in \cite{tasdizen2009principal} that by reducing the dimension of each patch (e.g., by applying PCA to the collection of patches), we can improve the speed and denoising performance. Similar to \cite{gastal2012adaptive,adams2010fast,adams2009gaussian}, we have also considered PCA-based nonlocal means in this work. Needless to say, the proposed algorithm can also work with full patches.  

The brute-force computation of \eqref{num} and \eqref{den} clearly requires $\mathcal{O}(S^d (n+\rho))$ operations per pixel.
In particular, the complexity scales exponentially with the window length $S$, which can be large for some smoothing applications.
In the last decade, several fast algorithms have been proposed for bilateral filtering of grayscale images \cite{paris2006fast,Chaudhury2011,durand2002fast,Porikli2008,Kamata2015,Yang2009}. 
The complexity can be cut down from $\mathcal{O}(S^d)$ to $\mathcal{O}(1)$ using these algorithms. 
Similarly, fast algorithms have been proposed for nonlocal means \cite{darbon2008,wang2006} that can reduce the complexity from $\mathcal{O}(S^d \rho)$ to $\mathcal{O}(S^d)$. 

\subsection{Previous Work}

\textit{Quantization methods}. Durand et al.~\cite{durand2002fast} proposed a novel framework for approximating \eqref{num} and \eqref{den} using clustering and interpolation. In terms of our notations, their approximation of \eqref{num} and \eqref{den} can be expressed as
\begin{equation}
 \label{numdurand}
\frac{1}{\hat{\eta}({\i})}  \sum_{k=1}^K c_k(\i) \Bigg( \sum_{\j\in W} \omega(\j)  \varphi\big(\p(\i-\j) - \amu_k\big) \f(\i-\j)\Bigg),
\end{equation}
and
\begin{equation}
\label{dendurand}
\hat{\eta}(\i) = \sum_{k=1}^K c_k(\i) \Bigg(\sum_{\j\in W} \omega(\j) \varphi\big(\p(\i-\j) - \amu_k\big)\Bigg).
\end{equation}
The centers $\amu_k$ are obtained by uniformly sampling the range space of $\p$, whereas the interpolation coefficient $c_k(\i)$ is determined from the distance between $\p(\i)$ and $\amu_k$. Notice that the inner summations in \eqref{numdurand} and \eqref{dendurand} can be expressed as convolutions, which are computed using FFT in~\cite{durand2002fast}. Furthermore, the entire processing is performed on the subsampled image, following which the output image is upsampled. The approximation turns out to be effective for grayscale images. This is because the range space is one-dimensional in this case, and hence a good approximation can be achieved using a small the number of samples $K$. Moreover, it suffices to interpolate just the two neighboring pixels. 
Paris et al.~\cite{paris2006fast} showed that the accuracy of~\cite{durand2002fast} can be improved by downsampling the intermediate images (instead of the input image) involved in the convolutions. Chen et al.~\cite{chen2007real} proposed to accelerate \cite{paris2006fast} by performing convolutions in the higher-dimensional 
spatio-range space. Yang et al.~\cite{Yang2009} observed that this framework can also be used with non-Gaussian range kernels, and that $\mathcal{O}(1)$ convolutions can be used to improve the computational complexity. 
As the range dimension $\rho$ increases, these methods however become computationally inefficient. In particular, an exponentially large $K$ is required to achieve a decent approximation, and linear interpolation requires $2^{\rho}$ neighboring samples. 
In~\cite{Yang2015}, Yang et al. extended this method to perform bilateral filtering of color images, and proposed a fast scheme for linear interpolation.
However, \cite{Yang2015} is based on uniform quantization, which is not optimal given the strong correlation between the color channels. 
In fact, a more efficient algorithm for color filtering was later proposed in~\cite{mozerov2015global}, where clustering is used to find $\amu_k$. Moreover, instead of interpolation, they used local statistics prior to improve the accuracy. However, the algorithm  is used only for color bilateral filtering and its application to other forms of high-dimensional filtering is not reported in~\cite{mozerov2015global}. 
\begin{figure*}
\centering
\includegraphics[width=0.9\linewidth]{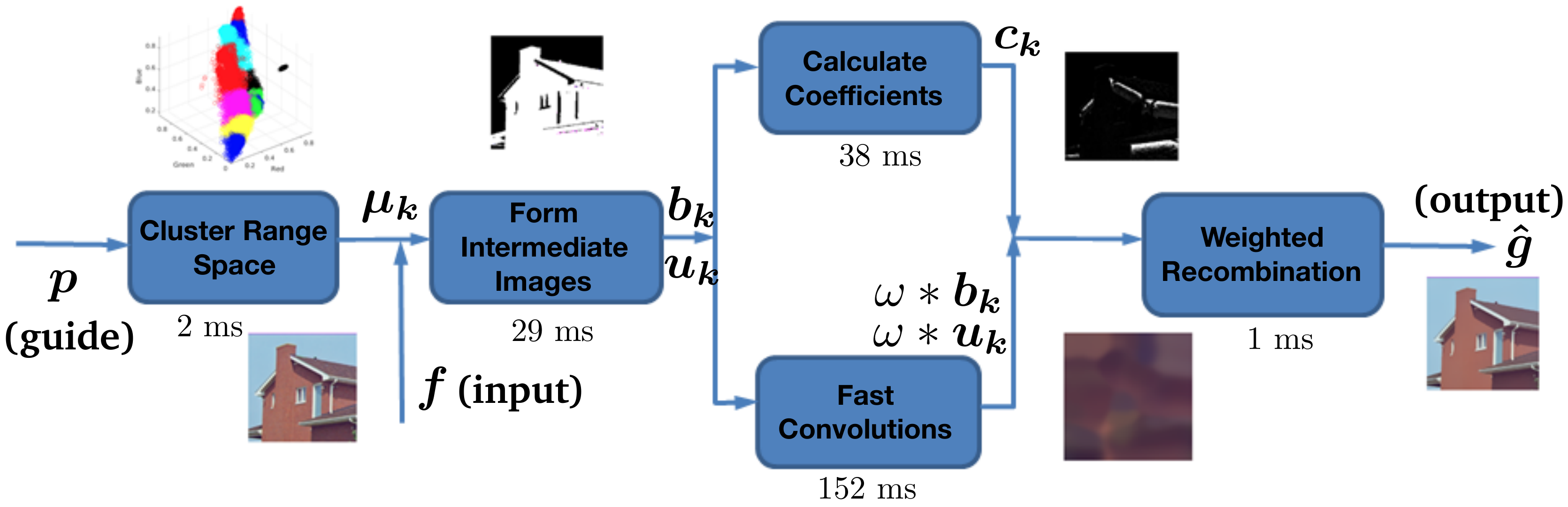}
\caption{Flowchart of our algorithm with the main modules. Also shown are the timings for the bilateral filtering of a $1$MP color image. The filter parameters are  $\sigma_s=10$ and $\sigma_r=75$. When $K=4$ (number of clusters), total time required is $253$ ms and the $\mathrm{PSNR}$ is $48.30$ dB.
The symbols used in the flowchart are defined in Section \ref{Proposed} (also see Algorithm \ref{fastalgo}).}
\label{flowchart}
\end{figure*}

\textit{Splat-blur-slice method}. Another line of work is \cite{gastal2012adaptive,adams2010fast,adams2009gaussian}, which is based on a slightly different form of approximation.
More precisely, they are based on the ``splat-blur-slice'' framework, which involves data partitioning (clustering or tessellation), fast convolutions, and interpolation as the core operations.  
These are considered to be the state-of-the-art fast algorithms for high-dimensional bilateral and nonlocal means filtering. 
The idea, originally proposed in~\cite{chen2007real}, is based on the observation that~\cite{paris2006fast} corresponds to convolutions in the joint spatio-range space. 
The general idea is to sample the input pixels in a different space (splatting), perform Gaussian convolutions (blurring), and resample the result back to the original pixel space (slicing). 
Adams et al. tessellated the domain and performed blurring using the $k$-d tree in \cite{adams2010fast} and the permutohedral lattice in \cite{adams2009gaussian}. 
Gastal et al.~\cite{gastal2012adaptive} divided and resampled the domain into non-linear manifolds, and performed blurring on them. 
This was shown to be faster than all other methods within the splat-blur-slice framework.

\textit{Kernel approximation}. In an altogether different direction, it was shown in \cite{Chaudhury2011,Porikli2008,Kamata2015,Chaudhury2016} that fast algorithms for bilateral filtering can be obtained by approximating the range kernel $\varphi$ using the so-called \textit{shiftable} functions, which includes polynomials and trigonometric functions. The above mentioned algorithms provide excellent accelerations for grayscale images, and are generally superior to algorithms based on data approximation. Unfortunately, it is difficult to approximate a high-dimensional Gaussian using low-order shiftable functions. For example, the Fourier expansion in \cite{Kamata2015} is quite effective in one dimension (grayscale images), 
but its straightforward extension to even three dimensions (color images) results in exponentially many terms.
This is referred to as the ``curse of dimensionality''.

\subsection{Contributions}

Existing algorithms have focussed on either approximating the data \cite{gastal2012adaptive,adams2010fast,adams2009gaussian,mozerov2015global,Yang2009} or the range kernel \cite{Chaudhury2011,Porikli2008,Kamata2015,ghosh2016fast,ghosh2018color}. 
In this paper, we combine the ideas of shiftability \cite{Chaudhury2011,Porikli2008,Chaudhury2016,Chaudhury2013} and data approximation \cite{durand2002fast,Yang2009} within a single framework to approximate \eqref{num} and \eqref{den}. 
In particular, we locally approximate the range kernel in \eqref{num} and \eqref{den} using weighted and shifted copies of a Gaussian, where the weights and shifts are determined from the data. 
More specifically, we use clustering to determine the shifts, whereby the correlation between data channels  is taken into account. Once the shifts have been fixed, 
we determine the weights (coefficients) using least-squares fitting, where the data is again taken into account. 
An important technical point is that we are required to solve a least-squares problem at each pixel, which can be expensive. We show how this can be done using just one matrix inversion. 
In summary, the key distinctions of our method in relation to previous approaches are as follows: 
\begin{itemize}
\item We use data-dependent methods to calculate both $\amu_k$ and $c_k$ in \eqref{numdurand} and \eqref{dendurand}. The latter is done in a heuristic fashion in \cite{gastal2012adaptive,adams2010fast,adams2009gaussian}. We note that our approximation also reduces to the form in \eqref{numdurand} and \eqref{dendurand}. However, the notion of shiftable approximation (in a sense which will be made precise in Section \ref{Proposed}) plays an important role. Namely, it allows us to interpret the coefficients in \eqref{numdurand} and \eqref{dendurand} from an approximation-theoretic point of view. This in turn forms the basis of the proposed optimization used to compute the coefficients.
\item Unlike~\cite{mozerov2015global,paris2006fast,durand2002fast,Yang2009,Yang2015}, our approach is scalable in the dimensions of the input $\f$ and the guide $\p$. 
\item An important difference with \cite{Chaudhury2011,Porikli2008,Kamata2015,Chaudhury2016} is that a different shiftable approximation is used at each pixel in our proposal, while the approximation is global in \cite{Chaudhury2011,Porikli2008,Kamata2015,Chaudhury2016}. This will be made precise in Section \ref{Proposed}.
\item Similar to  \cite{gastal2012adaptive,adams2010fast,adams2009gaussian}, our method also involves splatting (clustering), blurring (convolutions) and slicing (weighted recombinations). However, the important difference is that we perform slicing in an optimal manner,  
which is done in a heuristic fashion in these  methods. 
\end{itemize}

We explain the proposed shiftable approximation in detail in Section \ref{Proposed}. The algorithm emerging from the approximation is conceptually simple and easy to code. The flowchart of the processing pipeline is shown in Figure \ref{flowchart}.
The computation-intensive components are ``Cluster Range Space'' and ``Fast Convolutions'', for which highly optimized routines are available.  
Excluding clustering, the complexity is $\mathcal{O}(K(n+\rho))$, where $K$ is the number of clusters. Notice that there is no dependence on $S$. This is a big advantage compared to fast nonlocal means \cite{darbon2008,wang2006}, where the scaling is $\mathcal{O}(S^d)$.
The highlight of our proposal is that we are able to derive a bound on the approximation error for a particular variant of our algorithm \cite{pravin2017filtering}. In particular, the error is guaranteed to vanish if we use more clusters. 
This kind of guarantee is not offered by state-of-the-art algorithms for high-dimensional filtering \cite{gastal2012adaptive,adams2010fast,adams2009gaussian}, partly due to the complex nature of their formulation. 

We use the proposed fast algorithm for various applications such as smoothing \cite{tomasi1998bilateral}, denoising \cite{buades2005non}, low-light denoising \cite{zhuo2010enhancing}, hyperspectral filtering \cite{yuan2012hyperspectral}, and flow-field denoising \cite{westenberg2005denoising}. In particular, we demonstrate that  our algorithm is competitive with existing fast algorithms for color bilateral and nonlocal means  filtering  (cf. Figures \ref{Visualfig2} and \ref{Visualfig4}). We note that although our original target was high-dimensional filtering, our algorithm outperforms \cite{Yang2009}, which is considered as the state-of-the-art for bilateral filtering of grayscale images (cf. Figure \ref{Visualfig1}). 
Finally, we note that we have not used the particular form of the Gaussian kernel in the derivation of the fast algorithm. Therefore, it can also be used for non-Gaussian kernels, such as the exponential and the Lorentz kernel \cite{durand2002fast,yang2014hardware}.

\subsection{Organization}
This rest of the paper is organized as follows. 
In Section \ref{Proposed}, we describe the proposed approximation and the resulting fast algorithm. 
We also derive error bounds for a particular variant of the approximation.
In Section \ref{Experiments}, we apply the fast algorithm for high-dimensional bilateral and nonlocal means filtering, and compare its performance (timing and accuracy) with state-of-the-art algorithms. We conclude the paper with a summary of the results in Section \ref{Conclusion}.

\section{Proposed Method}
\label{Proposed}

Notice that, if the guide $\p$ has constant intensity value at each pixel, then \eqref{num} and \eqref{den} reduce to linear convolutions. 
This observation is essentially at the core of the fast algorithms in  \cite{paris2006fast,durand2002fast,Yang2009}. On the other hand, the fast algorithms in \cite{Chaudhury2011,Kamata2015,Chaudhury2016} are derived using a completely different idea, where the univariate Gaussian kernel $\varphi$ is approximated using a polynomial or a  trigonometric function $\psi$. 
These functions are \textit{shiftable} in the following sense: There exists basis functions $\psi_1,\ldots,\psi_K$ such that, for any $\tau \in \mathbb{R}$, we can write
\begin{equation}
\label{shiftable}
\psi(x - \tau) =  \sum_{k=1}^K c_{k}(\tau) \psi_k(x),
\end{equation}
where the coefficients $c_1,\ldots,c_K$ depend only on $\tau$. One can readily see that polynomials and trigonometric functions,
\begin{equation}
\label{polytrig}
\psi(x) = \sum_{n=0}^N p_n x^n \quad \text{and} \quad \psi(x) = \sum_{n=0}^N q_n\! \cos(n \omega_0 x),
\end{equation}
are shiftable. The utility of expansion \eqref{shiftable} is that it allows us to factor out $\p(\i)$ from the kernel. In particular, by replacing $\varphi$ by $\psi$, it was shown in  \cite{Chaudhury2011} that we can compute the bilateral filter using fast convolutions. The Taylor polynomial of $\varphi$ was used for $\psi$ in \cite{Porikli2008}, and the Fourier approximation of $\varphi$ was used in \cite{Chaudhury2011,Kamata2015,Chaudhury2016}. 

\subsection{Shiftable Approximation}

As mentioned earlier, it is rather difficult to obtain shiftable approximations for high-dimensional Gaussians that are practically useful. 
We can use separable extensions of \eqref{polytrig} to generate such approximations.
However, the difficulty is that the number of terms grows as $N^\rho$ in this case, where $N$ is the order in \eqref{polytrig}.
We address this fundamental problem using a data-centric approach. First, we do not use a shiftable (trigonometric or polynomial) approximation of $\varphi$ \cite{Chaudhury2011,Kamata2015}. Instead, for some fixed pixel $\i \in \Omega$, we consider the multidimensional Gaussian $\varphi(\x - \p(\i))$ centered at $\p(\i)$ appearing in \eqref{num} and \eqref{den}.  Similar to \eqref{shiftable}, we wish to find $\psi_1,\ldots,\psi_K$ such that
\begin{equation}
\label{shiftableHD}
\varphi(\x - \p(\i)) \approx \sum_{k=1}^K c_{k}(\i) \psi_k(\x),
\end{equation}
where $c_1,\ldots,c_K$ depend only on $\p(\i)$. The key distinction with \cite{Chaudhury2011,Porikli2008,Kamata2015,Chaudhury2016} is that the functional approximation is defined locally in \eqref{shiftableHD}, namely, the approximation is different at each pixel. Moreover, \eqref{shiftableHD} is neither a polynomial nor a trigonometric function. In spite of this, we continue to
use the term ``shiftable'' to highlight the fact that the approximation is based on ``shifts'' of the basis functions $\psi_1,\ldots,\psi_K$.
In this work, we propose to use the translates of the original kernel as the basis functions, i.e., we set $\psi_k(\x)=\varphi(\x - \amu_k)$. One can in principle use different basis functions, but we will not investigate this possibility in this paper.

The important consideration is that the shifts $\{\amu_k\}$ are set globally.
Since the range kernel is defined via the guide $\p$, we propose to cluster the range space of $\p$ and use the cluster centroids for $\{\amu_k\}$. That is we partition the range space
\begin{equation}
\label{range}
\Theta = \big\{\p(\i) : \i \in \Omega \big\},
\end{equation}
into clusters $\C_1, ....,\C_K$. We set $\amu_k$ to be the centroid of $\C_k$. 
In summary, for each $\i$, we require that
\begin{equation}
\label{approx}
\varphi(\x - \p(\i)) \approx \sum_{k=1}^K c_{k}(\i) \varphi(\x - \amu_k).
\end{equation}
At this point, notice the formal resemblance between \eqref{shiftable} and \eqref{approx}. Hence, we will refer to  \eqref{approx} as a shiftable approximation in the rest of the discussion, even though it is not shiftable in the sense of \eqref{shiftable}.

Note that we need a good approximation in  \eqref{approx} only for $\x \in \Theta_{\i}= \{\p(\i-\j): \j \in W\}$. This is because the samples appearing in \eqref{num} and \eqref{den} are $\varphi(\x - \p(\i))$, where $\x$ takes values in $\Theta_{\i}$. To determine the coefficients, it is therefore natural to consider the following problem:
\begin{equation}
\label{LS}
\underset{\c(\i) \in \mathbb{R}^K}{\min} \ \sum_{\x \in \Theta_{\i}} \Big(\varphi(\x - \p(\i)) - \sum_{k=1}^K c_{k}(\i) \varphi(\x - \amu_k)\Big)^2.
\end{equation}
\begin{figure*}
\centering
\subfloat{\includegraphics[width=0.4\linewidth]{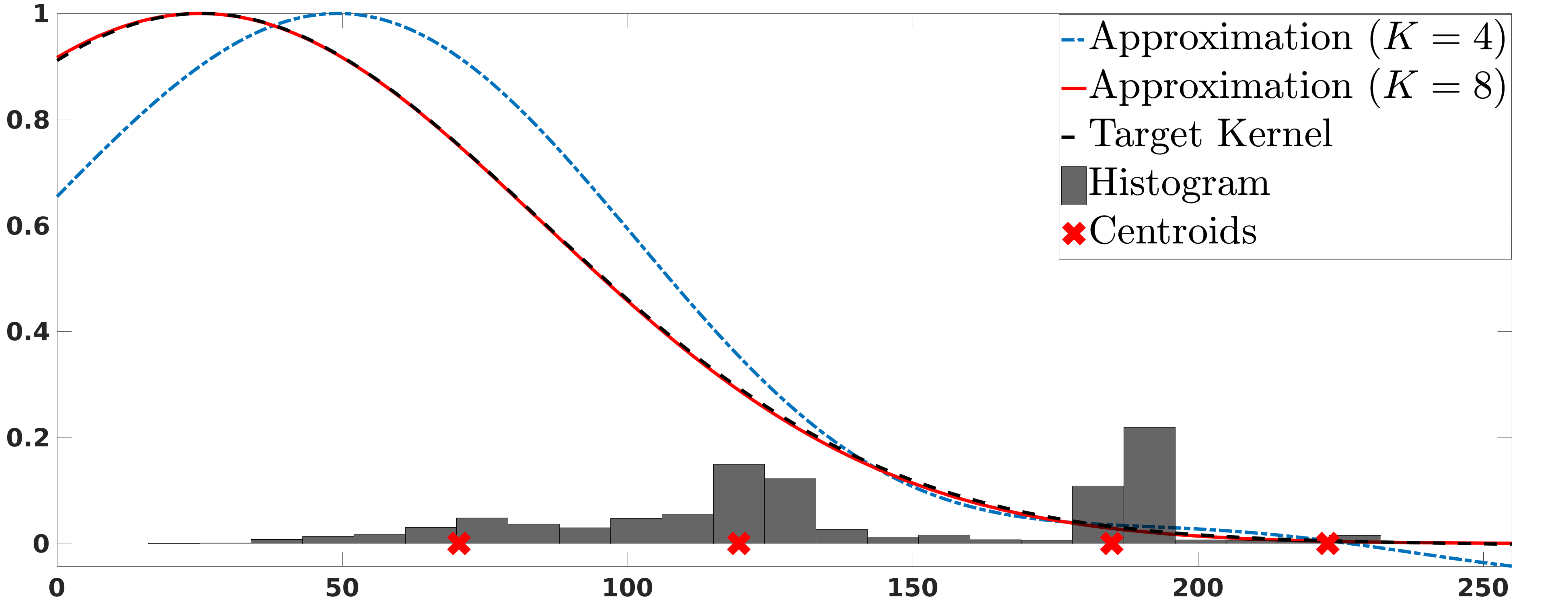}} \hspace{3.5mm}
\subfloat{\includegraphics[width=0.4\linewidth]{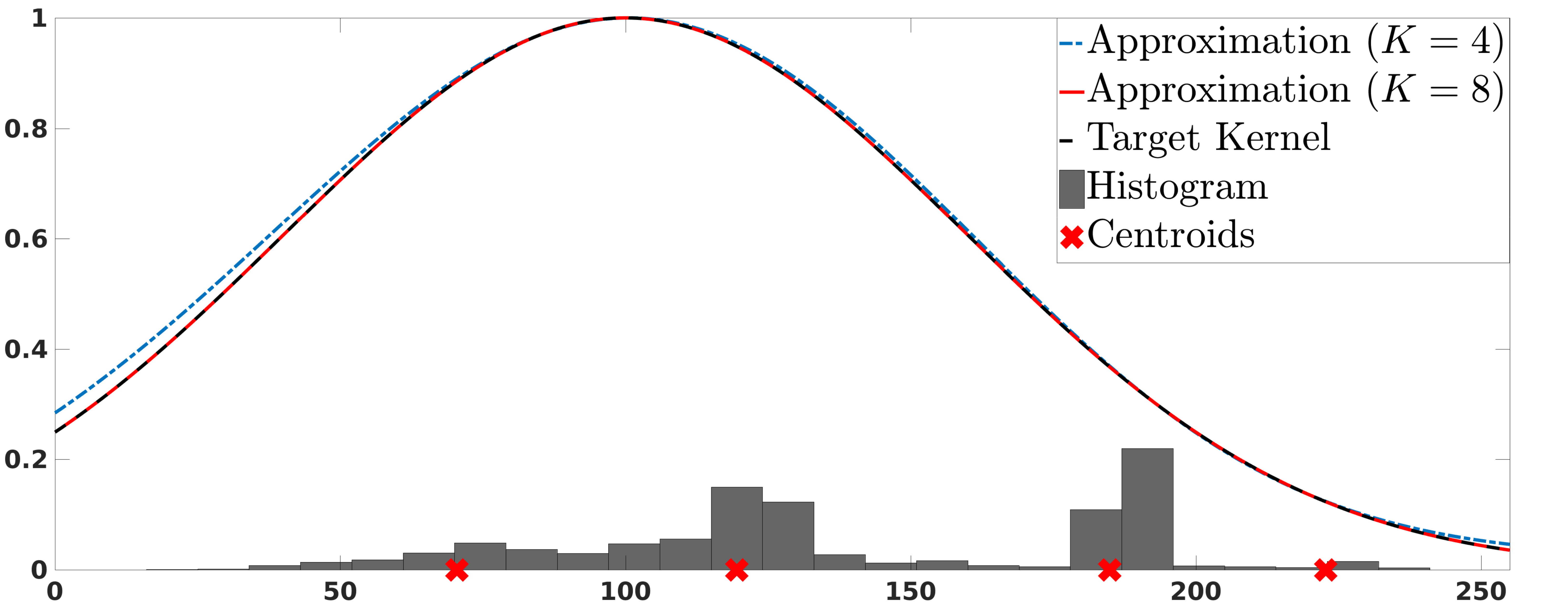}} 
\caption{Proposed approximation of the Gaussian range kernel $\varphi(x-\p(\i))$ in \eqref{approx} for $\p(\i)=25$ (left) and $\p(\i)=100$ (right). The shifts $\{\mu_k\}$ are obtained by clustering a grayscale image whose histogram is shown above. As expected, the approximation is better for order $K=8$ compared to $K=4$. In particular, the approximation  is quite accurate over the entire dynamic 
range $[0,255]$ when $K=8$. However, notice that for $K=4$, the approximation is better in the interval $[100,200]$, where the density is higher. As the approximation is data-driven, the error gets distributed in tune with the underlying histogram.}
\label{fig:approx}
\end{figure*}
\begin{figure}
\centering
\includegraphics[width=0.55\linewidth,height=0.3\linewidth]{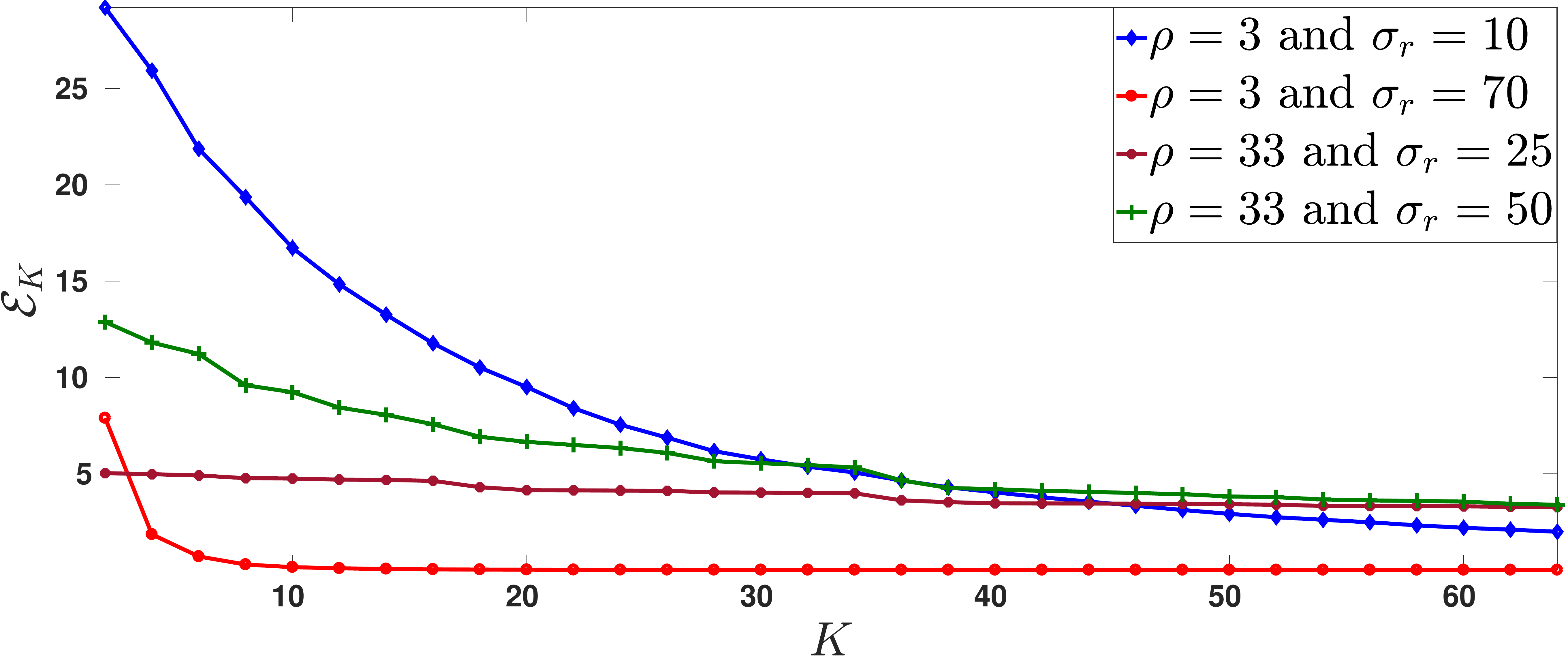} \hspace{1mm}
\caption{Error rates for the proposed approximation (see text for the definition of $\mathcal{E}_K$), where $\rho$ is the dimension  of the range space of the guide image $\p$. We notice that the approximation error falls off with the increase in order $K$. 
The error is averaged over the color images ($\rho=3$) from the Kodak dataset and hyperspectral images ($\rho=33$) from~\cite{hyperspectraldata}.}
\label{Approxerrorfig}
\label{fig1:approx}
\end{figure}
The difficulty is that this requires us to solve an expensive least-squares problem (the matrix to be inverted is large) for each $\i$. 
More importantly, we have a different inversion, at each pixel. This is time consuming and impractical. 
On the other hand, notice that $\Theta_i$ is included  in $\Theta$. Hence, we could perform the fitting over $\Theta$. Instead, we set $\Lambda =\{ \amu_1,\ldots,\amu_K\}$ and choose to fit over $\Lambda$, which are quantized representatives of the points in $\Theta$. 
In short, while the cluster centers $\Lambda$ most likely do not belong to $\Theta_i$, they are representative of the larger set $\Theta$.
Therefore, instead of \eqref{LS}, we consider the surrogate problem:
\begin{equation*}
\underset{\c(\i) \in \mathbb{R}^K}{\min} \ \sum_{\x \in \Lambda} \Big(\varphi(\x - \p(\i)) - \sum_{k=1}^K c_{k}(\i) \varphi(\x - \amu_k)\Big)^2.
\end{equation*}
In matrix notation, we can compactly write this as
\begin{equation}
 \label{opt}
\underset{\c(\i) \in \mathbb{R}^K}{\min} \quad \lVert \mathbf{A}\c(\i) - \b(\i) \rVert^2,
\end{equation}
where $\mathbf{A} \in \mathbb{R}^{K \times K}$ and $\b(\i) \in \mathbb{R}^K$ are given by
\begin{equation}
\label{defAb}
A_{kl} = \varphi(\amu_k - \amu_l) \quad \text{and} \quad b_k(\i) = \varphi(\amu_k - \p(\i)).
\end{equation}
The solution of \eqref{opt} is  $\c(\i)=\mathbf{A}^{\dagger} \b(\i)$, where $\mathbf{A}^{\dagger}$ is the pseudo-inverse of $\mathbf{A}$. 
In particular, we need to compute the pseudo-inverse just once; the coefficients at each pixel are obtained using a matrix-vector multiplication. 
In Figure \ref{Approxfig1}, we show that solving \eqref{opt} is much faster than solving \eqref{LS}, and that the coefficients from \eqref{opt} are close to those from \eqref{LS}.

The approximation result for univariate Gaussians are shown in Figure \ref{fig:approx}. Notice that our approximation depends on the distribution of $\p$ (because of the clustering), and the coefficients are obtained by solving a least square problem at each pixel $\i$. 
Hence, in Figure \ref{fig:approx}, a particular distribution is selected and the approximations for $\p(\i)=25$ and $\p(\i)=100$ are shown.
The error rates (error vs. $K$) for multidimensional Gaussians (covariance $\sigma_r^2 \mathbf{I}$) corresponding to dimensions $\rho=3$ and $33$ are shown in Figure \ref{Approxerrorfig}. The error $\mathcal{E}_K$ for a fixed $K$ is simply \eqref{LS} averaged over all the pixels.

To summarize, the steps in the proposed shiftable approximation are as follows:
\begin{itemize}
\item Cluster the range space $\Theta$ and use the centers for $\amu_k$. 
\item Set the basis functions as $\psi_k(\x)=\varphi(\x - \amu_k)$. 
\item Set up $\mathbf{A}$ using \eqref{defAb}  and compute its pseudo-inverse $\mathbf{A}^{\dagger}$. 
\item At each pixel $\i$, set $\b(\i)$ using \eqref{defAb} and $\c(\i)=\mathbf{A}^{\dagger} \b(\i)$.
\end{itemize}
We note that one can freely choose different shifts (and basis functions)  in \eqref{approx}  for different applications. 
That is, \eqref{approx} offers a broad approximation framework, where one can consider other rules for fixing the parameters (shifts and coefficients).

\subsection{Fast Algorithm}
\label{Algo}

We now develop a fast algorithm by replacing the kernel in \eqref{num} and \eqref{den} with the approximation in \eqref{approx}. 
For $1 \leq k \leq K$, define $\boldsymbol{v}_k: \Omega  \to \mathbb{R}^n$ and $r_k: \Omega  \to \mathbb{R}$ to be
\begin{equation}
\label{temp1}
\boldsymbol{v}_k(\i) = \smashoperator[r]{\sum_{\j\in W}} \omega(\j) \ \varphi( \p(\i-\j) - \amu_k) \f(\i-\j),
\end{equation}
\begin{equation}
\label{temp2}
\text{and} \ \ \ r_k(\i)={\smashoperator[r]{\sum_{\j\in W}} \omega(\j) \  \varphi(\p(\i-\j) - \amu_k)}.
\end{equation}
For $\x=\p(\i-\j)$, $\j \in W$, we replace $\varphi(\x-\p(\i))$ in \eqref{num} and \eqref{den} with the approximation in \eqref{approx}. 
After some manipulations (see Appendix \ref{derivation}), \eqref{num} and \eqref{den} are approximated as,
\begin{equation}
\label{HDapprox}
\hat{\g}(\i) =  \frac{1}{\hat{\eta}(\i)} \sum_{k=1}^K  c_{k}(\i) \boldsymbol{v}_k(\i),
\end{equation}
\begin{equation}
\label{approxDen}
\text{and} \ \ \ \hat{\eta}(\i) =  \sum_{k=1}^K  c_{k}(\i) r_k(\i).
\end{equation}

The advantage with \eqref{HDapprox} and  \eqref{approxDen} is clear once we notice that \eqref{temp1} and \eqref{temp2} can be expressed as convolutions.
Indeed, by defining $\boldsymbol{u}_k: \Omega  \to \mathbb{R}^n$ to be $\boldsymbol{u}_k(\i) = b_k (\i) \f(\i)$, where $b_k $ is given by \eqref{defAb}, we can write
\begin{equation*}
\boldsymbol{v}_k(\i) = (\omega \ast \boldsymbol{u}_k ) (\i)= \sum_{\j\in W} \omega(\j) \boldsymbol{u}_k (\i-\j),
\end{equation*}
\begin{equation*}
\text{and} \ \ \ r_k(\i) = (\omega \ast b_k) (\i)= \sum_{\j\in W} \omega(\j) b_k (\i-\j),
\end{equation*}
\begin{algorithm}
 \textbf{Input}: $\f : \Omega \to \mathbb{R}^n$ and $\p : \Omega \to \mathbb{R}^\rho$, kernels $\omega$ and $\varphi$\;
\textbf{Parameter}: Number of clusters $K$\;
\textbf{Output}: Approximation in \eqref{HDapprox}\;  
\tcp{CLUSTER RANGE SPACE}
$\{\amu_k\} \leftarrow$ cluster $\Theta$ in \eqref{range} using bisecting $K$-means\; \label{cluster}
\tcp{FORM INTERMEDIATE IMAGES}
Set up $\mathbf{A}$ in \eqref{defAb} using $\{\amu_k\}$ and $\varphi$, and compute $\mathbf{A}^{\dagger}$\;  \label{pinv}
\For{$\i \in \Omega$} {  \label{for1}
\For{$k=1,\ldots,K$}{
$b_k(\i) \leftarrow \varphi(\amu_k - \p(\i))$\;
$\boldsymbol{u}_k(\i) \leftarrow b_k (\i) \f(\i)$\; \label{inter}
}
}
\tcp{CALCULATE COEFFICIENTS}
\For{$\i \in \Omega$}{  \label{for2}
$\c(\i) \leftarrow \mathbf{A}^{\dagger}\b(\i)$\; \label{inv}
}
\tcp{FAST CONVOLUTIONS}
Initialize $\boldsymbol{v}: \Omega \to \mathbb{R}^n$ and $r: \Omega \to \mathbb{R}$ with zeros\;
\For{$k=1,\ldots,K$}{  \label{for3}
$\boldsymbol{v}\leftarrow \boldsymbol{v}\oplus \left[c_k \otimes (\omega \ast \boldsymbol{u}_k)\right]$\; \label{conv1}
$r \leftarrow r \oplus \left[c_k \otimes (\omega \ast b_k)\right]$\; \label{conv2}
}
$\hat{\g} \leftarrow \boldsymbol{v} \oslash r$.
\caption{Fast High-Dimensional Filtering.}
\label{fastalgo}  
\end{algorithm}
where $\ast$ denotes linear convolution. In summary, using \eqref{approx}, we have been able to approximate the high-dimensional filtering using weighted combinations of fast convolutions.
The overall process is summarized in Algorithm \ref{fastalgo}. 
Note that we have used $\oplus$, $\otimes$ and $\oslash$ to represent pointwise addition, multiplication, and division.
In steps \ref{conv1} and \ref{conv2}, $c_k$  denotes the mapping $c_k : \Omega \to \mathbb{R}$.
The dominant computation in Algorithm \ref{fastalgo} are the $(n+1)K$ convolutions in steps \ref{conv1} and \ref{conv2}. 
Clustering and inversion of $\mathbf{A}$ is performed just once. Notice that $b_k$, which is used for computing the coefficients in step \ref{inv}, is anyways required to form the intermediate images in step \ref{inter}.
The flowchart of our algorithm, along with typical timings for a megapixel image, is shown in Figure \ref{flowchart}. 
Note that the steps in Figure \ref{flowchart} (resp. loops in Algorithm \ref{fastalgo}) can be parallelized: clustering the range space (loop \ref{for1}), finding coefficients (loop \ref{for2}), and performing convolutions (loop \ref{for3}).

\subsection{Implementation}

We have used bisecting $K$-means \cite{Tan2005} for iteratively clustering the range space in step \ref{cluster}. In particular, the cluster with largest variance is divided into two parts (using $2$-means clustering) at every iteration. This ensures that we do not end up with few large clusters at the end. 
We initialize the $2$-means clustering using a pair of points with the largest separation. We picked bisecting $K$-means over other clustering algorithms as its cost is linear in $K$, and is generally faster than other clustering algorithms \cite{Tan2005}. Importantly, its computational overhead is negligible compared to the time required for the convolutions. Needless to say, we can use any efficient clustering method and the fast algorithm would still go through. 

We have used the Matlab routine \textit{pinv} to compute $\mathbf{A}^{\dagger}$ in step \ref{pinv}. 
As is well-known, when $\omega$ is a box or Gaussian, we can convolve using $\mathcal{O}(1)$ operations, i.e., the cost does not depend on the size of spatial filter \cite{deriche1993recursively,young1995recursive}. 
The Gaussian convolutions in steps \ref{conv1} and \ref{conv2} are implemented using Young's $\mathcal{O}(1)$ algorithm \cite{young1995recursive}.
For box convolutions, we used a standard recursive procedure which require just four operations per pixel for any box size.
Since the leading computations are the convolutions, and there are $\mathcal{O}(Kn)$ convolutions, the overall complexity  is $\mathcal{O}(K(n+\rho))$; the additional $\mathcal{O}(K\rho)$ term is for computing \eqref{defAb}.
Importantly, we are able to obtain a speedup of $S^d/K$ over the brute-force computations of \eqref{num} and \eqref{den}. 
The storage complexity of Algorithm \ref{fastalgo} is clearly linear in $n$, $K,$ as well as the image size. The MATLAB implementation of our algorithm is available here: https://github.com/pravin1390/FastHDFilter.

\subsection{Approximation Accuracy}

We note that the algorithm in \cite{pravin2017filtering} is a variant of our method.
Indeed, for $1 \leq k \leq K$, if we set the coefficients in \eqref{approx} to be
\begin{equation}
\label{coeffICIP}
c_k(\i) =
\begin{cases}
   1 & \text{if $\p(\i) \in {\C}_k$,} \\
   0 & \text{else},
\end{cases}
\end{equation}
then we obtain the fast algorithm in \cite{pravin2017filtering}. In other words, the filtering is performed on a cluster-by-cluster basis in this case. 
Correspondingly, \eqref{HDapprox} reduces to
\begin{equation}
\label{HDapprox2}
\hat{\g}(\i) = \frac{\boldsymbol{v}_s(\i)}{r_s(\i)},
\end{equation}
assuming that $\p(\i)$ is in cluster ${\C}_s$.

In  \eqref{HDapprox} and  \eqref{approxDen}, we weight the results from the $K$ clusters, where the weights are obtained using the optimization in \eqref{opt}. 
We will demonstrate in Section \ref{Experiments} that the weighting helps in reducing the approximation error. Unfortunately, it also makes the analysis of the algorithm difficult. Intuitively though, one would expect the error from \eqref{HDapprox} to be less than that from 
 \eqref{HDapprox2}. For the latter approximation, we have the following result \cite{pravin2017filtering}.
\begin{theorem}[Error bound]
\label{theorem} Suppose that $\omega$ and $\varphi$ are non-negative, and 
that the latter is Lipschitz continuous, i.e., for some $L >0$, $\lvert \varphi(\x)  - \varphi(\y) \rvert \leq L \lVert \x - \y \rVert$, 
for any $\x$ and $\y$ in $\mathbb{R}^{\rho}$. 
Then, for some constant $C >0$,
\begin{equation}
 \label{bound}
\sum_{\i \in \Omega} \|\hat{\g}(\i)-\g(\i)\|^2 \leq C L^2 n |W|^2  {{E}}_K,
\end{equation}
where $\hat{\g}$ is given by \eqref{HDapprox2}, and $E_K$ is the clustering error:
\begin{equation}
{E}_K = \sum_{k=1}^K  \sum_{\p(\i) \in \C_k} \!\! {\|\p(\i) - \amu_k \|}^2.
\end{equation}
\end{theorem}
We note that the assumptions in Theorem \ref{theorem} are valid when $\omega$ is a box or Gaussian, and $\varphi$ is a multidimensional Gaussian. For completeness, we provide the derivation of Theorem \ref{theorem}  in Appendix \ref{thoremproof}. For several clustering methods, the clustering error $E_K$ vanishes as $K \rightarrow \infty$ \cite{Tan2005}. For any of these methods, we can approximate \eqref{num} with arbitrary accuracy provided $K$ is sufficiently large. We note that such a guarantee is not available for existing fast  
approximation algorithms for high-dimensional filtering \cite{gastal2012adaptive,adams2010fast,adams2009gaussian}. 

\section{Numerical Results}
\label{Experiments}
We now report some numerical results for high-dimensional filtering. We have used the Matlab implementation of Algorithm \ref{fastalgo}.
For fair timing comparisons with Adaptive Manifolds (AM) \cite{gastal2012adaptive} and Global Color Sparseness (GCS) \cite{mozerov2015global}, we have used the Matlab code provided by the authors. 
The timing analysis was performed on an Intel 4-core 3.4 GHz machine with 32 GB memory.
The grayscale and color images used for our experiments were obtained from standard databases\footnote{https://goo.gl/821N2G, https://goo.gl/2fcNmu, https://goo.gl/MvxCMX.}. The infrared and hyperspectral images are the ones used in \cite{gastal2012adaptive} and \cite{hyperspectraldata,ImageSource4}. 
To compare the filtering accuracy with existing methods, we have fixed the timings by adjusting the respective approximation order. 
The objective of this section is to demonstrate that, in spite of its simple formulation, our algorithm is competitive with existing fast approximations of bilateral and nonlocal means filtering.

We have used an isotropic Gaussian range kernel for bilateral and nonlocal means filtering :
\begin{equation*}
\varphi( \x) = \exp\left(-\lVert \x \rVert^2/2\sigma_r^2 \right),
\end{equation*}
where the standard deviation $\sigma_r$ is used to control the effective width (influence) of the kernel. We recall from \eqref{num} and \eqref{den} that the input to $\varphi$ is the difference $\x=\p(\i)-\p(\i-\j)$.
In one of the experiments, $\varphi( \x)$ is an anisotropic Gaussian; we have explicitly mentioned this at the right place. 
The spatial kernel $\omega$ for bilateral filtering is also Gaussian:
\begin{equation*}
\omega(\i) = \exp\left(-\lVert \i \rVert^2/2\sigma_s^2 \right),
\end{equation*}
where $\sigma_s$ is again the standard deviation.
For nonlocal means (NLM), $\omega$ is a box filter, namely, no spatial weighting is used. The filter width for bilateral filtering is $S=3\sigma_s$, while that for NLM is explicitly mentioned (typically, $S=10$).

Following \cite{Porikli2008,Yang2009}, the error between the outputs of the brute-force implementation and the fast algorithm is measured using the peak signal-to-noise ratio (PSNR). This is given by $\mathrm{PSNR}= - 10 \log_{10}(\mathrm{MSE})$, where
\begin{equation}
\label{rmse}
\mathrm{MSE}= |\Omega|^{-1} \sum_{\i  \in \Omega}\|\hat{\g}(\i)-\g(\i)\|^2,
\end{equation}
and $|\Omega|$ is the number of pixels. Note that $\mathrm{PSNR}$ is also used to measure denoising performance;  in this case, $\hat{\g}(\i)$ and $\g(\i)$ are the denoised and clean images  in \eqref{rmse}. 
It should be evident from the context as to what is being measured using $\mathrm{PSNR}$.
We also use $\mathrm{SSIM}$ \cite{wang2004} for measuring visual quality. 

\subsection{Grayscale Bilateral Filtering}

\begin{figure}[!htp]
\centering
\subfloat[Input ($256 \times 256$).]{\includegraphics[width=0.34\linewidth]{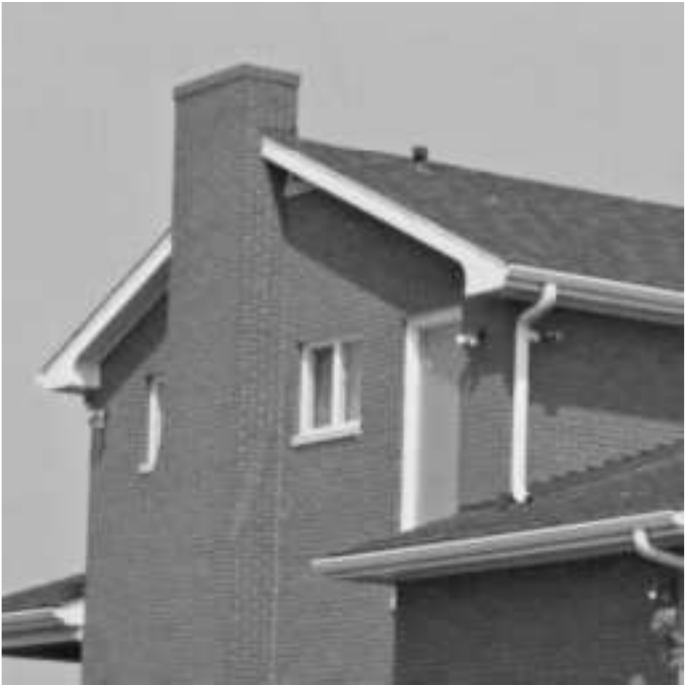}} \hspace{2mm}
\subfloat[Brute-force filtering.]{\includegraphics[width=0.34\linewidth]{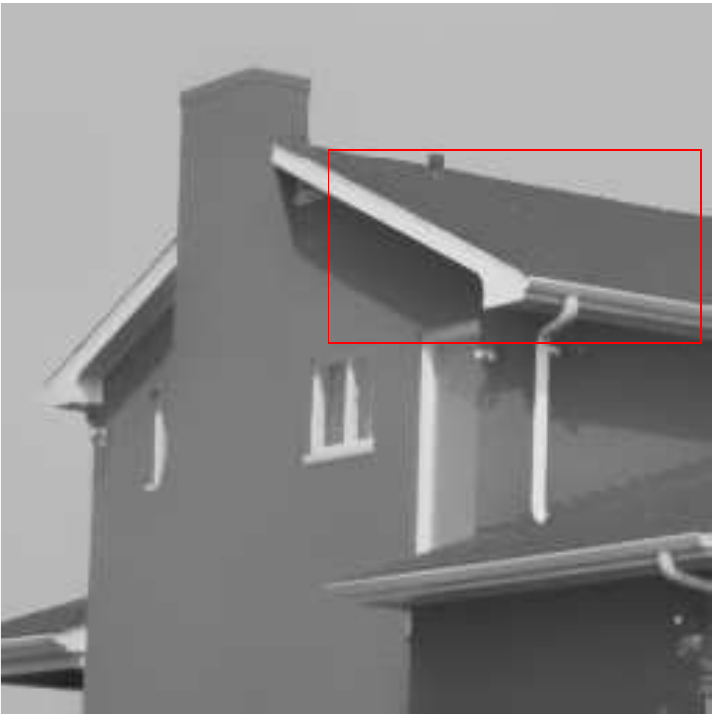}} 

\subfloat[\textbf{Proposed ($\textbf{43.5}$ dB).}]{\includegraphics[width=0.34\linewidth]{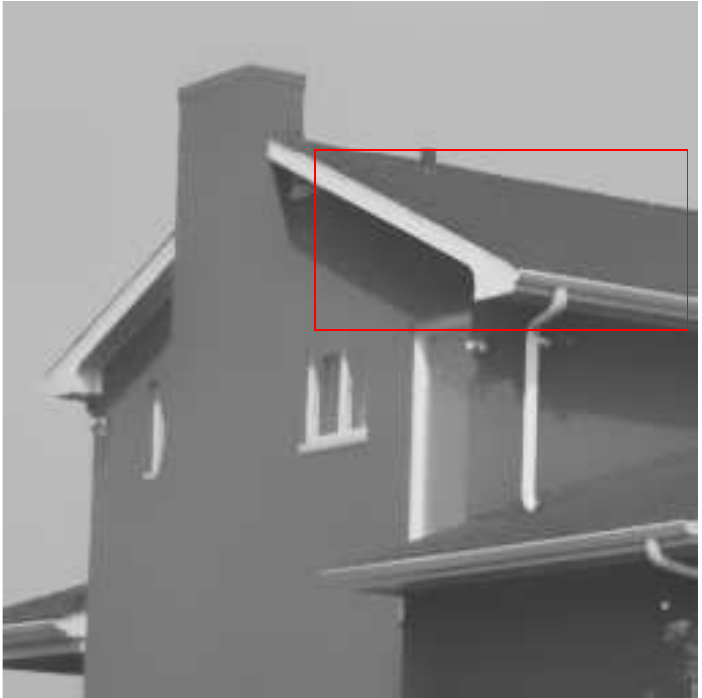}} \hspace{2mm}
\subfloat[Yang ($34$ dB).]{\includegraphics[width=0.34\linewidth]{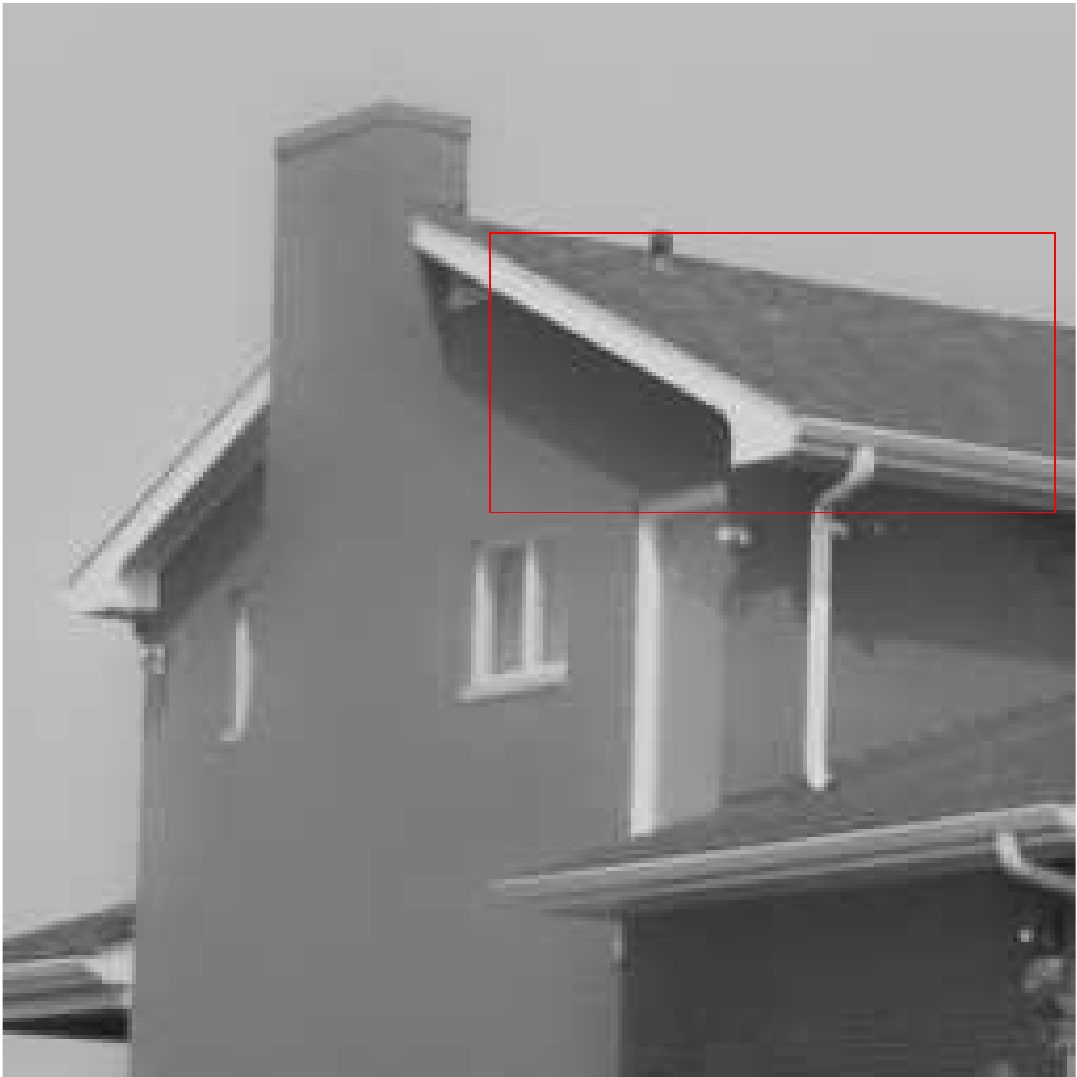}} 
\caption{Visual comparison of bilateral filtering for the  \textit{House} image. Filter parameters: $\sigma_s = 10$ and $\sigma_r = 30$. 
For a fair comparison, we have used four clusters for the proposed method and four bins for Yang's method \cite{Yang2009}.
Notice that our result is better than Yang's, both visually (compare boxed areas) and in terms of $\mathrm{PSNR}$. 
This is because we use non-uniform quantization (clustering) and data-driven approximation (see Figure \ref{fig:approx}).}
\label{Visualfig1}
\end{figure}

As mentioned earlier, although the proposed method is targeted at high-dimensional filtering, we can outperform \cite{Yang2009}, which is considered the state-of-the-art for bilateral filtering of grayscale images. 
We demonstrate this in Table \ref{timingbil2} and Figure \ref{Visualfig1} using a couple of examples.
In particular, notice in Figure \ref{Visualfig1} that artifacts are visible in Yang's result (when $K$ is small.). 
Our $\mathrm{PSNR}$ is almost $10$ dB higher, and our output does not show any artifacts. 
A detailed PSNR comparison is provided in Table \ref{timingbil2}. We notice that, for the same timing, our PSNR is consistently better than \cite{Yang2009}.
Note that we have set $K=4$ as the number of clusters (resp. quantization bins) for our method (resp. \cite{Yang2009}). 

\setlength\tabcolsep{1.5pt}
\begin{table}[!htp]
\caption{Comparison of approximation accuracy ($\mathrm{PSNR}$) for bilateral filtering of \textit{Barbara} ($512 \times 512$). 
Average timing is $530$ ms for Yang and $550$ ms for proposed method.}
\centering
\resizebox{\columnwidth}{!}{%
 \begin{tabular}{|c | c | c  c  c  c  c  c  c  c  c  c|}
 \hline
 $\sigma_s$\textbackslash{}$\sigma_r$ & & $10$ & $20$ & $30$ & $40$ & $50$ & $60$ & $70$ & $80$ & $90$ & $100$\\ [0.5ex]
 \hline
 & & \multicolumn{9}{c}{\textbf{Yang \cite{Yang2009}}} & \\
 \hline
 $10$ & $\mathrm{\textbf{PSNR}}$ & $51.48$ & $53.01$ & $53.48$ & $53.65$ & $53.48$ & $53.17$ & $53.01$ & $52.71$ & $52.28$ & $52.01$ \\ [0.1ex]
 \hline
 $30$ & $\mathrm{\textbf{PSNR}}$ & $51.23$ & $53.17$ & $53.81$ & $53.98$ & $53.98$ & $53.81$ & $53.81$ & $53.98$ & $53.98$ & $53.98$ \\ [0.1ex]
 \hline
 $30$ & $\mathrm{\textbf{PSNR}}$ & $50.86$ & $53.32$ & $53.81$ & $53.98$ & $53.82$ & $53.65$ & $53.65$ & $53.82$ & $53.98$ & $54.33$ \\ [0.1ex]
 \hline 
 & & \multicolumn{9}{c}{\textbf{Proposed}} & \\
 \hline
 $10$ & $\mathrm{\textbf{PSNR}}$ & $56.08$ & $64.61$ & $61.69$ & $59.50$ & $57.50$ & $56.09$ & $54.88$ & $53.81$ & $53.32$ & $52.71$ \\ [0.1ex]
 \hline
 $30$ & $\mathrm{\textbf{PSNR}}$ & $56.53$ & $68.13$ & $64.61$ & $62.11$ & $60.53$ & $59.19$ & $58.03$ & $57.25$ & $56.77$ & $56.09$ \\ [0.1ex]
 \hline
 $30$ & $\mathrm{\textbf{PSNR}}$ & $57.01$ & $70.07$ & $67.30$ & $64.61$ & $62.56$ & $60.89$ & $60.17$ & $59.19$ & $58.88$ & $58.59$ \\ [0.1ex]
 \hline
\end{tabular}
}
\label{timingbil2}
\end{table}

For completeness, we have reported the timings for different $\sigma_s$ in Table \ref{timingbil1}, where we have used Young's fast algorithm for  performing the Gaussian convolutions \cite{young1995recursive}. 
As expected, the timing essentially does not  change with $\sigma_s$.
\setlength\tabcolsep{0.5pt}
\begin{table}[!htp]
\caption{timings for bilateral filtering of \textit{Barbara} for varying $\sigma_s$ ($\sigma_r=100, K=16$). }
\centering
% \resizebox{\columnwidth}{!}{%
 \begin{tabular}{|c ||  c@{\hskip 0.11in}   c@{\hskip 0.11in}   c@{\hskip 0.11in}   c@{\hskip 0.11in}   c@{\hskip 0.11in}   c@{\hskip 0.11in}   c@{\hskip 0.11in}   c|}
 \hline
 $\sigma_s$ & $10$ & $20$ & $30$ & $40$ & $50$ & $60$ & $70$ & $80$\\ [0.5ex]
 \hline
 \textbf{Time (ms)} & $496$ & $498$ & $475$ & $486$ & $500$ & $489$ & $485$ & $481$\\ [0.1ex]
 \hline
\end{tabular}
%}
\label{timingbil1}
\end{table}

\begin{figure}[!htp]
\centering
\subfloat[Input ($256 \times 256$).]{\includegraphics[width=0.39\linewidth]{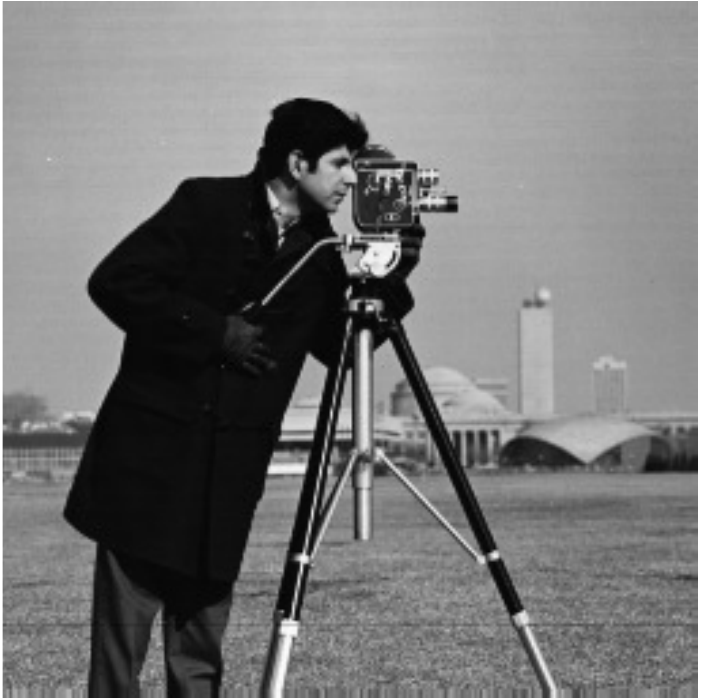}} \hspace{2mm}
\subfloat[Brute-force ($495$ ms).]{\includegraphics[width=0.39\linewidth]{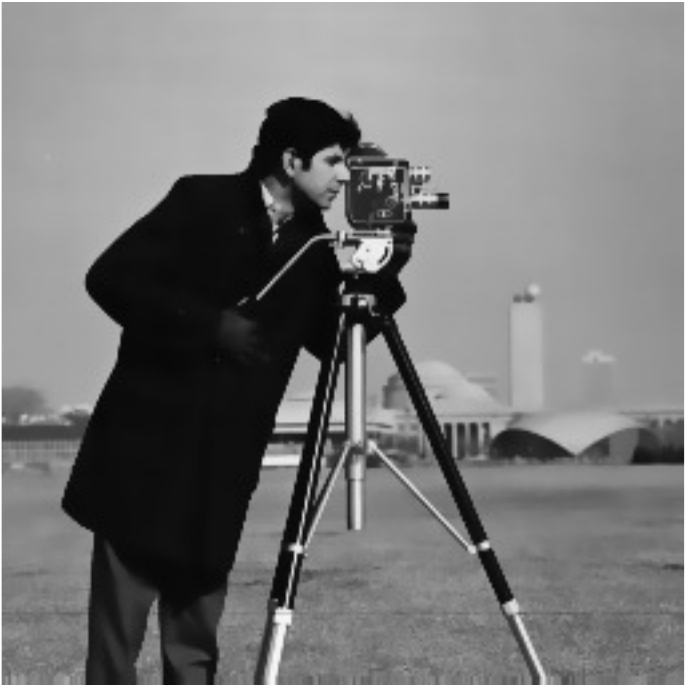}} \\
\subfloat[Using \eqref{LS} ($5$ sec, $35.3$ dB).]{\includegraphics[width=0.39\linewidth]{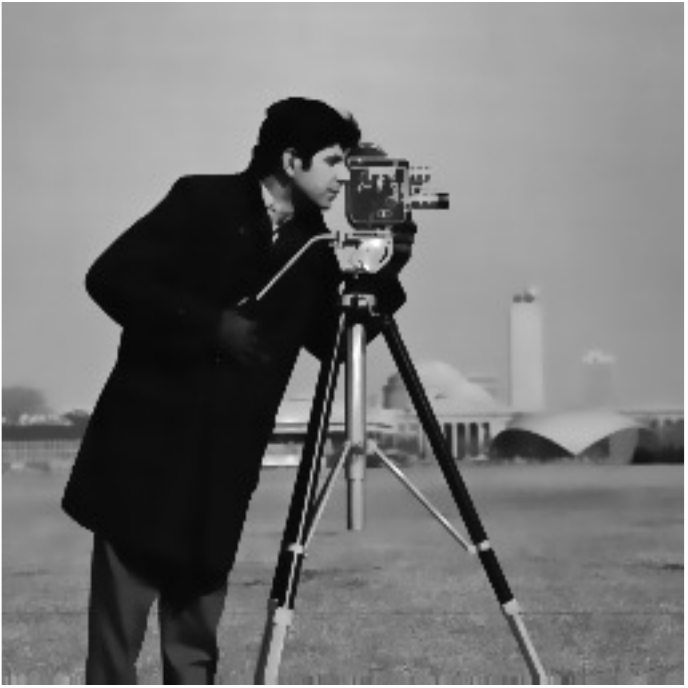}} \hspace{2mm}
\subfloat[\textbf{Using \eqref{opt} ($\textbf{27}$ ms, $\textbf{32.6}$ dB).}]{\includegraphics[width=0.39\linewidth]{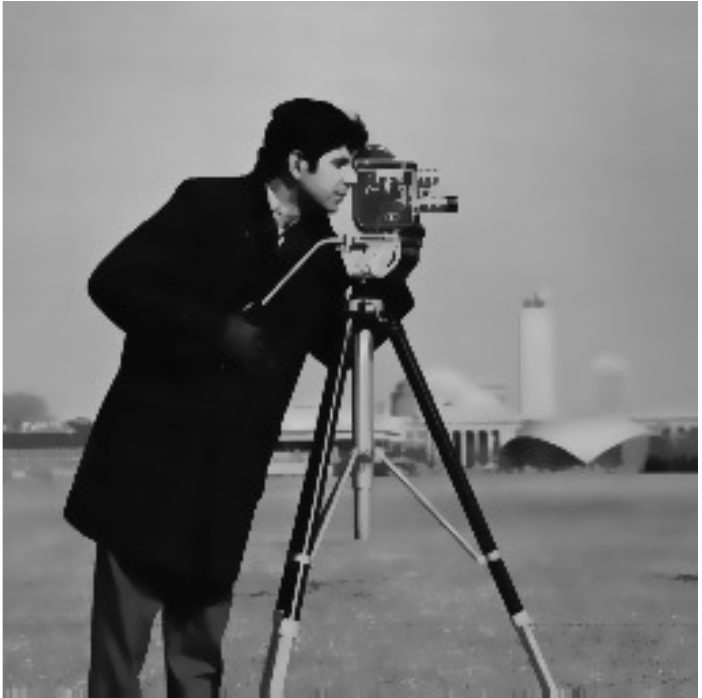}} 
\caption{Bilateral filtering of a grayscale image. Parameters: $\sigma_s = 2,\sigma_r = 20,$ and $K=4$. The coefficients are obtained from \eqref{LS} and \eqref{opt} in (c) and (d).}
\label{Approxfig1}
\end{figure}

In Figure \ref{Approxfig1}, we have compared the bilateral filtering results obtained using \eqref{LS} and \eqref{opt}. Since \eqref{opt} is an approximation of \eqref{LS}, we see a drop in PSNR using \eqref{LS}, but importantly there is a significant speed up using \eqref{opt}. 
In fact, the time required using \eqref{LS} is more than that of the brute-force implementation.

\subsection{Color Bilateral Filtering}

\begin{figure}[!htp]
 \centering
%\hspace{11mm}
\includegraphics[width=0.55\linewidth]{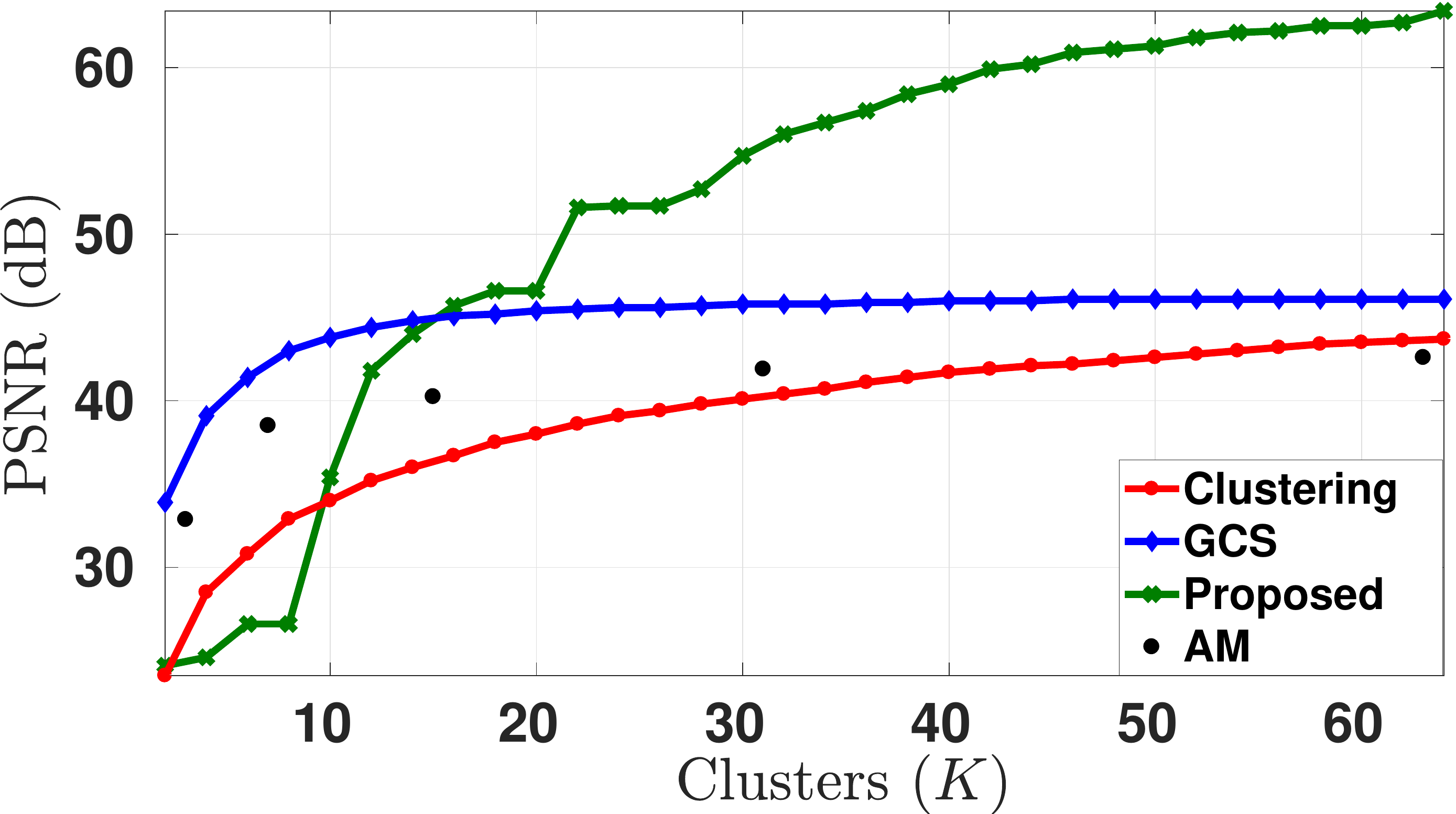}
\vspace{-2mm}
\caption{Scaling of filtering accuracy (average $\mathrm{PSNR}$) with number of clusters (manifolds for AM), for bilateral filtering of color images from the Kodak dataset. The parameters are $\sigma_s=20$ and $\sigma_r=40$. }
\label{compare1}
\end{figure}

We next perform fast bilateral filtering of color images, for which the state-of-the-art methods are AM \cite{gastal2012adaptive} and GCS \cite{mozerov2015global}. 
We have also compared with our previous work \cite{pravin2017filtering}, which we simply refer to as ``Clustering''.
In Figure \ref{compare1}, we have studied the effect of  the number of clusters  on the filtering accuracy, for our method, GCS, AM and \cite{pravin2017filtering}. 
While GCS performs better for small $K$ (coarse approximation), our method outperforms GCS when $K>15$. 
Note that the number of manifolds can only take dyadic (discrete) values \cite{gastal2012adaptive}.
 
\begin{figure}[!htp]
\begin{minipage}{0.42\linewidth}
				\centering
 \setlength\tabcolsep{1.5pt}		
  \begin{tabular}{|c || c | c | c|}
 \hline
size\textbackslash{}$K$ & $4$ & $8$ & $16$ \\ [0.5ex]
 \hline
 $1080$P & $33.54$ & $48.68$ & $53.86$ \\ [0.1ex]
 \hline
 $4$K & $31.29$ & $47.42$ & $53.16$ \\ [0.1ex]
 \hline
 $8$K & $34.25$ & $46.90$ & $54.03$ \\ [0.1ex]
 \hline
\end{tabular}
% \vspace{0.5mm}
\captionof{table}{$\mathrm{PSNR}$ for color images ($\sigma_s=10$, $\sigma_r=50$).}
\label{timingbil5}
	\end{minipage}\hfill
	\begin{minipage}{0.52\linewidth}
	        \vspace{-2mm}
		\centering
		\includegraphics[width=0.9\linewidth,height=0.57\linewidth]{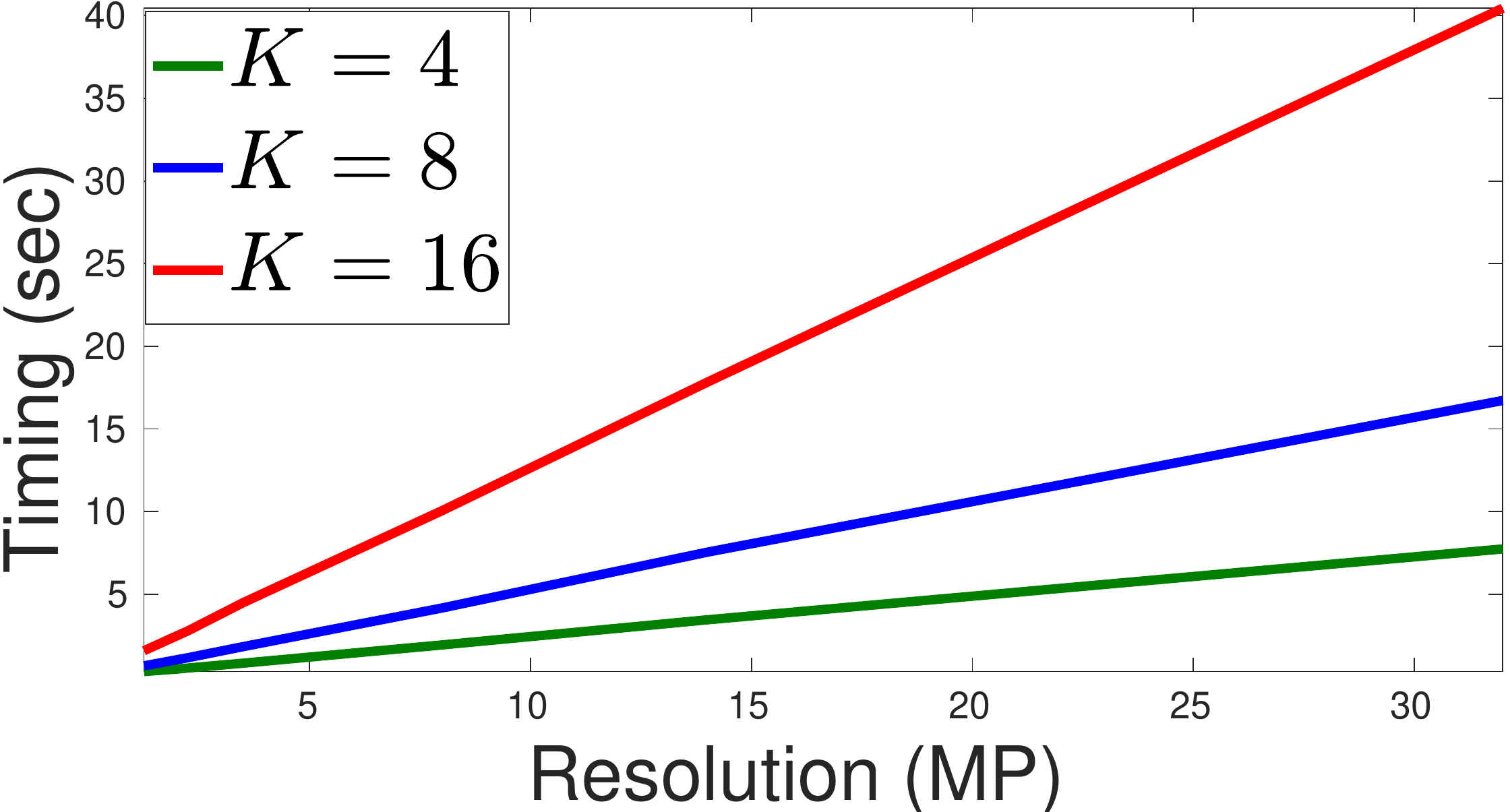} \vspace{-3mm}
		\caption{Timing (seconds) vs. resolution (MP) for color filtering.}
		\label{figuretimingres}
	\end{minipage}
\end{figure}
In Table \ref{timingbil5}, we report the filtering accuracy for full HD, $4K$ ultra HD, and $8K$ ultra HD by varying the number of clusters $K$. For a given resolution and $K$, we have averaged the PSNR over six high resolution images\footnote{https://hdqwalls.com} with that resolution. 
Notice that, irrespective of the image resolution, the PSNR values are above $40$ dB with just $8$ clusters. 
In Figure \ref{figuretimingres}, the timings for filtering color images are plotted for varying image sizes ($1.5$ MP to $32$ MP). This plot verifies the linear dependence of the timing on the image size, irrespective of the number of clusters used.

\begin{figure}[!htp]
\centering
\includegraphics[width=0.55\linewidth]{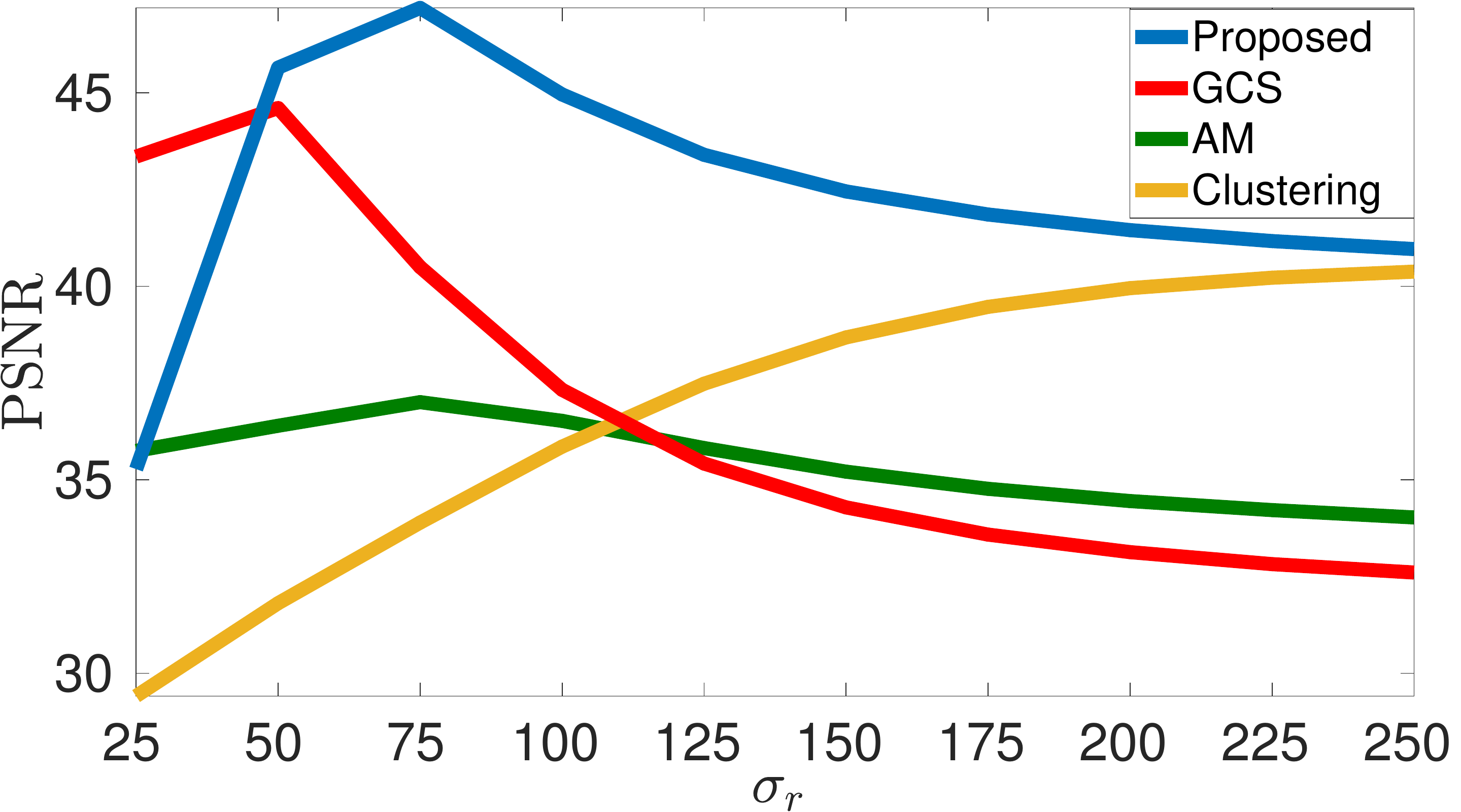}
\vspace{-2mm}
\caption{Approximation accuracy ($\mathrm{PSNR}$) for bilateral filtering of color \textit{Barbara} ($3 \times 512 \times 512$), at different $\sigma_r$ values averaged over $\sigma_s=10,20,30$. The average timings are: Proposed: $475$ ms, Global Color Sparseness: $490$ ms, Adaptive Manifolds: $370$ ms and Clustering \cite{pravin2017filtering}: $370$ ms.}
\label{Timingplot}
\hspace{-2mm}
\end{figure}

In Figure \ref{Timingplot}, we have compared the $\mathrm{PSNR}$ and timings  for bilateral filtering a color image, at different $\sigma_s$ and $\sigma_r$ values. We have used $K=16$ clusters for all three methods: proposed, GCS, and \cite{pravin2017filtering}. We use the default settings for the number of manifolds in AM.
Notice that we are better by $5\mbox{-}10$ dB, while the timings are roughly the same. 
In particular, notice the PSNR improvement that we obtain over \cite{pravin2017filtering} by optimizing the coefficients, i.e., by using \eqref{HDapprox} in place of \eqref{HDapprox2}.
Interestingly, \cite{pravin2017filtering} can outperform GCS and AM at large $\sigma_r$ values.

A visual comparison for color bilateral filtering was already shown in Figure \ref{Visualfig2}, along with the error images, timings, and $\mathrm{PSNR}$s. 
In particular, notice that the error from our method is smaller than AM and GCS.
This is interesting as our method is conceptually simpler than these algorithms --- AM has a complex formulation and GCS is based on two-pass filtering.

\begin{figure*}
\centering
\subfloat[$K=2$ ($15$ ms, $22$ dB).]{\includegraphics[width=0.185\linewidth]{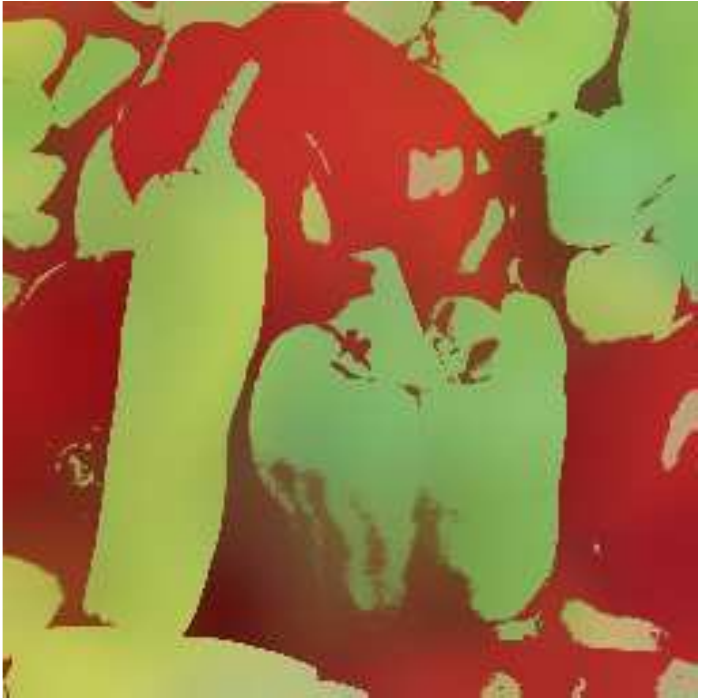}} \hspace{0.1mm}
\subfloat[$K=4$ ($28$ ms, $29$ dB).]{\includegraphics[width=0.185\linewidth]{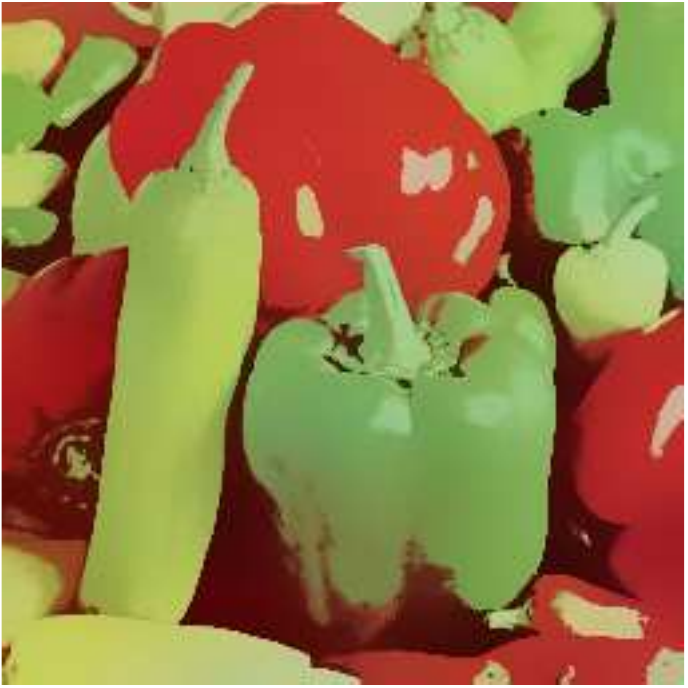}} \hspace{0.1mm}
\subfloat[$K=8$ ($52$ ms, $37$ dB).]{\includegraphics[width=0.185\linewidth]{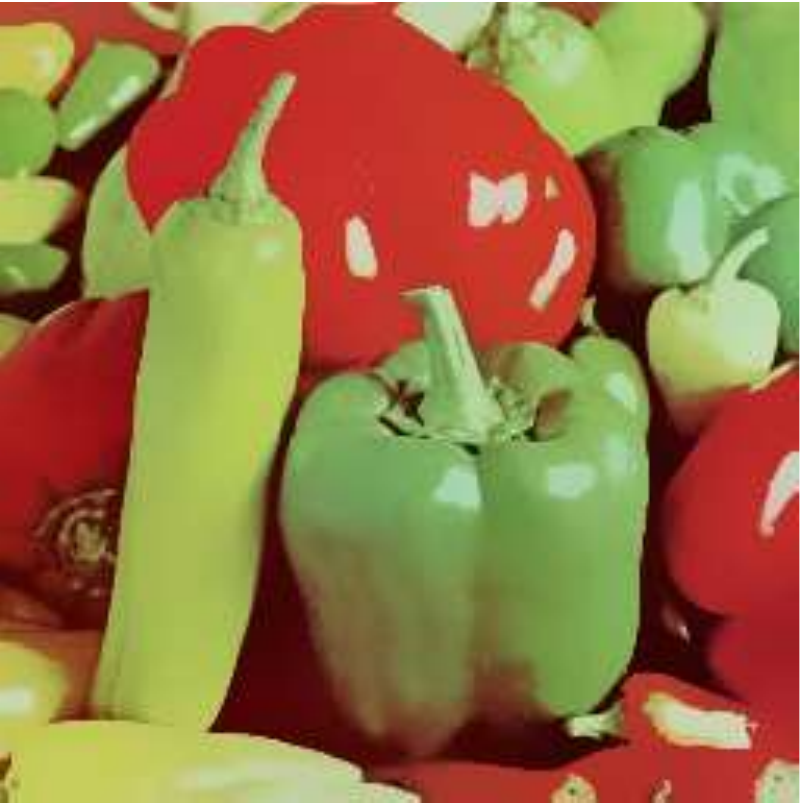}} \hspace{0.1mm}
\subfloat[$K=16$ ($100$ ms, $44$ dB).]{\includegraphics[width=0.185\linewidth]{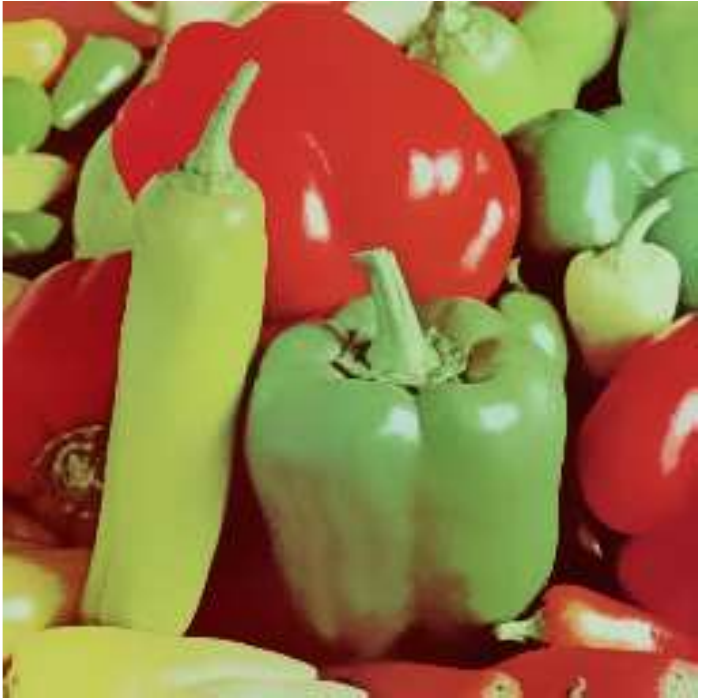}} \hspace{0.1mm}
\subfloat[Brute-force.]{\includegraphics[width=0.185\linewidth]{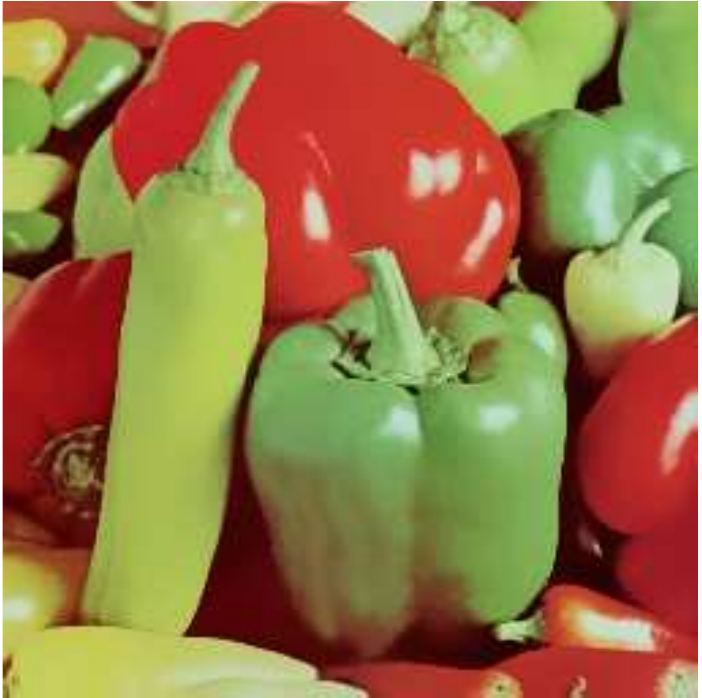}} 
\caption{Bilateral filtering of \textit{Peppers} for different values of $K$ ($\sigma_s=10$ and $\sigma_r=40$). Notice how the accuracy improves with $K$.}
\label{Visualfig3}
\end{figure*}

In Figure \ref{Visualfig3}, a visual result is provided to highlight how the approximation accuracy scales with $K$.   
Notice that the accuracy improves significantly as $K$ increases from $2$ to $16$. In fact, the approximation already resembles the result of brute-force filtering when $K$ is $8$. 

\begin{figure*}[!htp]
\centering
\subfloat[{\scriptsize Clean/Noisy ($22$ dB, $\sigma{\tiny =}20$}).]{\includegraphics[width=0.185\linewidth]{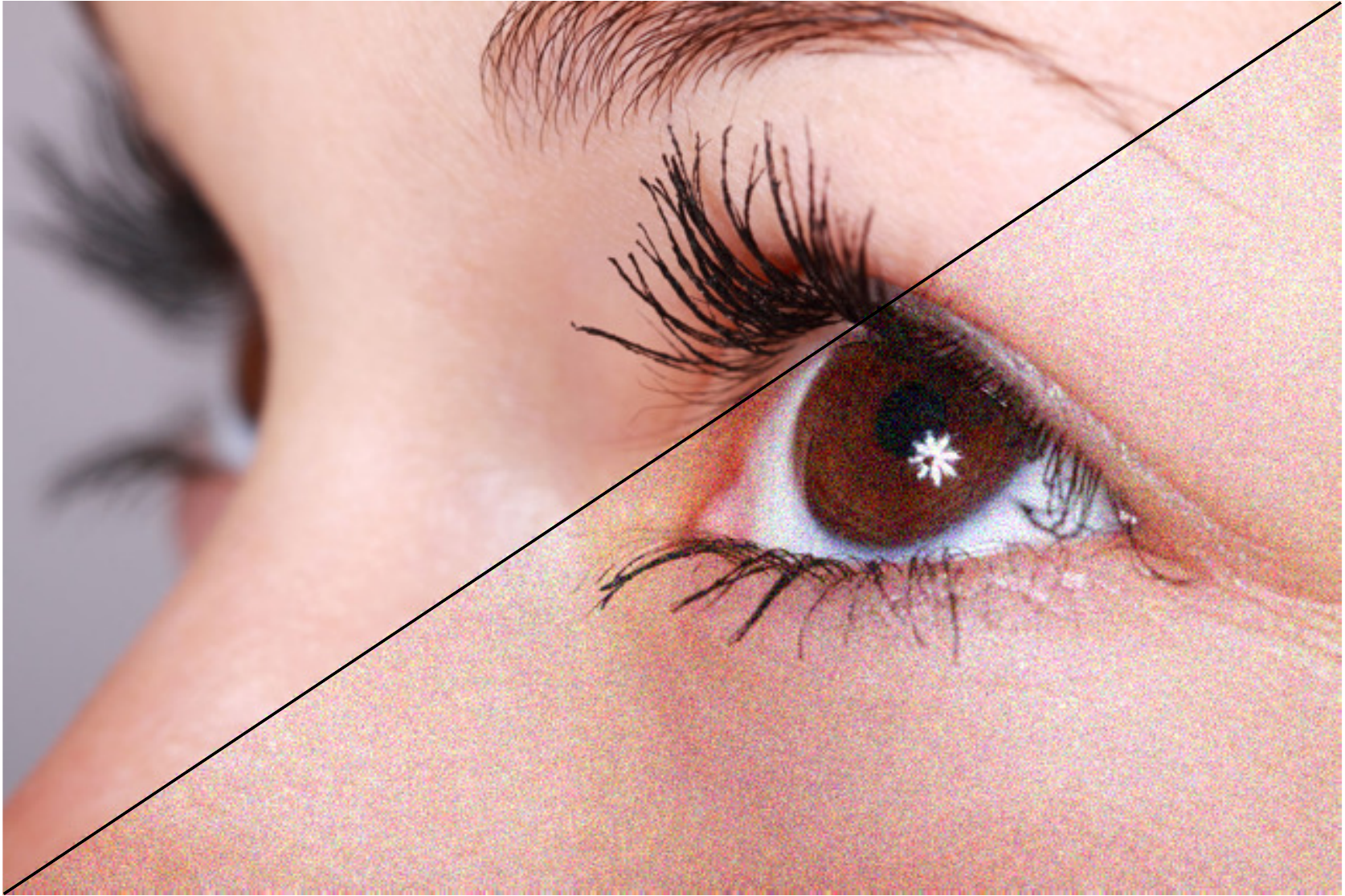}} \hspace{0.01mm}
\subfloat[{\scriptsize \cite{tasdizen2009principal} ($8$ min, $33.3$ dB, $0.92$)}.]{\includegraphics[width=0.185\linewidth]{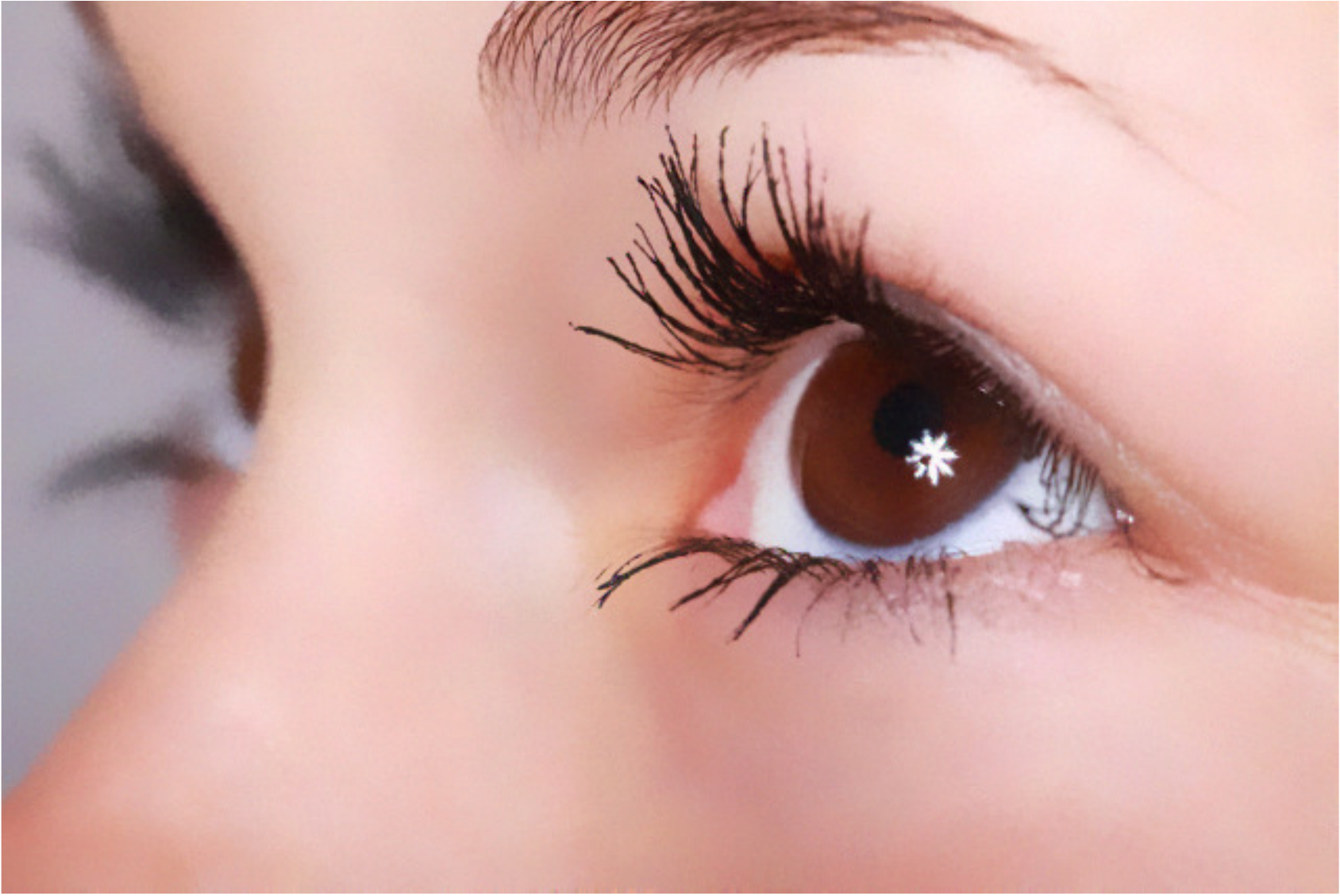}} \hspace{0.01mm}
\subfloat[{\scriptsize \textbf{Ours ($\textbf{1.7}$ sec, $\textbf{31.2}$ dB, $\textbf{0.92}$)}}.]{\includegraphics[width=0.185\linewidth]{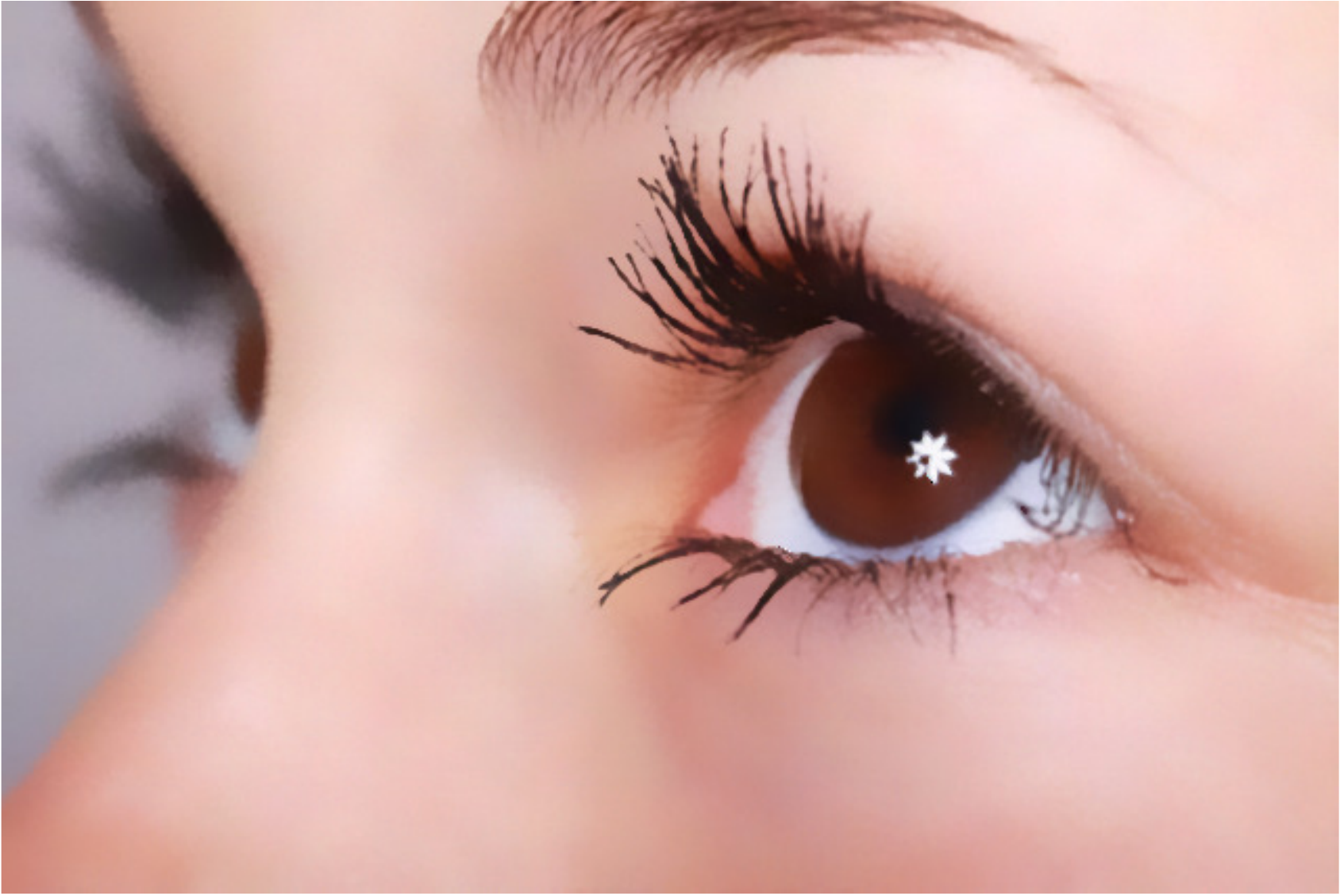}} \hspace{0.01mm}
\subfloat[{\scriptsize AM ($3$ sec, $30.7$ dB, $0.90$)}.]{\includegraphics[width=0.185\linewidth]{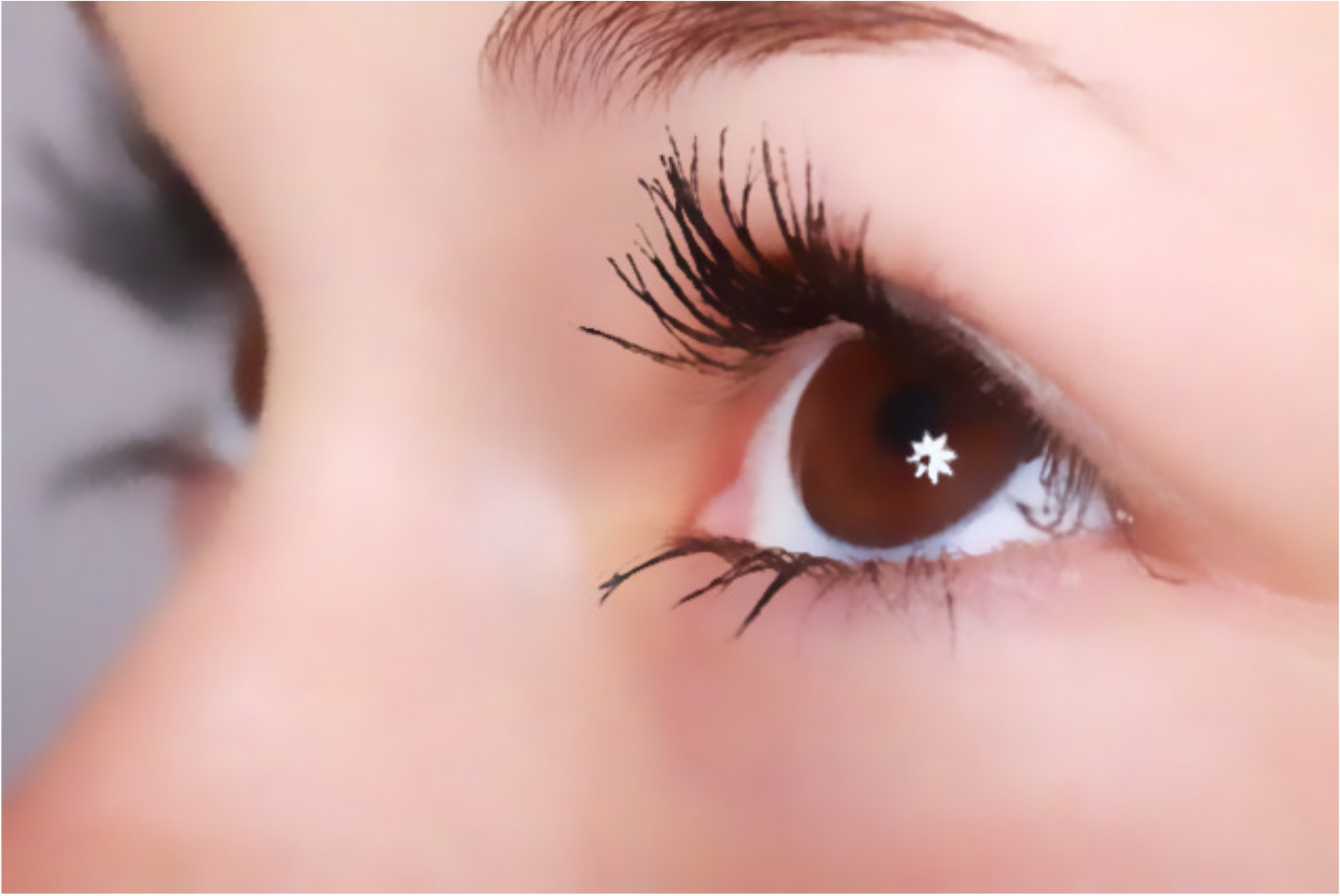}} \hspace{0.01mm}
\subfloat[{\scriptsize \cite{zhang2017beyond} ($3.5$sec, $36.2$ dB,  $0.93$)}.]{\includegraphics[width=0.185\linewidth]{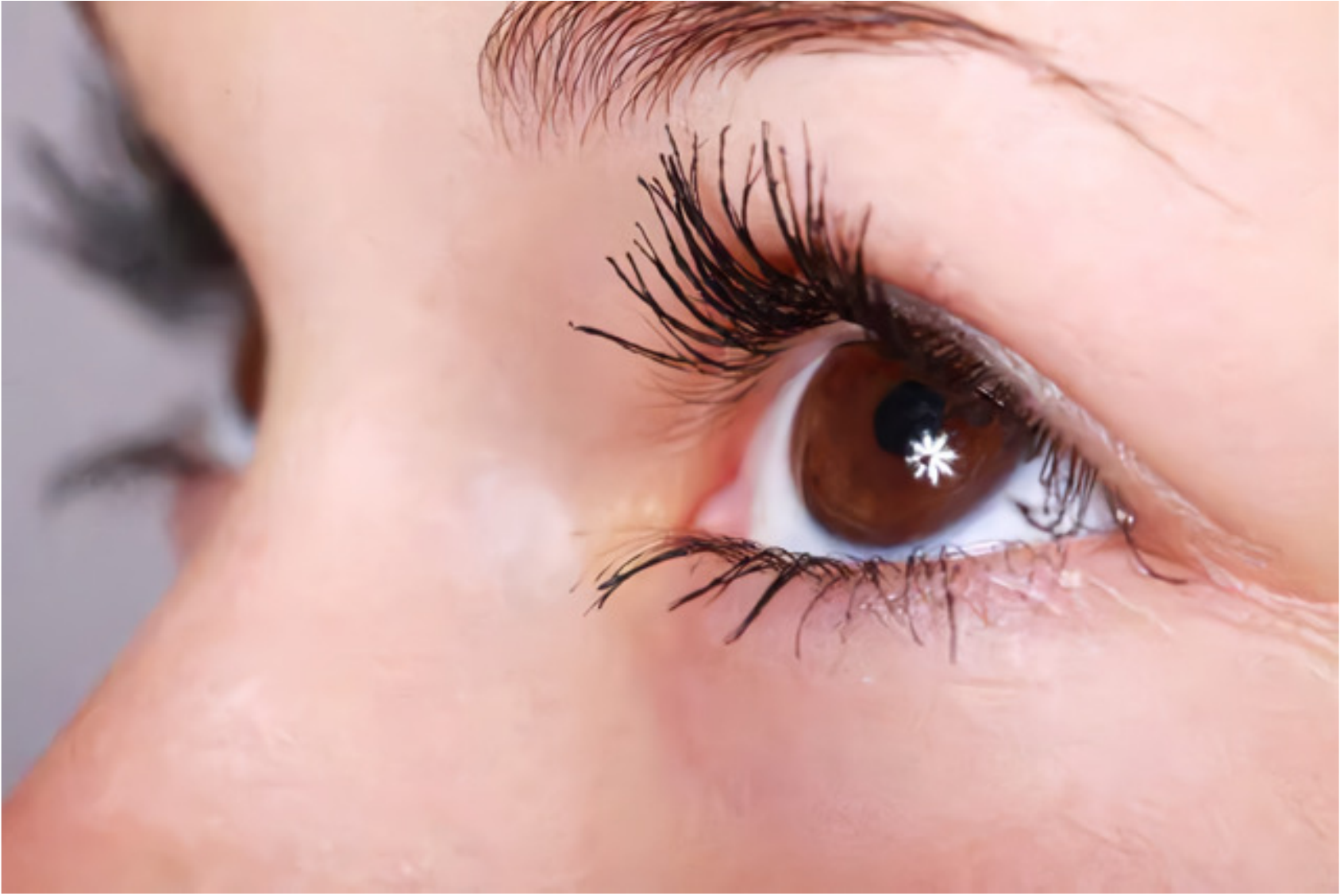}} 

\subfloat[{\scriptsize Clean/Noisy ($12$ dB, $\sigma{\tiny =}63$}).]{\includegraphics[width=0.185\linewidth]{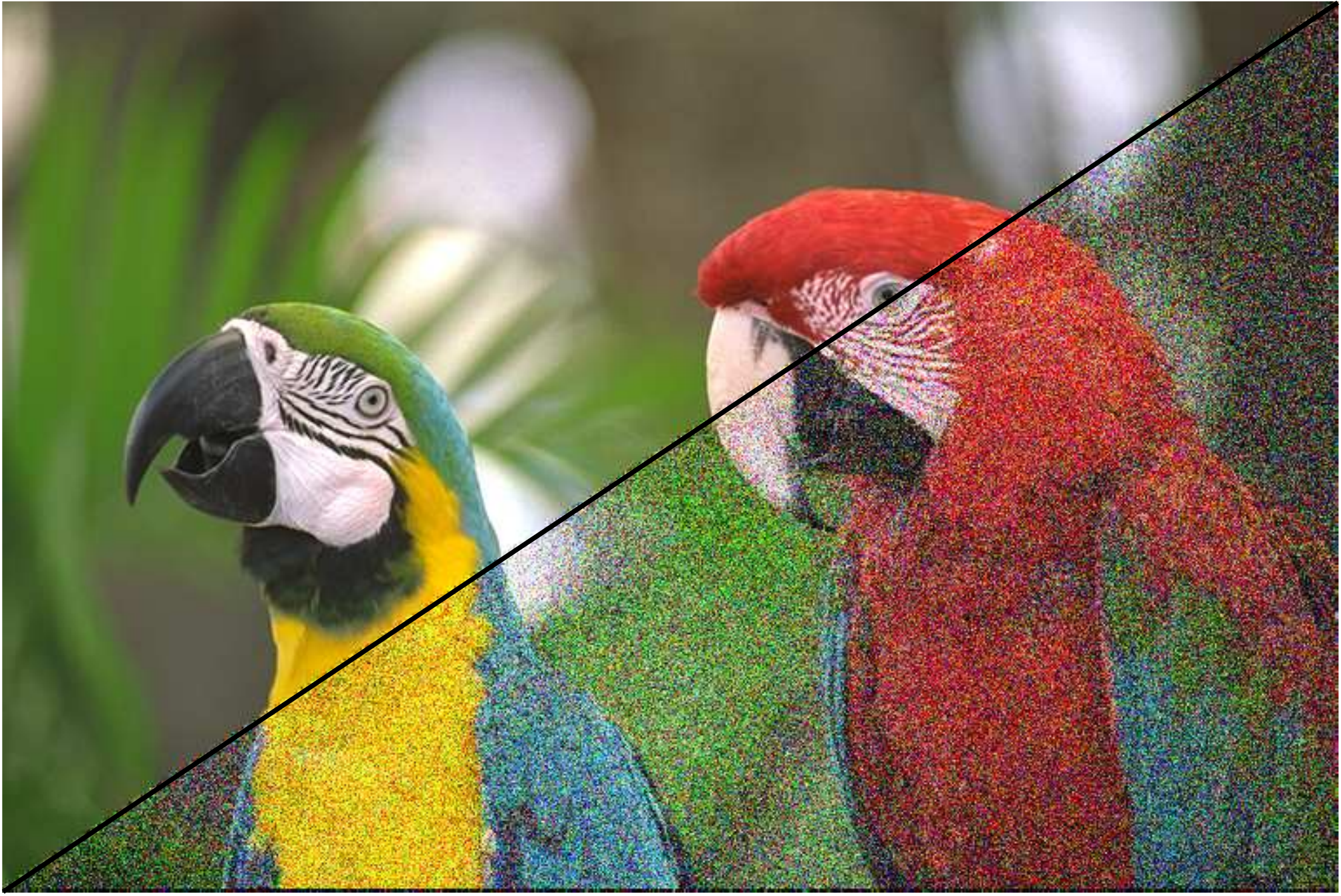}} \hspace{0.01mm}
\subfloat[{\scriptsize \cite{tasdizen2009principal} ($9$ min, $26.7$ dB, $0.77$)}.]{\includegraphics[width=0.185\linewidth]{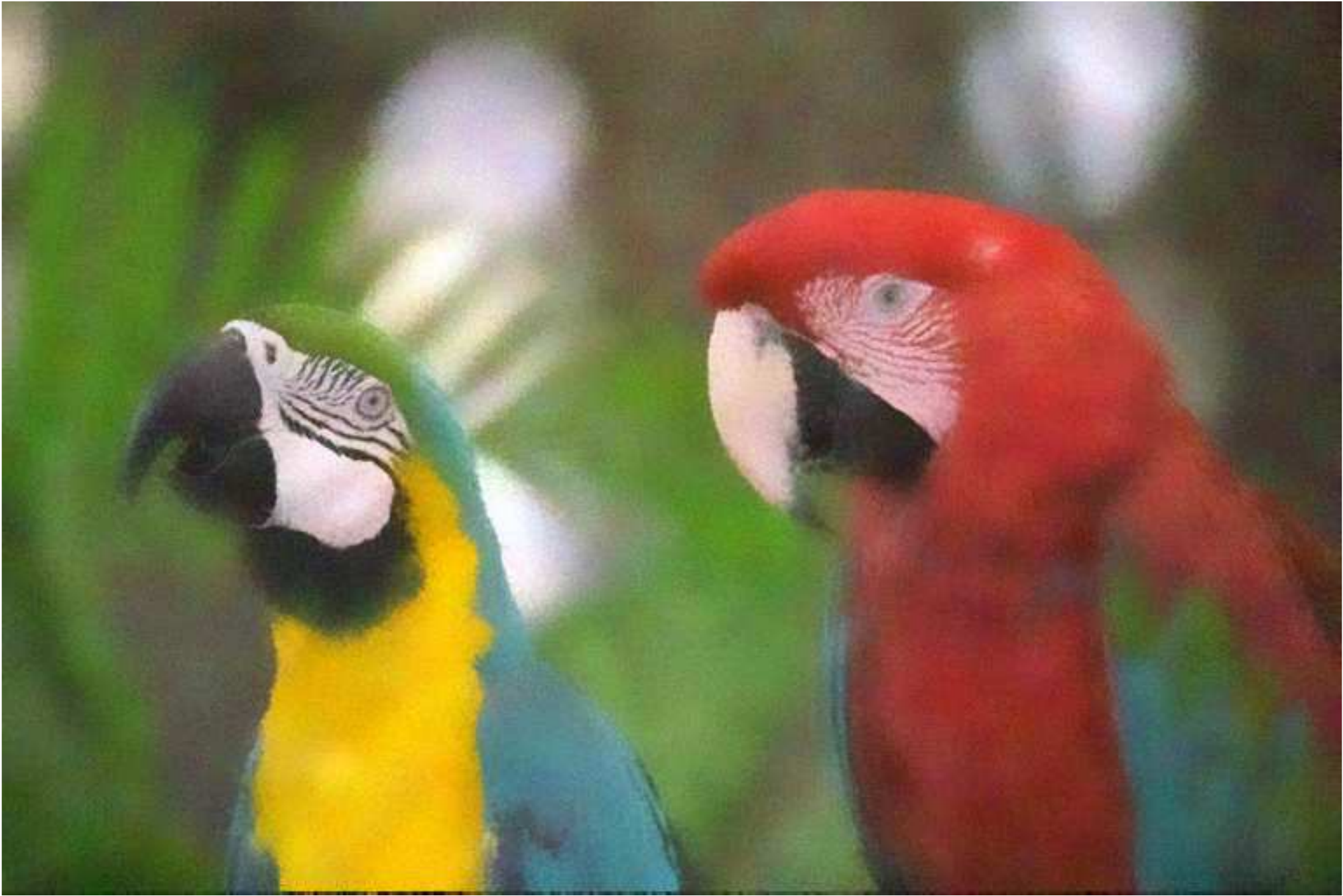}} \hspace{0.01mm}
\subfloat[{\scriptsize \textbf{Ours ($\textbf{2.7}$ sec, $\textbf{26.8}$ dB, $\textbf{0.82}$)}}.]{\includegraphics[width=0.185\linewidth]{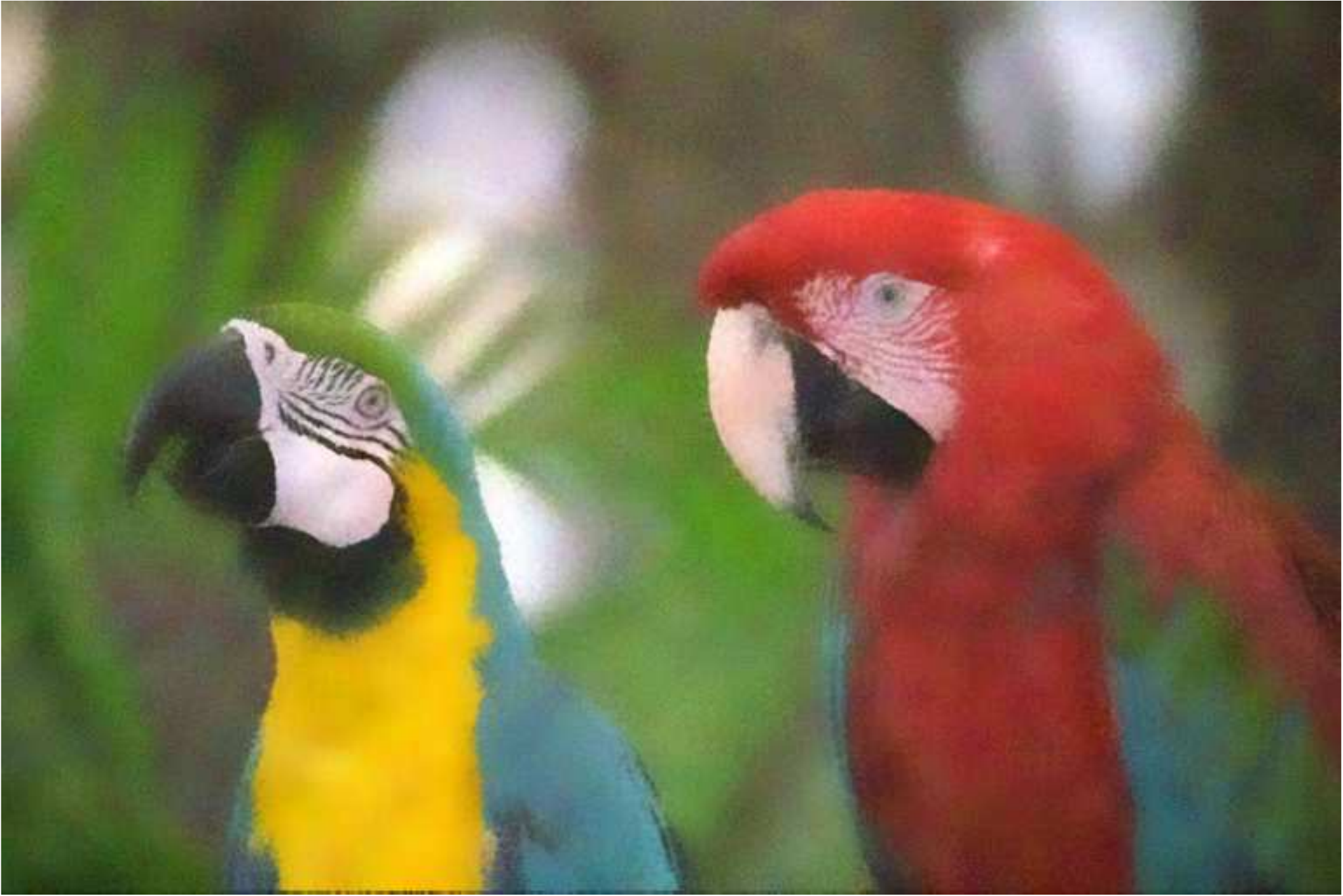}} \hspace{0.01mm}
\subfloat[{\scriptsize AM ($4.2$ sec, $25$ dB, $0.70$)}.]{\includegraphics[width=0.185\linewidth]{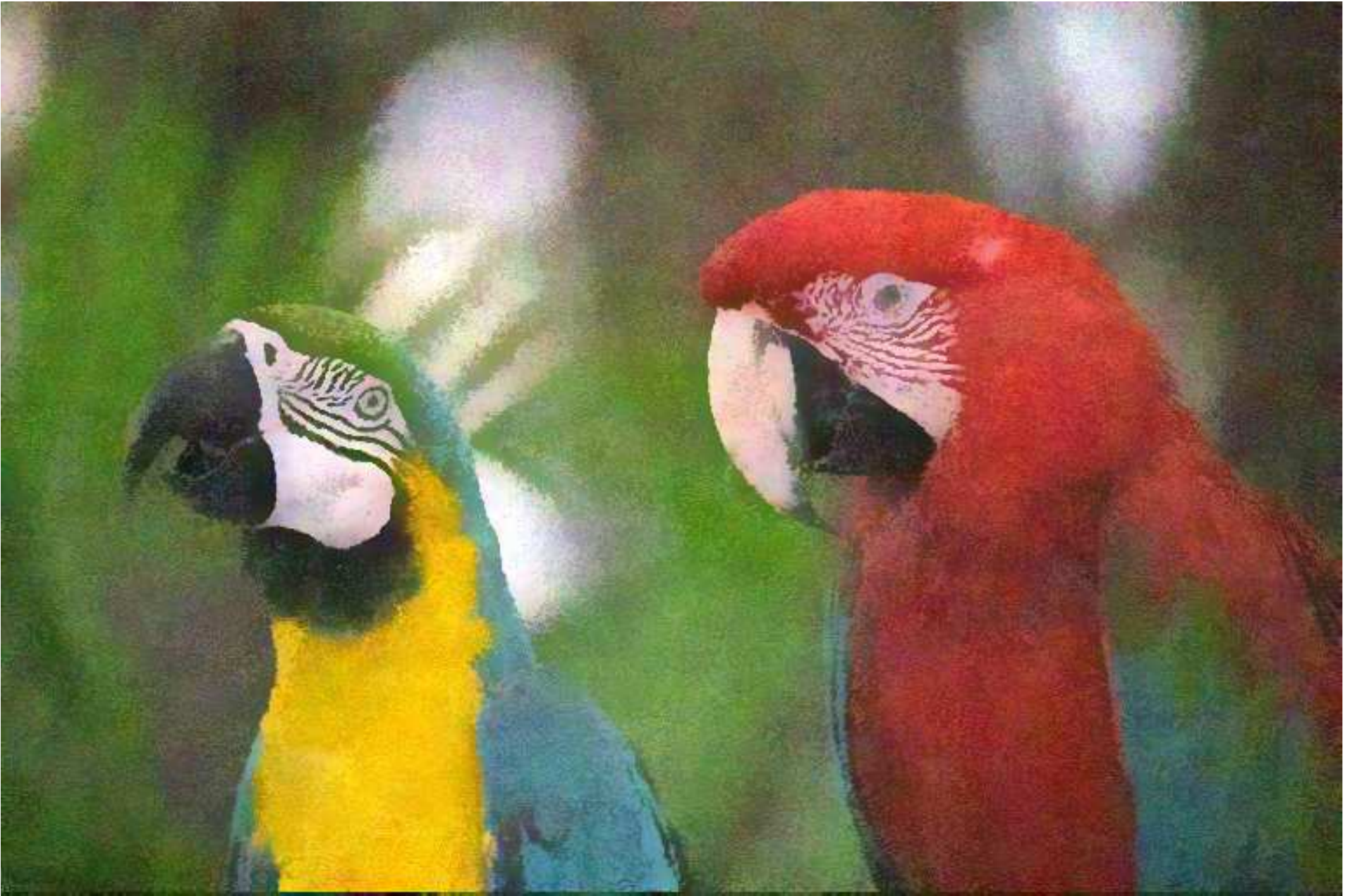}} \hspace{0.01mm}
\subfloat[{\scriptsize \cite{zhang2017beyond} ($5$ sec, $31$ dB, $0.86$)}.]{\includegraphics[width=0.185\linewidth]{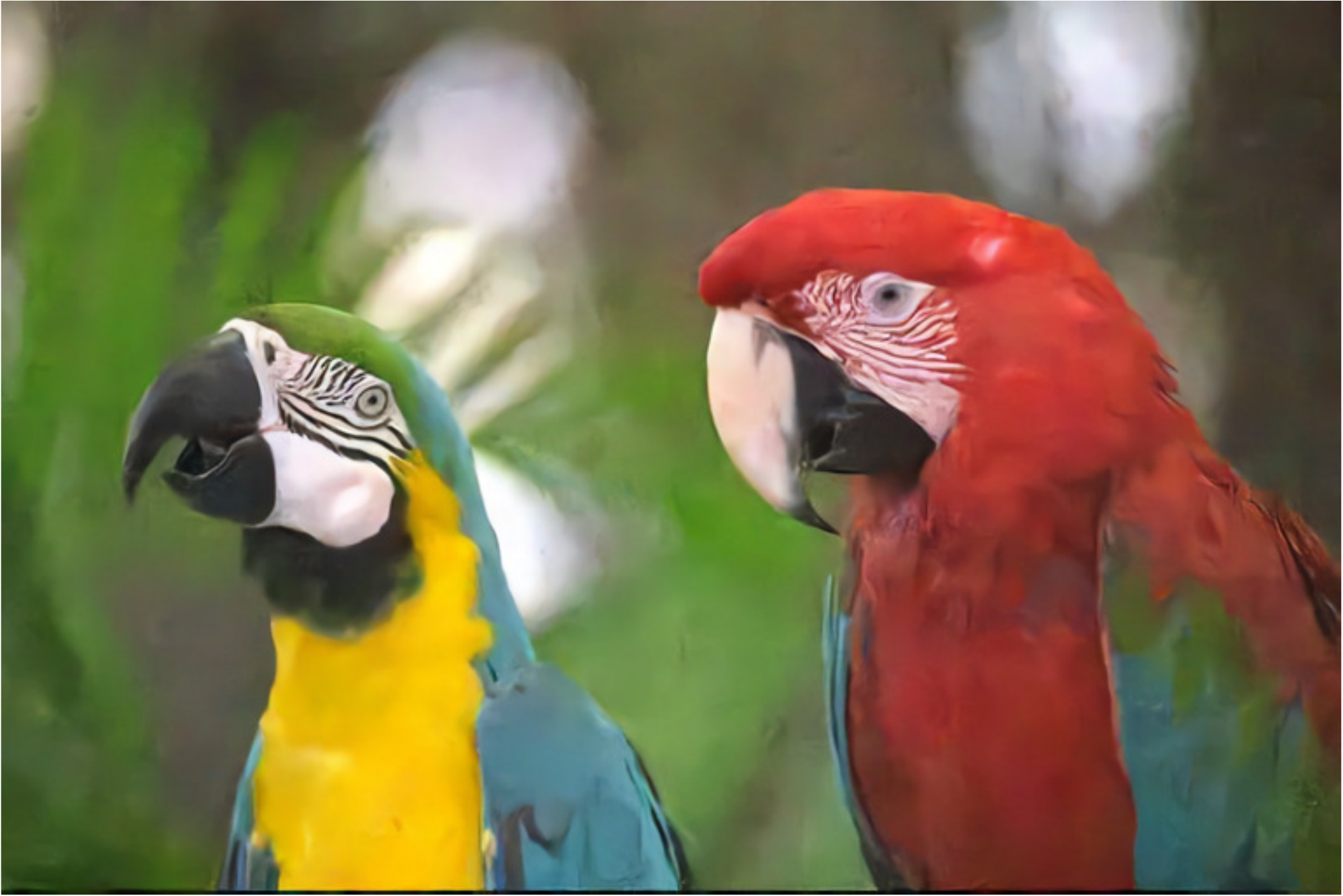}} 
\caption{Color image denoising (Gaussian noise) using PCA-NLM \cite{tasdizen2009principal}, where the patch and search sizes are $7 \times 7$ and $21 \times 21$. 
For PCA-NLM, AM and our method, PCA was used to reduce the range dimension from $3 \times 7  \times 7$ to $25$. 
We used $31$ clusters for our method. The number of manifolds was automatically set to $15$ in the AM code \cite{gastal2012adaptive}. 
The run-time, $\mathrm{PSNR}$ and SSIM for AM with $31$ manifolds are $6$ sec, $31.1$ dB and $0.90$ (top image) and  $8.3$ sec, $23.7$ dB and $0.69$ (bottom image). 
For comparison, we have also shown the result from a state-of-the-art Gaussian denoiser \cite{zhang2017beyond}. 
}
\label{Visualfig4}
\end{figure*}

\subsection{Color Nonlocal Means}

In NLM, $\f(\i)$ is the noisy image (corrupted with additive Gaussian noise) and $\p(\i)$ is a square patch of pixels around $\i$.
In particular, if the patch is $m \times m$, then the dimension of $\p(\i)$ is $\rho=3m^2$ for a color image.
A popular trick to accelerate  NLM (called PCA-NLM) is to project the patch onto a lower-dimensional space using PCA \cite{tasdizen2009principal}. 
We note that PCA was used for reducing the patch dimension (as explained earlier) for both AM and our algorithm. 

A visual comparison is provided in Figure \ref{Visualfig4} for low and high noise, where the superior performance of our method over AM is evident. We have also compared with a state-of-the-art Gaussian denoiser (CPU implementation\footnote{https://github.com/cszn/FFDNet}) based on deep neural nets \cite{zhang2017beyond}. It is not surprising that the result from \cite{zhang2017beyond} is better both in terms of PSNR and visual quality. However, we are somewhat faster.
Figure \ref{Visualfig4} also suggests that our method is visually closer to PCA-NLM at both low and high noise levels, albeit with a significant speedup ($10$ min to $2.7$ sec).

\begin{figure}
\centering
\subfloat{\includegraphics[width=0.5\linewidth]{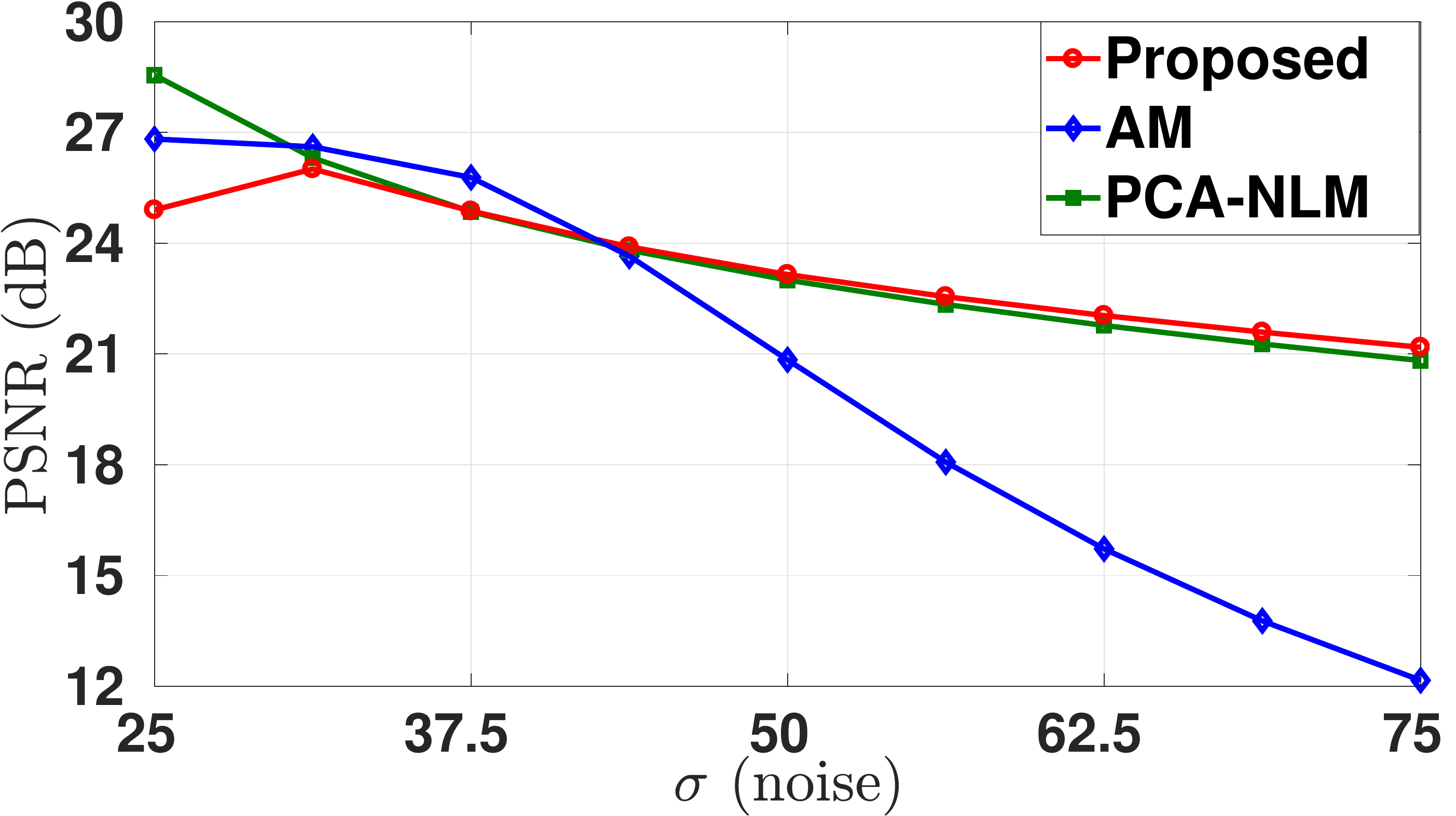}}
\subfloat{\includegraphics[width=0.5\linewidth]{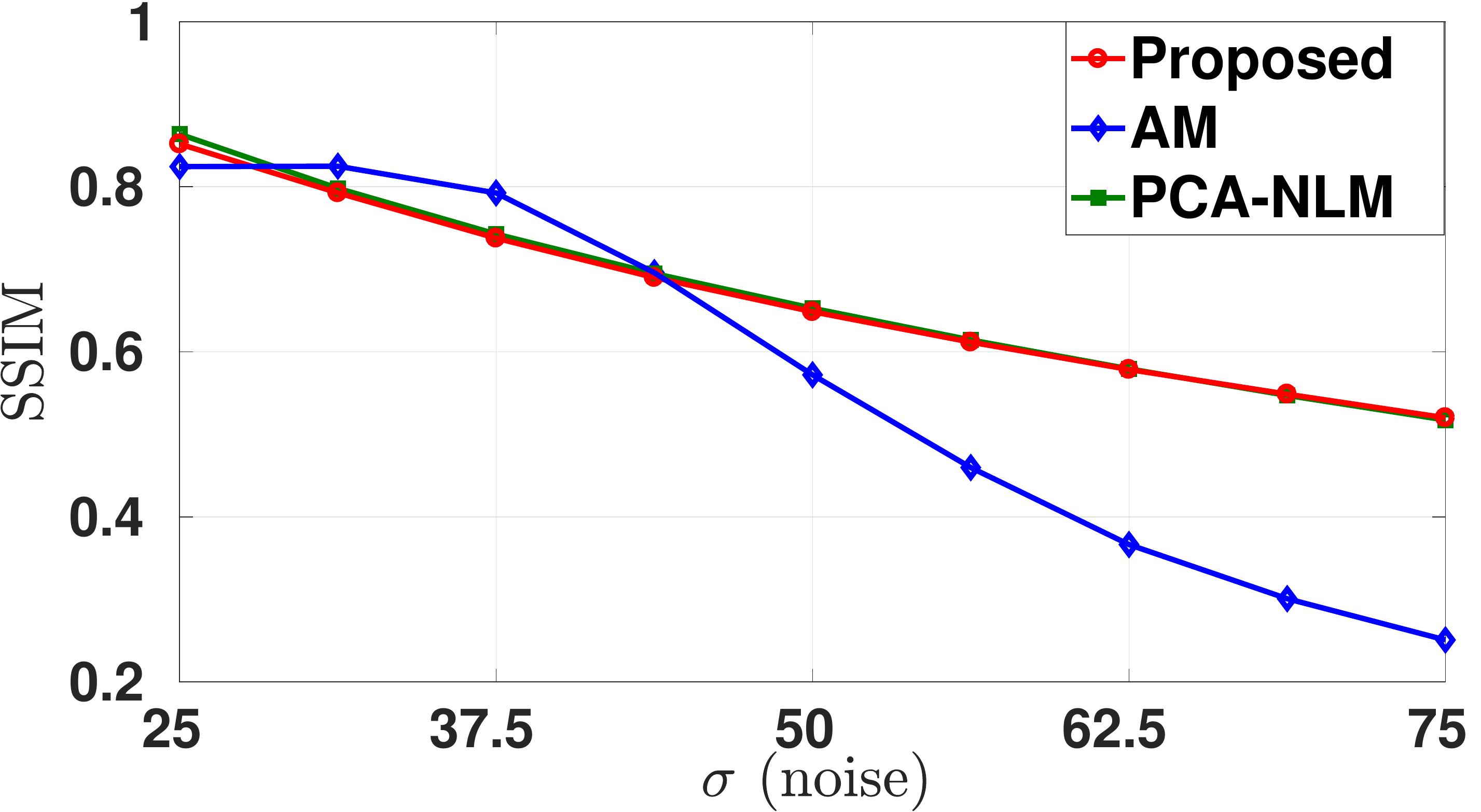}} 
\caption{Comparison of the denoising performances of PCA-NLM \cite{tasdizen2009principal}, Adaptive Manifolds, and the proposed method. The noise level is denoted by $\sigma (\times 255)$. The PCA dimension was set to $6$ for all methods. The $\mathrm{PSNR}$ and $\mathrm{SSIM}$ \cite{wang2004} values are averaged over the images from the Kodak dataset. 
The parameters used are $K = 31$, $S = 10,$ and $m = 3$.}
\label{VisualfigPSNR4}
\end{figure}

In Figure \ref{VisualfigPSNR4}, we have reported Gaussian denoising results on the Kodak dataset for different noise levels ($\sigma$).  
While the proposed method and AM perform similarly for small $\sigma$, the former is more robust when $\sigma$ is large. 
In fact, the denoising performance of AM degrades rapidly with the increase in $\sigma$. Another important point is that our result is close to PCA-NLM in terms of PSNR and SSIM for all noise levels.
\subsection{Hyperspectral Denoising}

\begin{figure}[!htp]
\centering
\subfloat[Input \cite{hyperspectraldata}.]{\includegraphics[width=0.38\linewidth]{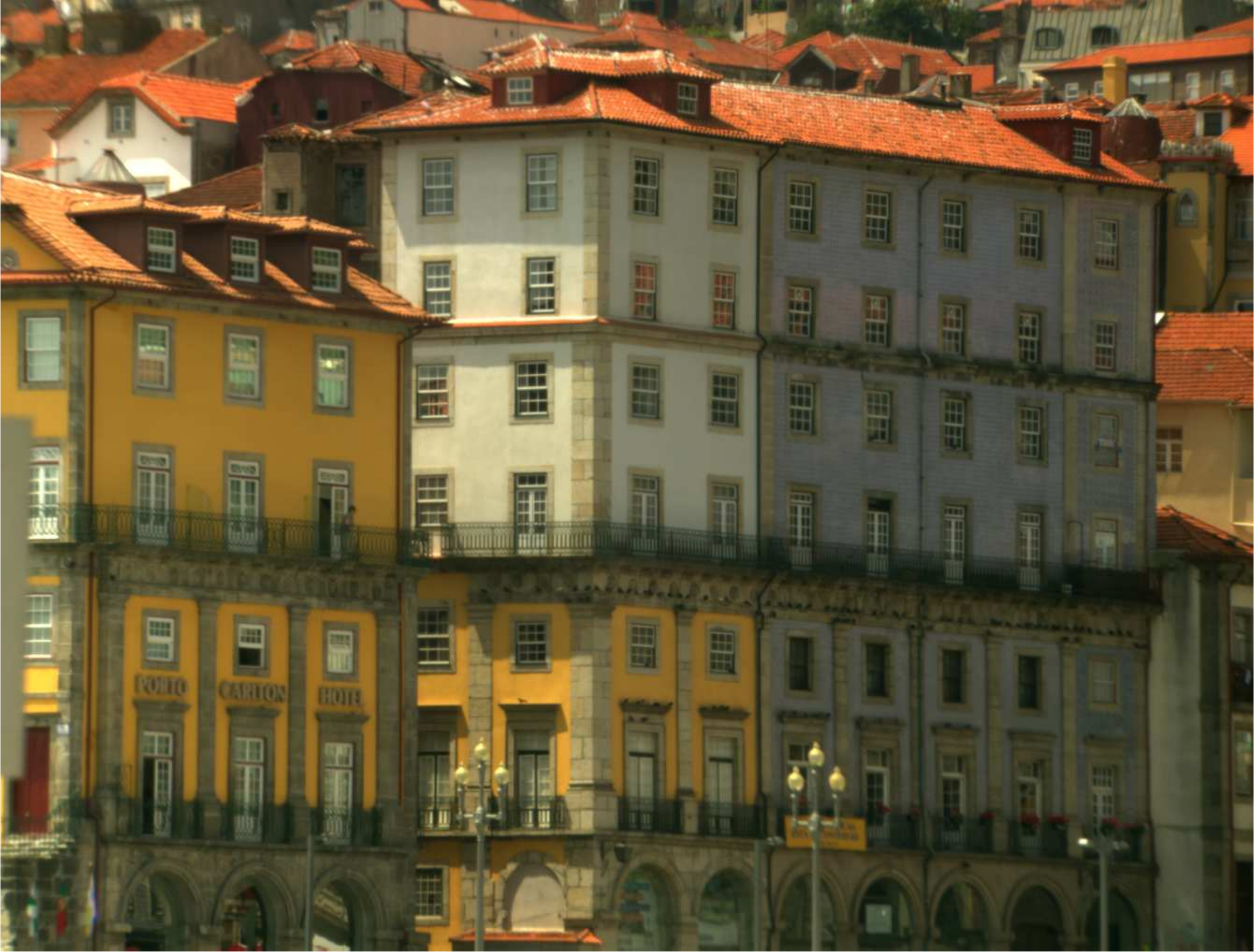}} \hspace{2.5mm}
\subfloat[440nm band.]{\includegraphics[width=0.38\linewidth]{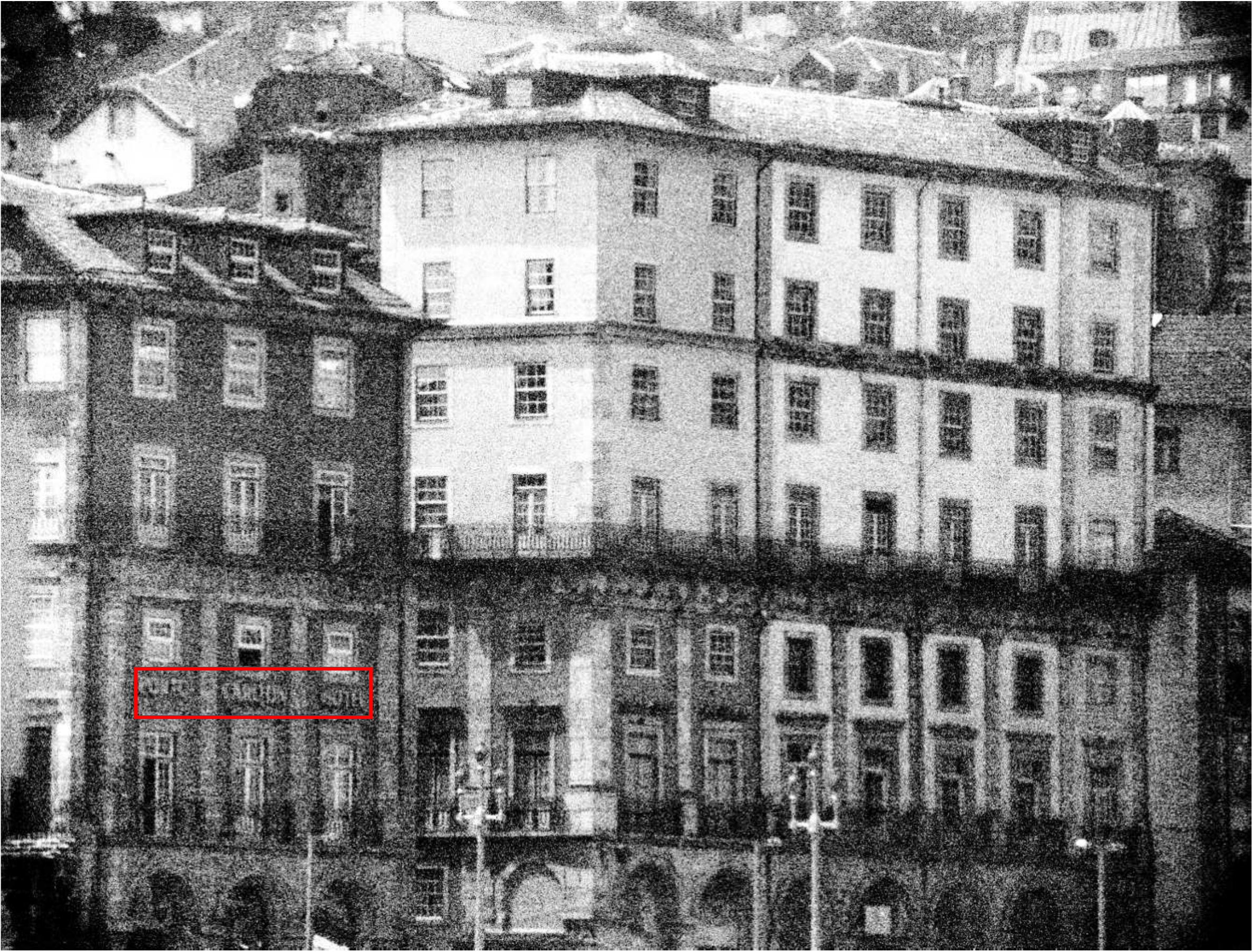}} 

\subfloat[Brute-force ($27$ min).]{\includegraphics[width=0.38\linewidth]{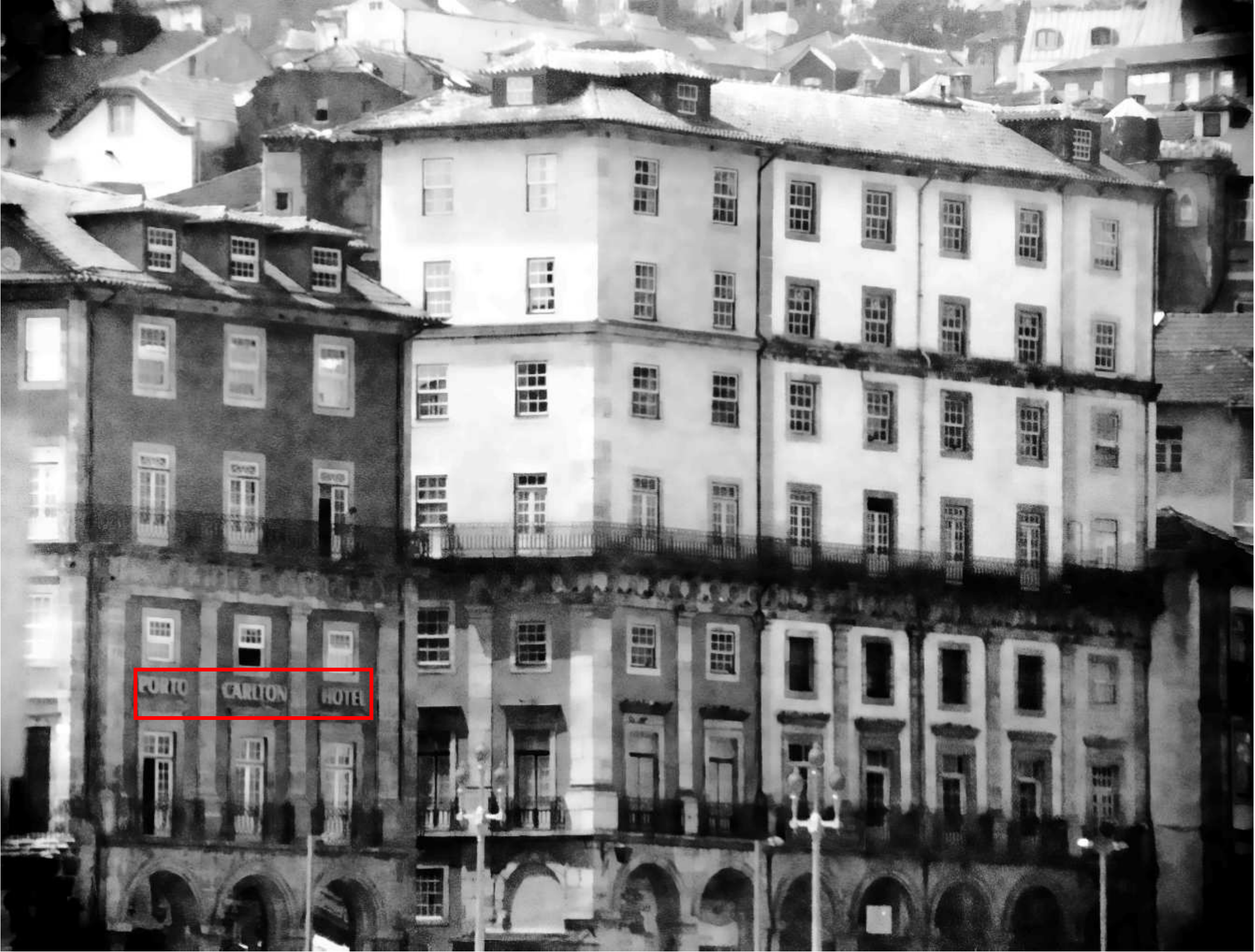}} \hspace{2.5mm}
\subfloat[\textbf{Proposed ($\textbf{15}$ sec).}]{\includegraphics[width=0.38\linewidth]{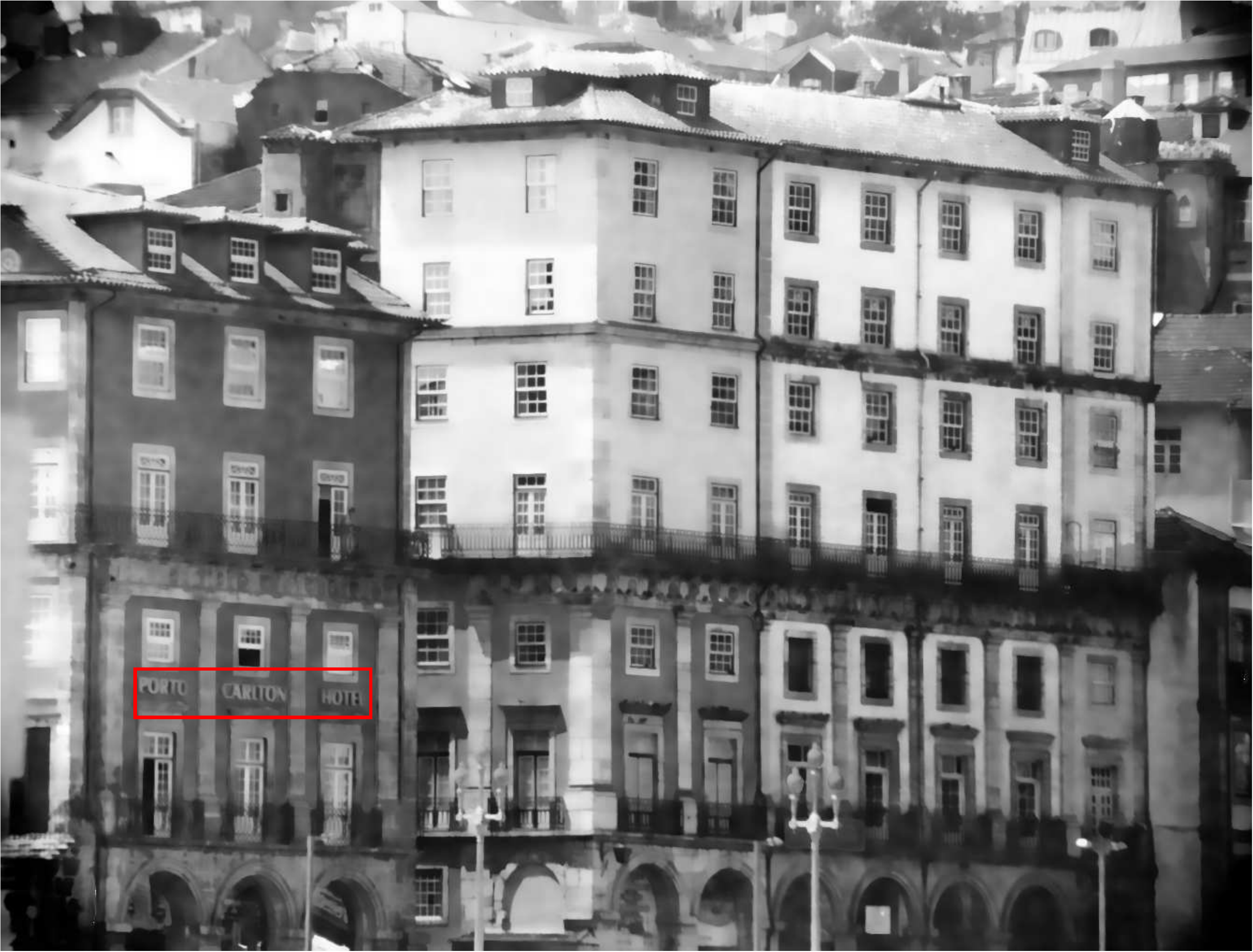}} 
\caption{Denoising of a hyperspectral image $(1340 \times 1017 \times 33)$ using bilateral filtering ($\sigma_s = 5$ and $\sigma_r = 100$).  
The image in (b) shows one of the noisy bands; the same band  after filtering is shown in (c) and (d). We have shown  one of the bands just for visualization; the filtering was performed on the entire hyperspectral image and not on a band-by-band basis.}
\label{Visualfig6}
\end{figure}

We next perform denoising of hyperspectral images using the bilateral filter. For such images, the range dimension (i.e., the number of spectral bands) is high. 
In Figure \ref{Visualfig6}, denoising results are exclusively shown to highlight the speedup of our approximation over brute-force implementation for a $33$-band image. Notice the dramatic reduction in timing compared to the brute-force implementation. Moreover, notice that the denoising quality is in fact quite good for our method (compare the text in the boxed regions). 

\begin{figure*}[!htp]
\captionsetup[subfigure]{labelformat=empty}
\centering
\subfloat{\includegraphics[width=0.05\linewidth]{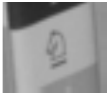}} 
\subfloat{\includegraphics[width=0.11\linewidth,height=0.044\linewidth]{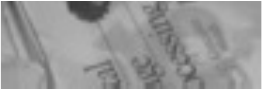}} \hspace{1mm}
\subfloat{\includegraphics[width=0.05\linewidth]{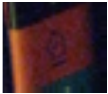}} 
\subfloat{\includegraphics[width=0.11\linewidth,height=0.044\linewidth]{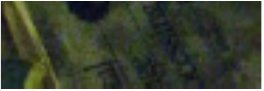}} \hspace{1mm}
\subfloat{\includegraphics[width=0.05\linewidth]{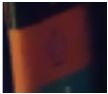}}
\subfloat{\includegraphics[width=0.11\linewidth,height=0.044\linewidth]{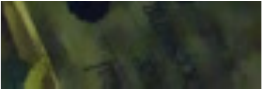}} \hspace{1mm}
\subfloat{\includegraphics[width=0.05\linewidth]{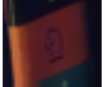}} 
\subfloat{\includegraphics[width=0.11\linewidth,height=0.044\linewidth]{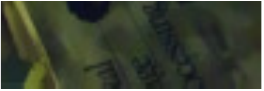}} \hspace{1mm}
\subfloat{\includegraphics[width=0.05\linewidth]{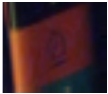}} 
\subfloat{\includegraphics[width=0.11\linewidth,height=0.044\linewidth]{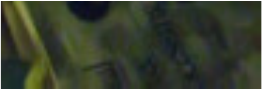}} 
\\[-1ex]
\subfloat[(a) Infrared input.]{\includegraphics[width=0.16\linewidth]{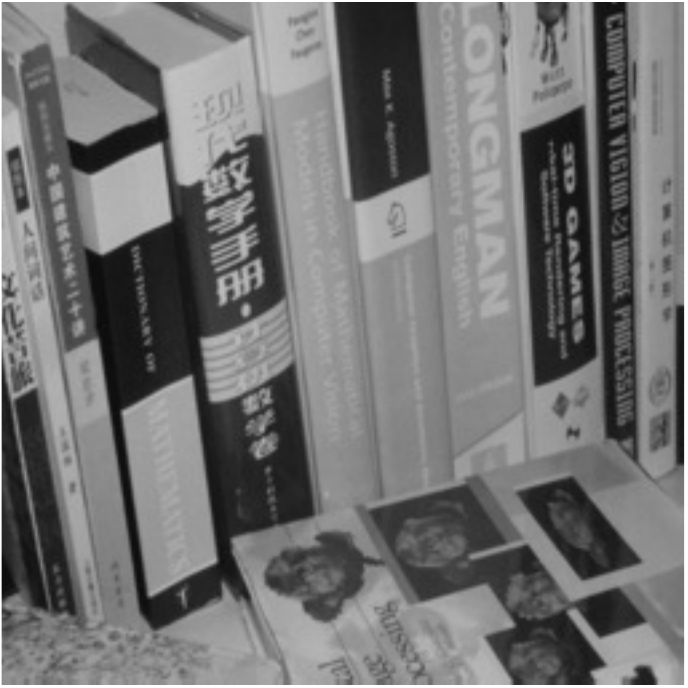}} \hspace{1mm}
\subfloat[(b) Noisy low-light input.]{\includegraphics[width=0.16\linewidth]{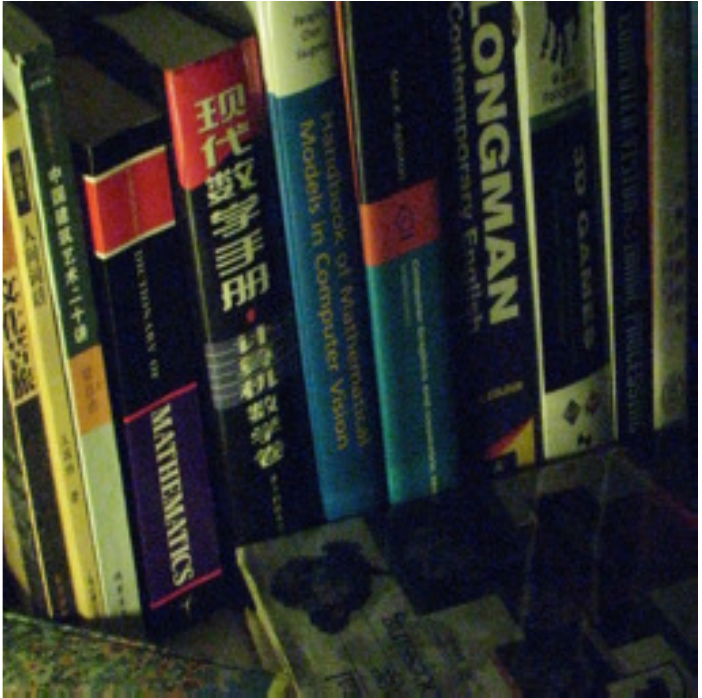}} \hspace{1mm}
\subfloat[(c) Without infrared.]{\includegraphics[width=0.16\linewidth]{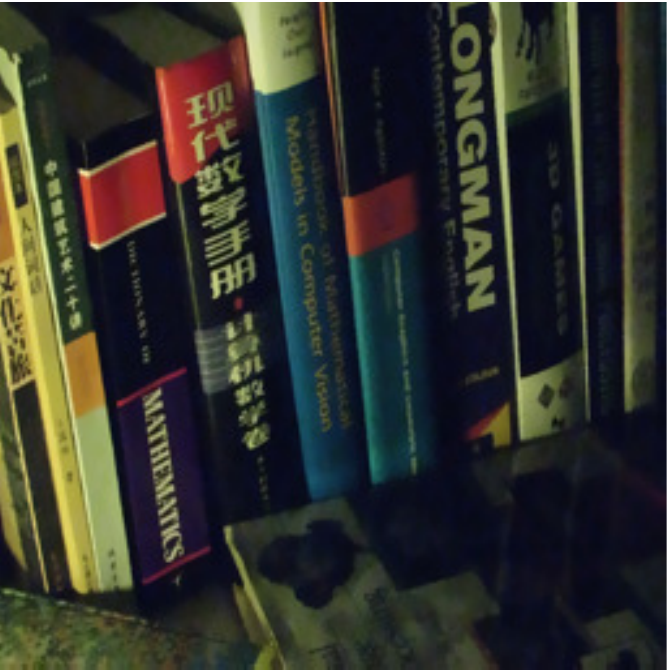}} \hspace{1mm}
\subfloat[\textbf{(d) With infrared data.}]{\includegraphics[width=0.16\linewidth]{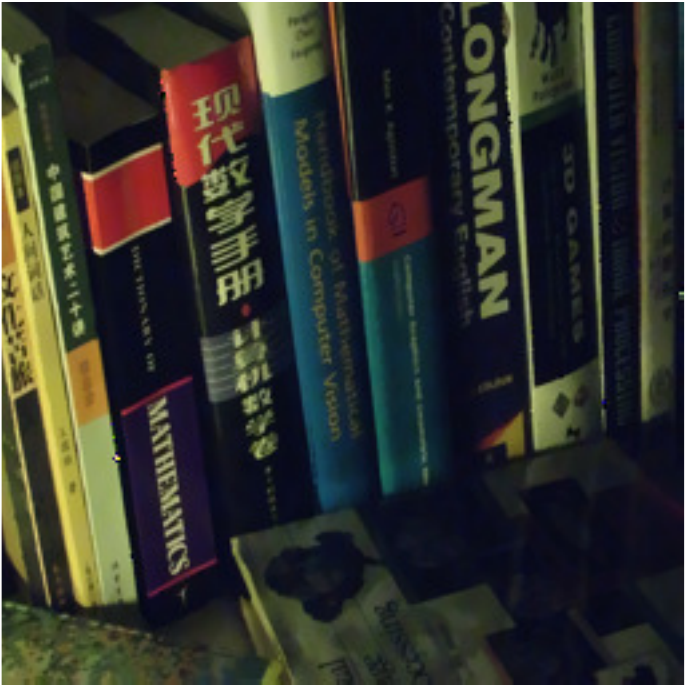}}\hspace{1mm}
\subfloat[(e) \cite{Lin2015brightening}.]{\includegraphics[width=0.16\linewidth]{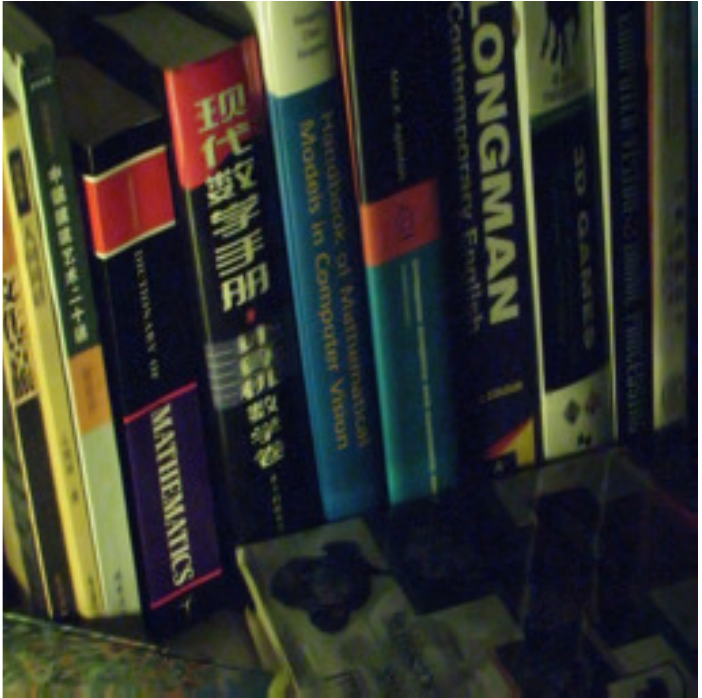}}
\caption{Nonlocal means denoising of a low-light image using \cite{Lin2015brightening} and the proposed method, with and without infrared data \cite{zhuo2010enhancing}. 
Notice that we obtain better denoising by taking the infrared information into account (see zoomed sections).}
\label{Visualfig5}
\end{figure*}   
\begin{figure}
\centering
\subfloat[Clean/Noisy($\sigma = 0.1$).]{\includegraphics[width=0.36\linewidth,height=0.46\linewidth]{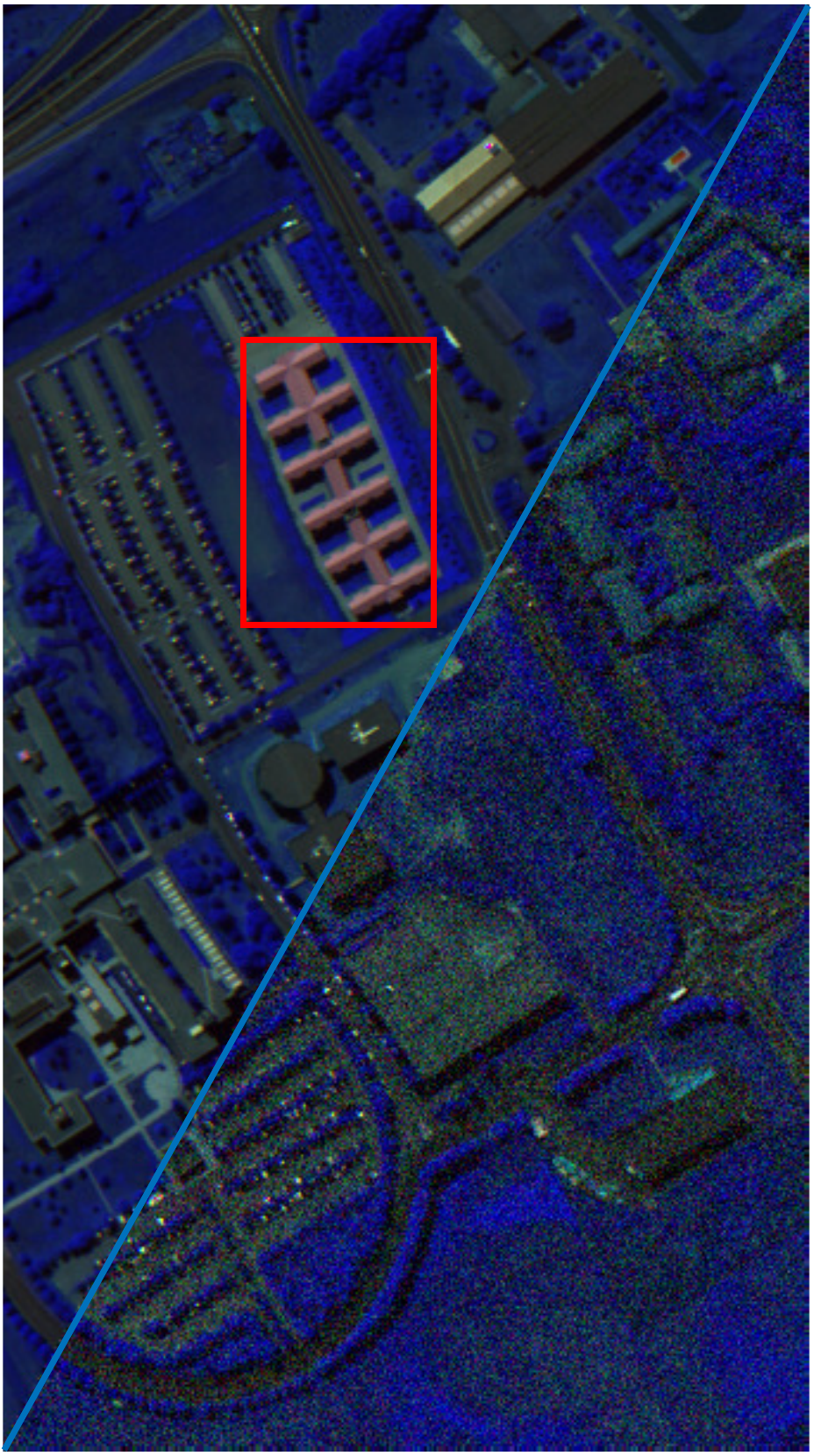}} \hspace{2mm}
\subfloat[\textbf{Proposed.}]{\includegraphics[width=0.36\linewidth,height=0.46\linewidth]{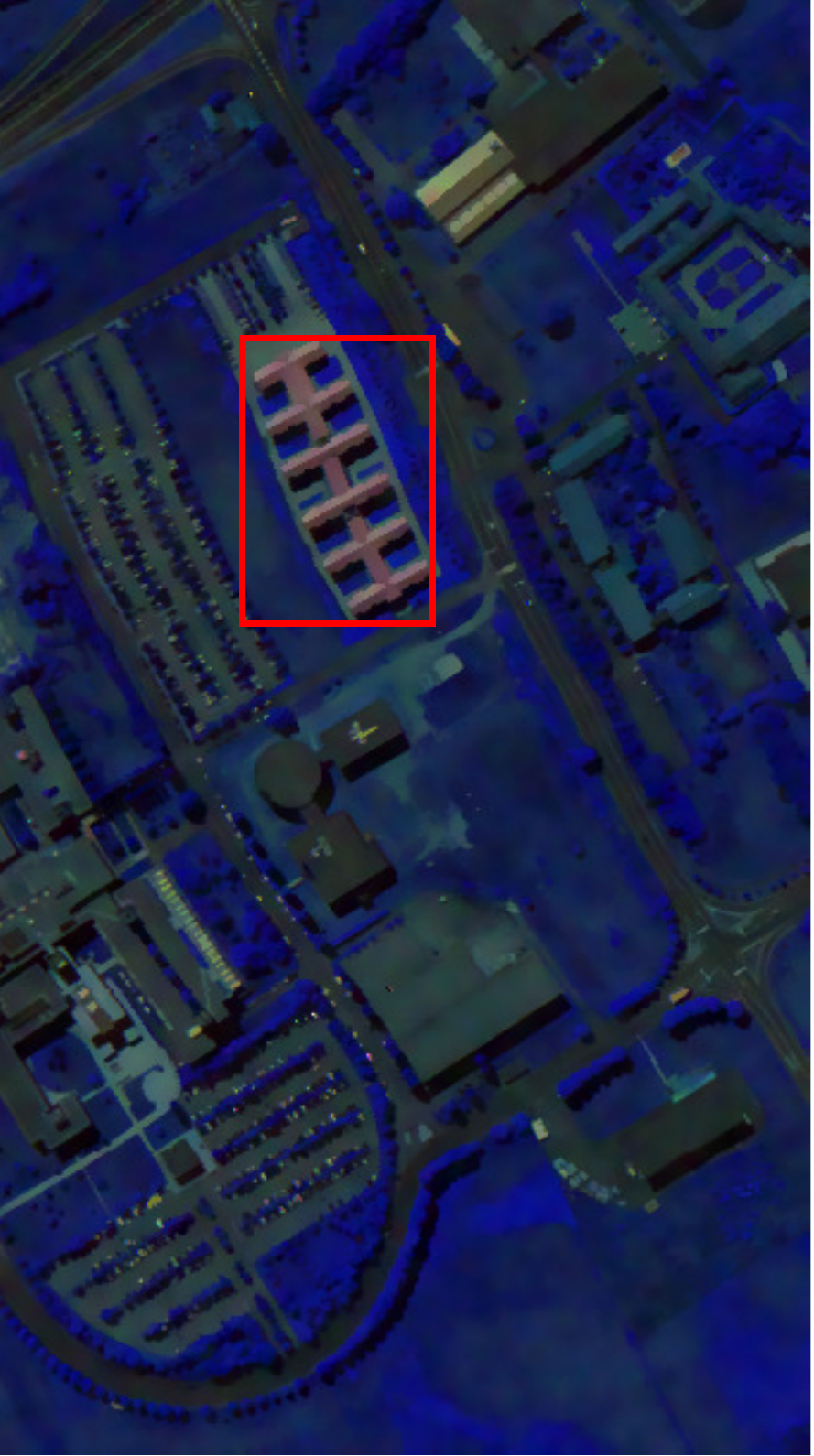}}

\subfloat[\cite{fan2017hyperspectral}.]{\includegraphics[width=0.36\linewidth,height=0.46\linewidth]{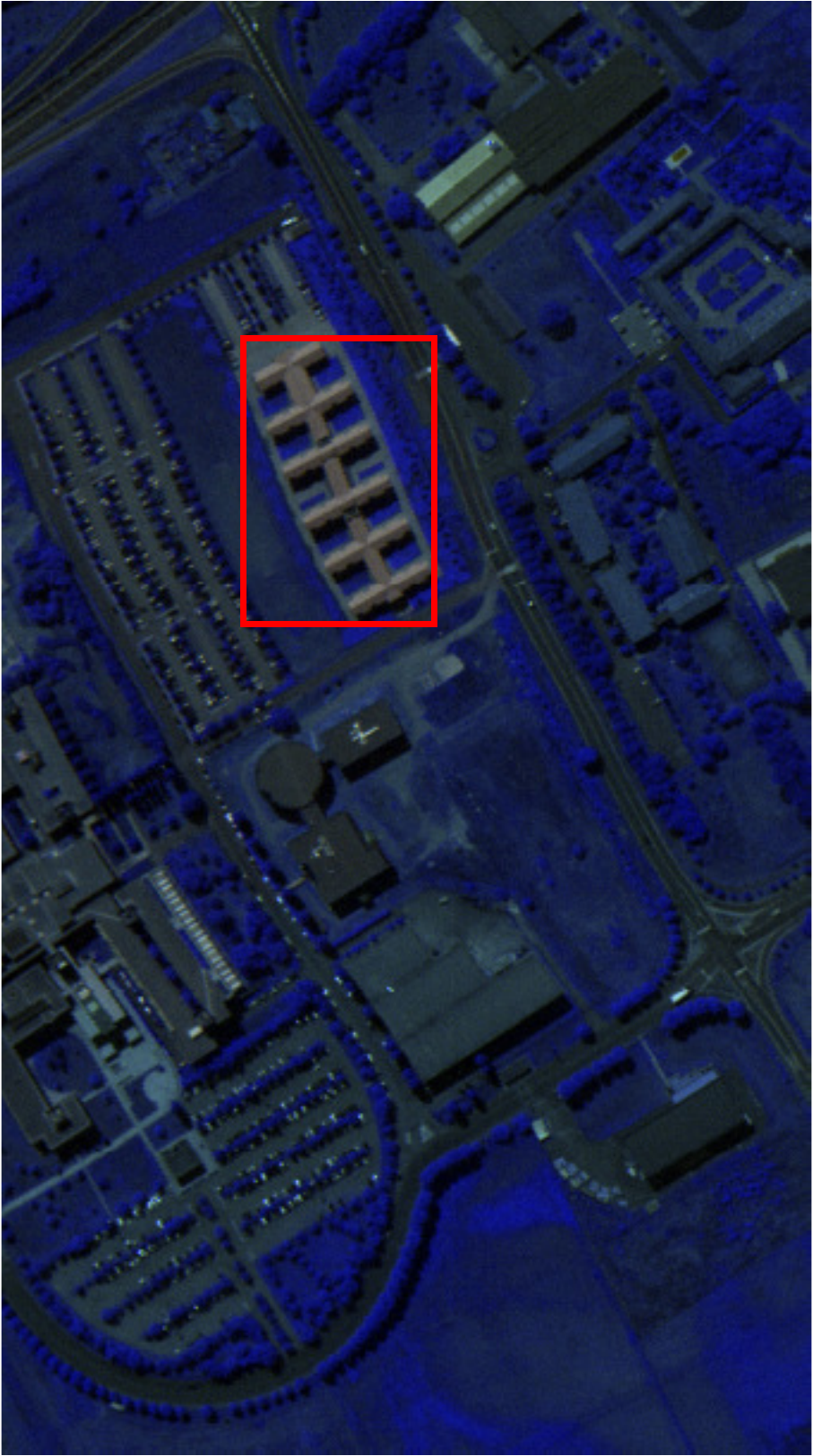}} \hspace{2mm}
\subfloat[\cite{zhao2015hyperspectral}.]{\includegraphics[width=0.36\linewidth,height=0.46\linewidth]{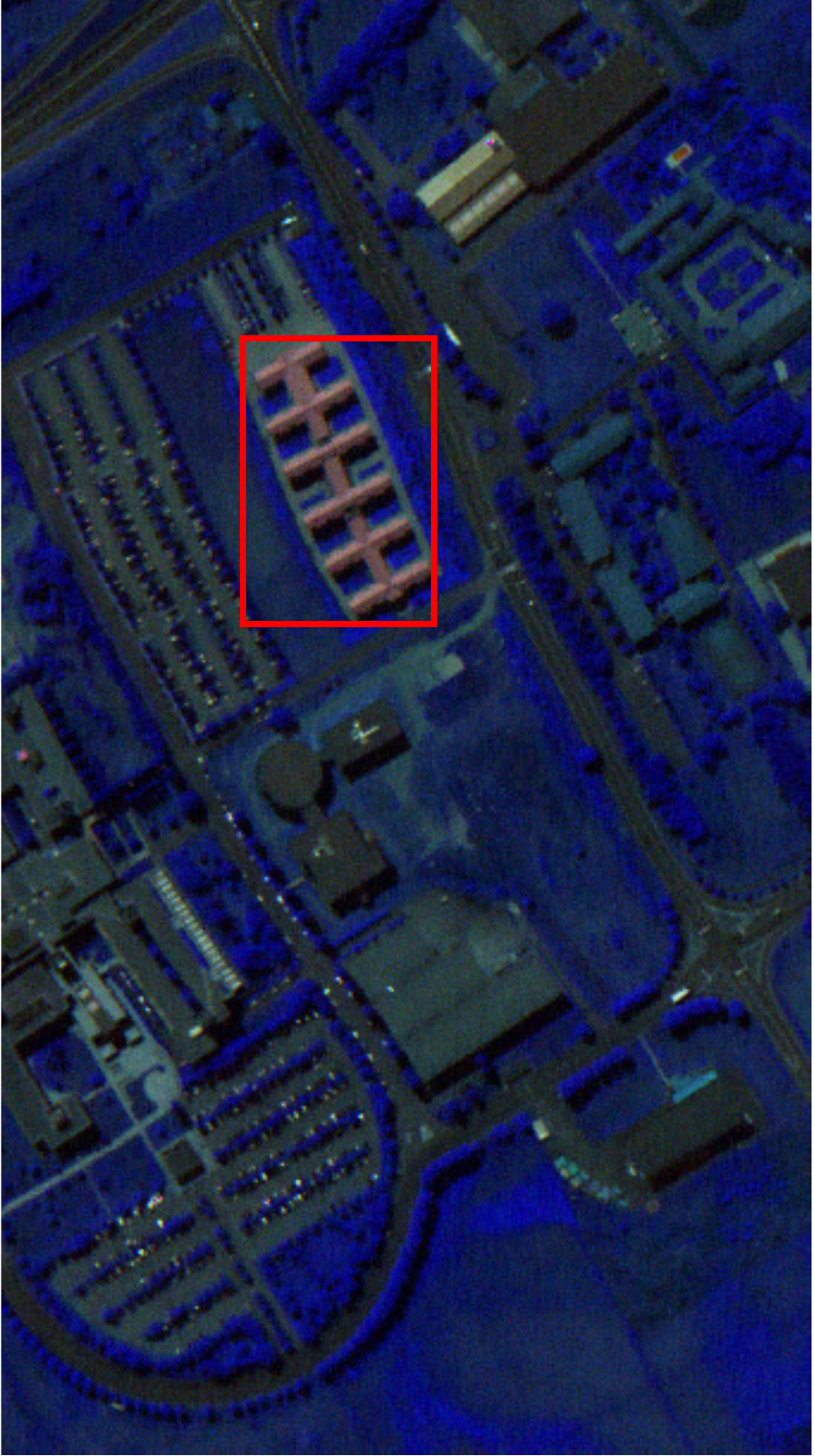}} 
\caption{Hyperspectral denoising of a natural image corrupted with Gaussian noise. Image size: ($610 \times 340) \times 200$ bands. For our method, $\sigma_s=3$, $\sigma_r=100,$ and $K=32$. 
We used $4$ iterations for \cite{fan2017hyperspectral} and $1$ iteration for \cite{zhao2015hyperspectral}. 
The (timing, MPSNR, MSSIM) for (b), (c) and (d) are \textbf{($\textbf{36}$ sec, $\textbf{31.29}$ dB, $\textbf{0.87}$)}, ($192$ sec, $30.54$ dB, $0.89$) and ($1250$ sec, $30.09$ dB, $0.74$).}
\label{Hyperspectral1}
\end{figure} 
Moreover, we also compare with recent optimization-based denoising methods \cite{fan2017hyperspectral,zhao2015hyperspectral}, where parameters are tuned accordingly. 
Visual and quantitative comparisons are shown in Figure \ref{Hyperspectral1} (Pavia dataset). 
For quantitative comparisons, we have used MPSNR and MSSIM, which are simply the PSNR and SSIM values averaged over the spectral bands. 
We notice that the restoration obtained using our method is better than \cite{fan2017hyperspectral,zhao2015hyperspectral} (source code made public by authors), 
which is supported by the metrics shown in the figures. The same is visually evident from a comparison of the boxed regions in Figure \ref{Hyperspectral1}. In particular, the color is not restored properly in \cite{fan2017hyperspectral}, and grains can be seen in \cite{zhao2015hyperspectral}. As expected, we are much faster than these iterative methods, since we perform the filtering in one shot. 

\subsection{Low-light Denoising}
Finally, we use our fast algorithm for NLM denoising of low-light images using additional infrared data \cite{zhuo2010enhancing}.  
In Figure \ref{Visualfig5}, we have shown a visual result which compares our method, both with and without infrared data, and \cite{Lin2015brightening} (an optimization method). 
We set the patch and search window sizes as $7 \times 7$ and $17 \times 17$. 
We used an anisotropic Gaussian kernel, where $\sigma_r = 50$ for the low-light data and $\sigma_r=10$ for the infrared data. 
The dimension was reduced from $3 \times 7 \times 7$ to $6$ using PCA, and the infrared data was added as the seventh dimension. We used $32$ clusters for our method. 
In Figure \ref{Visualfig5}, notice that some salient features are lost if we solely use the low-light input as the guide. 
Instead, if we add the infrared data to the guide, then the restored image is sharper and the features are more apparent compared to \cite{Lin2015brightening} and NLM without infrared data. 

\subsection{Flow-field Denoising}
\begin{figure}[!htp]
\centering
\subfloat[Clean flow-field.]{\includegraphics[width=0.45\linewidth]{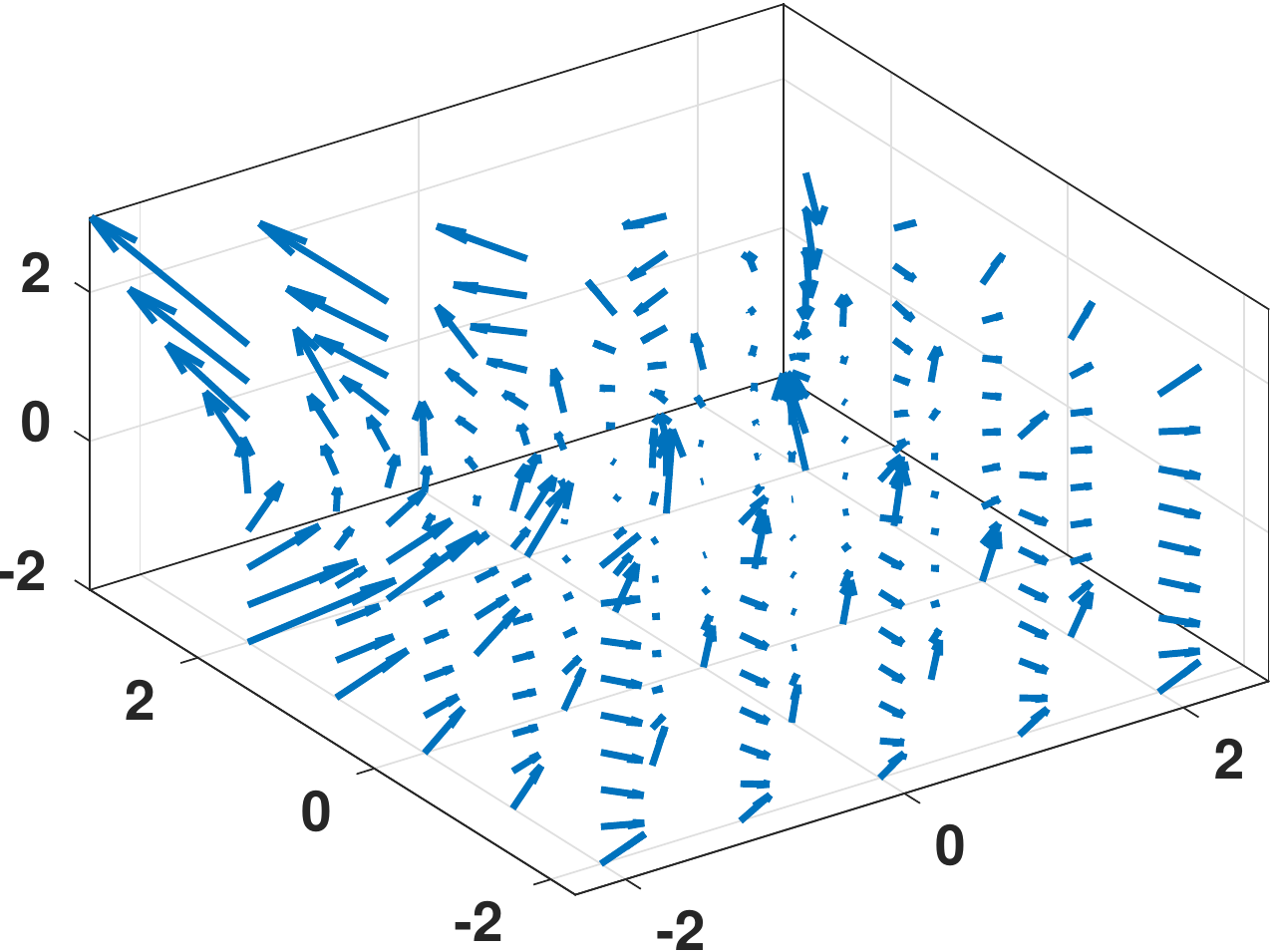}} \hspace{2mm}
\subfloat[Noisy flow-field ($\sigma= 7.5$).]{\includegraphics[width=0.45\linewidth]{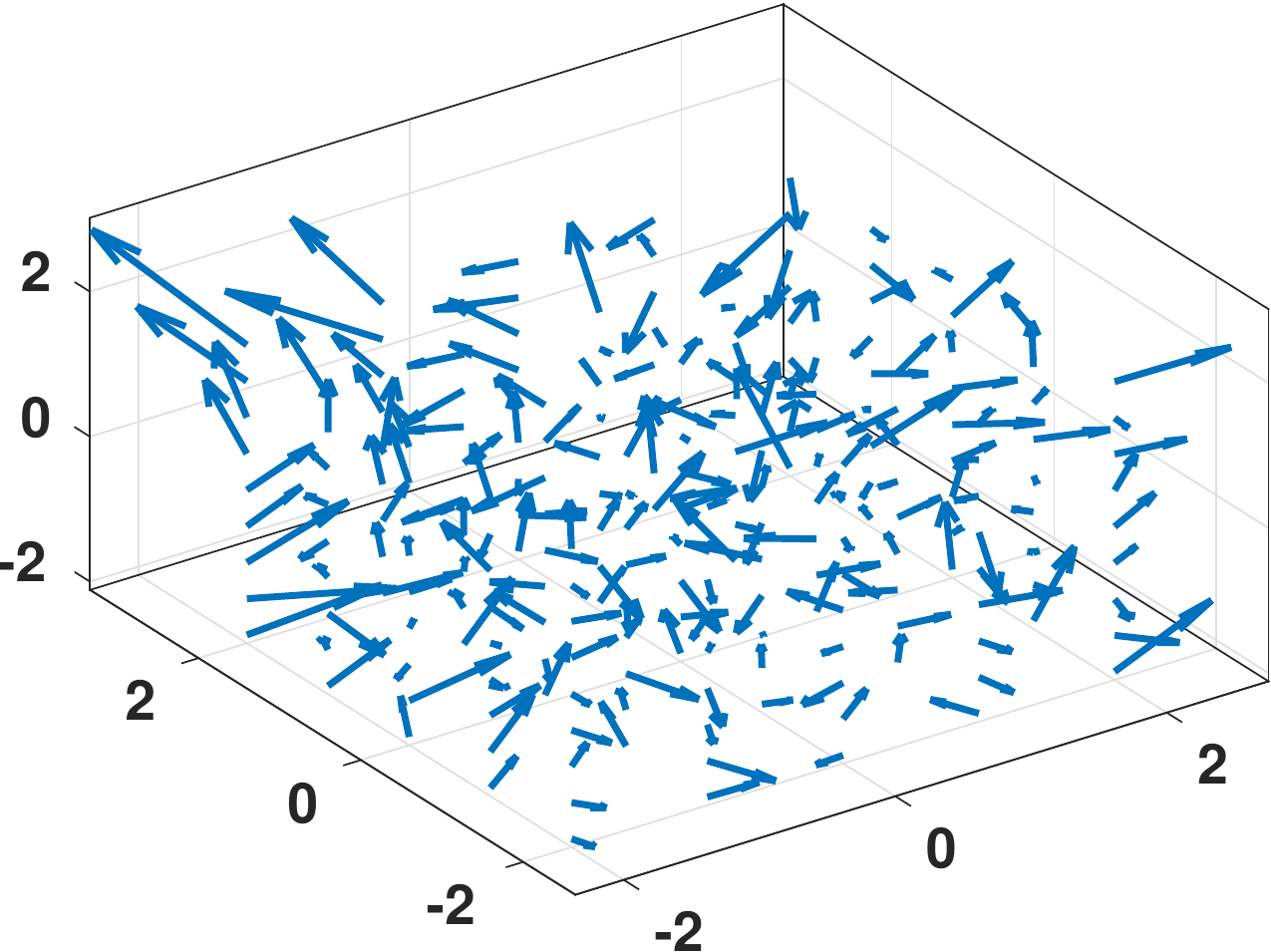}} 

\subfloat[NLM output.]{\includegraphics[width=0.45\linewidth]{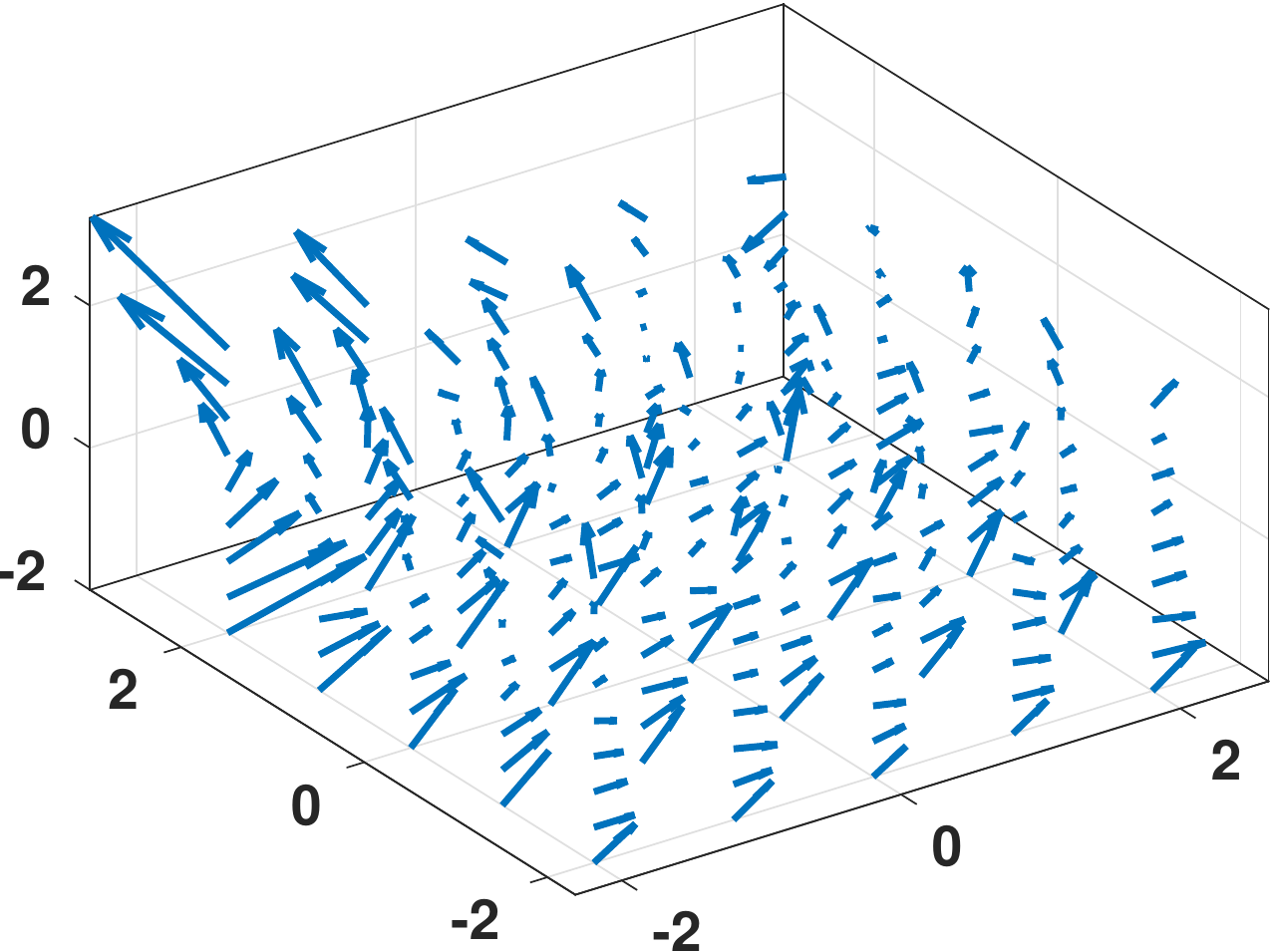}} \hspace{2mm} 
\subfloat[Gaussian smoothing.]{\includegraphics[width=0.45\linewidth]{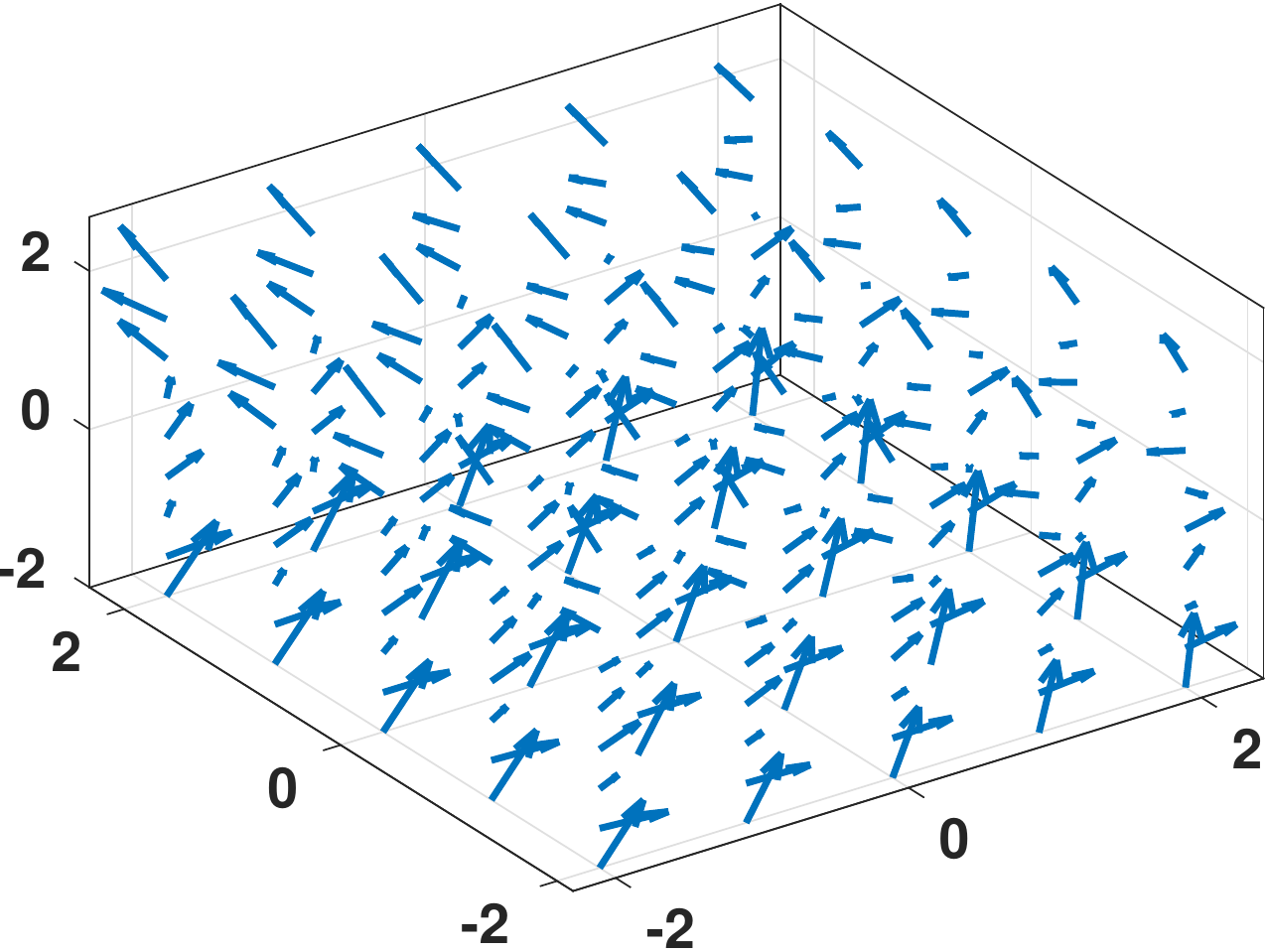}}
\caption{Denoising of a synthetic flow-field using proposed NLM approximation and Gaussian smoothing. 
The synthetic flow-field used is  $u(x, y, z)=x^2-ze^y, v(x, y, z)=y^3-xze^y,$ and $w(x, y, z)=z^4-xe^y$.
The patch and search window sizes for NLM are $7 \times 7 \times 7$ and $21 \times 21 \times 21$, and $\sigma_r = 7.5$. 
We have used PCA to reduce the dimension of the patch from $3 \times 7^3$ to $6$.
Notice that isotropic Gaussian smoothing fails to preserve sharp changes in the flow direction.}
\label{Visualfig9}
\end{figure}

Finally, we apply the proposed method for flow-field denoising. In particular, sharp directional changes in the flow can be preserved much better using NLM, while simultaneously removing the noise (with Gaussian filtering as the baseline).
An instance of flow-field denoising \cite{westenberg2005denoising}  is shown in Figure \ref{Visualfig9} and compared with Gaussian smoothing. 

\section{Conclusion}
\label{Conclusion}

We proposed a framework for fast high-dimensional filtering by approximating both the data and the kernel.
In particular, we derived an algorithm that fuses the scalability of the former with the approximation capability of the latter. 
At the core of our algorithm is the concept of shiftable approximation, which allows us to interpret the coefficients in the framework of Durand et al. \cite{durand2002fast} from an approximation-theoretic point of view. 
We proposed an efficient method for determining the shifts (centers) and the coefficients using $K$-means clustering (inspired by \cite{mozerov2015global}) and data-driven optimization.
Though the proposed algorithm is conceptually simple and easy to implement (about $15$ lines of code), it was shown to yield promising results for diverse applications. In particular, our algorithm was shown to be competitive with state-of-the-art methods for fast bilateral and nonlocal means filtering of color images.  

\section*{Acknowledgements}
The authors thank the editor and the anonymous reviewers for their comments and  suggestions. We also thank the authors of \cite{zhang2017beyond,gastal2012adaptive,mozerov2015global,fan2017hyperspectral} for distributing their code. 
We also thank Jingxiang Yang for helping us debug the code of \cite{zhao2015hyperspectral}.

\section{Appendix}

\subsection{Derivation of \eqref{HDapprox} and \eqref{approxDen}}
\label{derivation}

We recall that the basis of the approximation is the following: For $\x=\p(\i-\j), \j \in W$, we replace $\varphi(\x-\p(\i))$ in \eqref{num} and \eqref{den} with $\sum_{k=1}^K c_{k}(\i) \varphi(\x - \amu_k)$.
That is, we approximate the numerator of \eqref{num} with
\begin{equation}
\label{t1}
\sum_{\j \in W} \omega(\j) \left(\sum_{k=1}^K c_{k}(\i) \varphi(\p(\i-\j) - \amu_k)\right) \f(\i-\j),
\end{equation}
\begin{equation}
\label{t2}
\text{and \eqref{den} with} \ \sum_{\j \in W} \omega(\j) \left(\sum_{k=1}^K c_{k}(\i) \varphi(\p(\i-\j) - \amu_k)\right).
\end{equation}
Exchanging the sums, we can write \eqref{t1} as
\begin{align*}
& \indent \sum_{k=1}^K   c_{k}(\i)  \left(\sum_{\j \in W}  \omega(\j) \varphi(\p(\i-\j) - \amu_k) \f(\i-\j)\right) \\
&=\sum_{k=1}^K   c_{k}(\i)\boldsymbol{v}_k(\i),
\end{align*}
where $\boldsymbol{v}_k$ is defined in \eqref{temp1}. Similarly, we can write \eqref{t2} as $\sum_{k=1}^K   c_{k}(\i)r_k(\i)$,
where $r_k$ is defined in \eqref{temp2}. This completes the derivation of \eqref{HDapprox} and \eqref{approxDen}.

\subsection{Proof of Theorem \ref{theorem}}
\label{thoremproof}
For notational convenience, let  $\g(\i) = \boldsymbol{\xi}(\i)/\eta(\i)$, where
\begin{equation*}
\boldsymbol{\xi}(\i)=\sum_{\j\in W} \omega(\j) \ \varphi\big(\p(\i-\j) - \p(\i)\big) \f(\i-\j),
\end{equation*}
and $\eta$ is as defined in \eqref{den}. For some fixed $\i \in \Omega$, assume that $\p(\i) \in \mathcal{C}_s$ where $1 \leq s \leq K$. 
Then, following \eqref{HDapprox2}, 
\begin{equation}
\label{numApprox}
\hat{\g}(\i) = \frac{\boldsymbol{v}_s(\i)}{r_s(\i)}.
\end{equation}
Note that we can write
\begin{eqnarray*}
\hat{\g}(\i) - \g(\i) = \frac{1}{\eta(\i)} \Big(\hat{\g}(\i) \big( \eta(\i)-r_s(\i) \big) + \big(\boldsymbol{v}_s(\i) - \boldsymbol{\xi}(\i)\big)\Big).
\end{eqnarray*}
By triangle inequality, we can bound $\lVert{\hat{\g}}(\i) - \g(\i)\rVert$ using
\begin{equation}
\label{main}
\frac{1}{\lvert\eta(\i)\rvert} \Big( \sqrt{nR}\ \lvert \eta(\i)-r_s(\i) \rvert  + \lVert \boldsymbol{v}_s(\i) - \boldsymbol{\xi}(\i) \rVert \Big),
\end{equation}
where we have used the fact that $\hat{\g}(\i) \in [0,R]^n$. This follows from \eqref{temp1}, \eqref{temp2}, and \eqref{numApprox}. Now
\begin{equation*}
\boldsymbol{v}_s(\i)-\boldsymbol{\xi}(\i)=\sum_{\j\in W} \omega(\j)  \delta(\i,\j)  \f(\i-\j),
\end{equation*}
\begin{equation*}
\text{where} \ \delta(\i,\j) =\varphi\left( \p(\i-\j) - \p(\i)\right)  -  \varphi\left(\p(\i-\j) - \amu_s\right).
\end{equation*}
\begin{equation*}
\text{By Lipschitz property,} \ \lVert \delta(\i,\j)  \rVert \leq L  \lVert \p(\i) - \amu_s \rVert.
\end{equation*}
Note that we can assume that each $\omega(\j)$ is in $[0,R]$. This is simply because the weights appear both in the numerator and denominator in \eqref{num} and \eqref{numApprox}. Moreover, since the range of $\f$ is $[0,R]^n$, using triangle inequality again, we obtain
\begin{align}
\lVert \boldsymbol{v}_s(\i)-\boldsymbol{\xi}(\i) \rVert \leq L \ \lvert W \rvert \sqrt{nR}  \lVert \p(\i) - \amu_s \rVert.
\label{main1}
\end{align}
Similarly,
\begin{equation}
\ \lvert {\eta}(\i)- r_s(\i) \rvert \leq \  L \lvert  W \rvert  \lVert \p(\i) - \amu_s \rVert.
\label{main2}
\end{equation}
Now, since $\omega$ and $\varphi$ are non-negative, $\eta(\i) \geq \omega(\boldsymbol{0}) \varphi(\boldsymbol{0})$.
Using the fact that $1/\lvert \eta(\i) \rvert \leq 1/\omega(\boldsymbol{0}) \varphi(\boldsymbol{0})$ along with \eqref{main}, \eqref{main1} and \eqref{main2}, we obtain
\begin{equation}
\label{last}
\lVert {\hat{\g}}(\i) - \g(\i)  \rVert \leq  C \lvert W \rvert \sqrt{n} L \lVert \p(\i) - \amu_s \rVert,
\end{equation}
where $C=2\sqrt{R}/\omega(\boldsymbol{0}) \varphi(\boldsymbol{0})$. On squaring \eqref{last} and summing over all pixels, we arrive at \eqref{bound}.

\bibliographystyle{IEEEtran}
\bibliography{manuscript}

\end{document}